\documentclass[11pt]{article}

\usepackage{acl}

\usepackage{times}
\usepackage{latexsym}

\usepackage[T1]{fontenc}

\usepackage[utf8]{inputenc}

\usepackage{microtype}

\usepackage{inconsolata}

\usepackage{graphicx}

\usepackage{bm}
\usepackage{amsmath}
\usepackage{amsfonts}
\usepackage{tcolorbox}
\usepackage{booktabs}
\usepackage{colortbl}
\usepackage{enumitem}
\usepackage{subcaption}
\usepackage[table]{xcolor}
\usepackage{url}

\definecolor{thm}{RGB}{69, 53, 193}

\newtcolorbox{findings}[1][]{
  colback=thm!5!white,
  colframe=thm!60!black,
  left=1.5mm,
  right=1.5mm,
  top=1mm,
  bottom=1mm,
}

\title{Revisiting Entropy in Reinforcement Learning for Large Reasoning Models}

\author{
Renren Jin$^{1}$, Pengzhi Gao$^{2}$, Yuqi Ren$^{1}$, Zhuowen Han$^{1}$, Tongxuan Zhang$^{3}$\\ 
\textbf{Wuwei Huang}$^{2}$\textbf{,} \textbf{Wei Liu}$^{2}$\textbf{,} \textbf{Jian Luan}$^{2}$\textbf{,} \textbf{Deyi Xiong}$^{1}$\thanks{~Corresponding author.}\\
$^{1}$TJUNLP Lab, School of Computer Science and Technology, Tianjin University, China\\
$^{2}$Independent Researcher\\
$^{3}$College of Computer and Information Engineering, Tianjin Normal University, China\\
\texttt{\{rrjin, dyxiong\}@tju.edu.cn}\\
}

\begin{document}
\maketitle
\begin{abstract}
Reinforcement learning with verifiable rewards (RLVR) has emerged as a prominent paradigm for enhancing the reasoning capabilities of large language models (LLMs). However, the entropy of LLMs usually collapses during RLVR training, leading to premature convergence to suboptimal local minima and hindering further performance improvement. Although various approaches have been proposed to mitigate entropy collapse, a comprehensive study of entropy in RLVR remains lacking. To bridge this gap, we conduct extensive experiments to investigate the entropy dynamics of LLMs trained with RLVR and analyze how model entropy correlates with response diversity, calibration, and performance across various benchmarks. Our results identify three factors that influence entropy: the clipping thresholds in the optimization objective, the number of off-policy updates, and the diversity of the training data. Furthermore, through both theoretical analysis and empirical validation, we demonstrate that tokens with positive advantages are the primary drivers of entropy collapse. Motivated by this insight, we propose Positive-Advantage Reweighting, a simple yet effective approach that regulates model entropy by adjusting the loss weights assigned to tokens with positive advantages during RLVR training, while maintaining competitive performance.\footnote{The source code is publicly available at \url{https://github.com/cordercorder/EntropyRL}.}
\end{abstract}

\section{Introduction}

Pioneered by OpenAI o1 \citep{DBLP:journals/corr/abs-2412-16720}, DeepSeek-R1 \citep{DBLP:journals/corr/abs-2501-12948}, and Kimi k1.5 \citep{DBLP:journals/corr/abs-2501-12599}, reinforcement learning with verifiable rewards (RLVR) has been widely employed to push the boundaries of reasoning capabilities in large language models (LLMs). LLMs trained with RLVR have demonstrated remarkable performance on tasks that require complex reasoning and offer readily verifiable outcomes, such as mathematics and coding \cite{DBLP:journals/corr/abs-2503-20783,deepcoder2025,DBLP:journals/corr/abs-2505-22312,yu2026knowrlboostingllmreasoning,DBLP:journals/corr/abs-2602-09953,DBLP:journals/corr/abs-2601-21476}.

However, although RLVR enhances the reasoning ability of LLMs, a growing body of research has shown that it can also drive LLMs toward entropy collapse, wherein the entropy of the model decreases substantially during training, ultimately reaching a markedly low level \citep{DBLP:journals/corr/abs-2503-14476,DBLP:journals/corr/abs-2508-11016}. Entropy collapse indicates that the probability mass over the model’s vocabulary becomes concentrated on a limited subset of tokens. This phenomenon further implies that LLMs increasingly prioritize exploitation over exploration. As a result, they fail to effectively explore novel reasoning paths during training, thereby potentially causing premature convergence to a local optimum.

To address entropy collapse in LLMs, numerous methods have been proposed. \citet{shen2025entropycontrolllmrlalgorithms} and \citet{DBLP:journals/corr/abs-2505-22312} incorporate an entropy maximization term into the RLVR objective. DAPO \citep{DBLP:journals/corr/abs-2503-14476} raises the upper clipping bound of the importance ratio to avoid clipping low-probability tokens. \citet{DBLP:journals/corr/abs-2505-22617} restrict parameter updates for tokens with high covariance between log-probability and advantage, alongside other approaches \citep{DBLP:journals/corr/abs-2506-01939,DBLP:journals/corr/abs-2508-10751,DBLP:journals/corr/abs-2508-02260,DBLP:journals/corr/abs-2508-11016,DBLP:journals/corr/abs-2506-01347}. Despite these advances, systematic investigations of entropy in RLVR remain scarce. Specifically, three critical questions are still underexplored: (1) \textbf{How does the entropy of LLMs trained with RLVR correlate with their performance?} (\S\ref{sec:RQ_1}) (2) \textbf{What factors govern entropy dynamics, both theoretically and empirically?} (\S\ref{sec:RQ_2}) and (3) \textbf{How can entropy be effectively regulated to improve the performance of LLMs?} (\S\ref{sec:RQ_3})

To investigate the above research questions, we conduct extensive experiments on RLVR. We find that the entropy of LLMs trained with RLVR is strongly correlated with response diversity, as LLMs with lower entropy tend to produce less diverse outputs (\S\ref{subsec:appendix_entropy_and_response_diversity}). During training, both response entropy and prompt entropy decrease, with in-domain prompt entropy declining more rapidly than that of out-of-domain prompts (\S\ref{subsec:appendix_entropy_dynamics_on_prompts}). Meanwhile, prompt entropy exhibits only a weak correlation with the accuracy of the corresponding responses (\S\ref{subsec:appendix_prompt_entropy_and_accuracy}). Notably, we observe that model performance can continue to improve without sacrificing entropy (\S\ref{subsec:performance_gains_without_trading_off_entropy}). In addition, entropy does not serve as a reliable proxy for model performance across most benchmarks, as the observed correlations between entropy and performance are highly task-dependent (\S\ref{subsec:correlations_between_entropy_and_model_performance}). We further observe that entropy collapse is closely associated with model miscalibration, and that more severe entropy collapse corresponds to stronger miscalibration (\S\ref{subsec:entropy_collapse_and_miscalibration}). Beyond these empirical findings, we identify three factors influencing entropy dynamics: (1) the clipping threshold (\S\ref{subsec:clipping_threshold}), (2) the number of off-policy updates (\S\ref{subsec:off_policy_updates}), and (3) training data diversity (\S\ref{subsec:training_data_diversity}). Remarkably, an LLM trained on approximately 600 samples can achieve performance comparable to one trained on around 17k samples (\S\ref{subsec:training_data_diversity}). Finally, through both theoretical analysis and empirical validation (\S\ref{subsec:theoretical_analysis} and \S\ref{subsec:empirical_analysis}), we demonstrate that tokens with positive advantages are the primary drivers of entropy collapse. Motivated by this insight, we propose \textbf{Positive-Advantage Reweighting}, which adjusts the loss weights of tokens with positive advantages to regulate model entropy while improving performance (\S\ref{subsec:positive-advantage_reweighting}).
Our contributions can be summarized as follows:
\begin{itemize}[itemsep=0pt, leftmargin=*]
\item We conduct extensive experiments to investigate the dynamics of prompt entropy and response entropy in LLMs trained with RLVR, revealing how entropy correlates with response diversity, model calibration, and performance.
\item We identify three factors influencing the entropy dynamics of LLMs during RLVR training: (1) clipping threshold, (2) off-policy updates, and (3) training data diversity.
\item We theoretically and empirically demonstrate that entropy collapse in RLVR primarily arises from positive-advantage tokens, and propose \textbf{Positive-Advantage Reweighting} to control LLM entropy and improve performance by dynamically reweighting the losses of such tokens.
\end{itemize}

\section{Related Work}

Entropy has long served as a regularization mechanism to encourage exploration in reinforcement learning \citep{DBLP:conf/aaai/ZiebartMBD08,DBLP:phd/us/Ziebart18}. In the era of LLMs \citep{DBLP:journals/corr/abs-2303-18223,DBLP:journals/corr/abs-2309-15025,DBLP:journals/corr/abs-2310-19736,DBLP:journals/corr/abs-2412-17686,DBLP:journals/corr/abs-2509-02547}, several studies use entropy as a signal to determine when agents should seek experience guidance and perform additional partial sampling \citep{DBLP:journals/corr/abs-2507-19849,DBLP:journals/corr/abs-2601-08605}. In a related vein, entropy maximization objectives have been incorporated into RLVR to encourage exploration and prevent premature convergence \citep{shen2025entropycontrolllmrlalgorithms,DBLP:journals/corr/abs-2505-22312}. To mitigate entropy collapse, Clip-Higher \citep{DBLP:journals/corr/abs-2503-14476} raises the upper clipping bound of the importance ratio to prevent low-probability tokens with positive advantages from being clipped. \citet{zhihu2025entropy} and \citet{DBLP:journals/corr/abs-2505-22617} provide theoretical insights into entropy dynamics, showing that tokens with strong positive covariance between their probabilities and corresponding advantages primarily drive entropy collapse. Building on these insights, Clip-Cov and KL-Cov \citep{DBLP:journals/corr/abs-2505-22617} limit updates on tokens exhibiting such covariance. Similarly, CE-GPPO \citep{su2025cegppocoordinatingentropygradientpreserving} mitigates entropy collapse by preserving the gradients of clipped tokens through a stop-gradient operation. Furthermore, \citet{DBLP:journals/corr/abs-2506-01939} train LLMs using only high-entropy tokens to enhance model performance, while \citet{DBLP:journals/corr/abs-2506-14758} incorporate entropy terms into the advantage to encourage exploration and improve reasoning capabilities.  Numerous other studies have also explored entropy-based mechanisms to further enhance the performance of LLMs trained with RLVR \citep{DBLP:journals/corr/abs-2508-11016,wang2025harnessinguncertaintyentropymodulatedpolicy,liu2025uniformheterogeneoustailoringpolicy}.

\section{Preliminaries}

\subsection{Entropy Regularization}

Let $\bm{x}$ denote the prompt and $\bm{y}$ the response generated by the LLM $\pi_{\theta}$ parameterized by $\theta$. We define $\mathcal{H} \left( \pi_{\theta} \right)$ as the token-level average entropy of $\pi_{\theta}$. Entropy regularization augments the GRPO objective with the term $\alpha \mathcal{H} \left( \pi_{\theta} \right)$, where $\alpha$ is the entropy regularization coefficient. The complete formulation of GRPO and its corresponding objective are provided in Appendix~\ref{subsec:appendix_GRPO}.

\subsection{Adaptive Entropy Regularization}
Although vanilla entropy regularization can mitigate entropy collapse, selecting an appropriate regularization coefficient is challenging \citep{DBLP:journals/corr/abs-2505-22312}: a large coefficient causes entropy to rise rapidly, while a small one renders the regularization ineffective. To address this issue, \citet{DBLP:journals/corr/abs-2505-22312} propose adaptive entropy regularization, which dynamically adjusts the coefficient according to the model’s entropy rather than fixing it during training. Let $\mathcal{H}_k \left( \pi_{\theta} \right)$ denote the entropy of the LLM $\pi_{\theta}$ at training step $k$. To prevent entropy from falling below a predefined threshold $\delta$, the entropy regularization coefficient $\alpha_k$ is defined as follows:
\begin{equation}
\small
    \alpha_{k} = c_k \cdot \mathbb{I} \left\{ \mathcal{H}_k \left( \pi_{\theta} \right) < \delta \right\},
\label{eq:adaptive_entropy_equ_1}
\end{equation}
where the adaptive coefficient $c_k$ is updated according to the following rule:
\begin{equation}
\small
c_{k+1} = c_k + \beta \cdot \mathbb{I} \left\{ \mathcal{H}_k \left( \pi_{\theta} \right) < \delta \right\} - \beta \cdot \mathbb{I} \left\{ \mathcal{H}_k \left( \pi_{\theta} \right) \ge \delta \right\}.
\label{eq:adaptive_entropy_equ_2}
\end{equation}
Equation (\ref{eq:adaptive_entropy_equ_1}) indicates that entropy regularization is applied only when the entropy falls below $\delta$, setting the coefficient to $c_k$. Equation (\ref{eq:adaptive_entropy_equ_2}) defines the update rule: if the entropy is below $\delta$, $c_k$ increases by $\beta$; otherwise, it decreases by $\beta$.

\section{Experimental Setup}
\label{sec:experimental_setup}

We trained Qwen2.5-Math-7B \citep{DBLP:journals/corr/abs-2409-12122} with GRPO using the veRL framework \citep{DBLP:conf/eurosys/ShengZYWZZPL025} on the DAPO-Math-17K dataset \citep{DBLP:journals/corr/abs-2503-14476}. We evaluated the trained models on both in-domain and out-of-domain benchmarks. The in-domain benchmarks included AIME 2024/2025 \citep{aime2024,aime2025}, MATH500 \citep{DBLP:conf/nips/HendrycksBKABTS21}, AMC 2023 \citep{amc2023}, and Minerva Math \citep{DBLP:conf/nips/LewkowyczADDMRS22}, while out-of-domain evaluation covered LiveCodeBench \citep{DBLP:conf/iclr/JainHGLYZWSSS25} for coding and IF-Eval \citep{DBLP:journals/corr/abs-2311-07911} for instruction following. Detailed experimental settings are provided in Appendix~\ref{sec:appendix_experimental_setup}.

\begin{figure}[t]
    \centering
    \includegraphics[width=\linewidth]{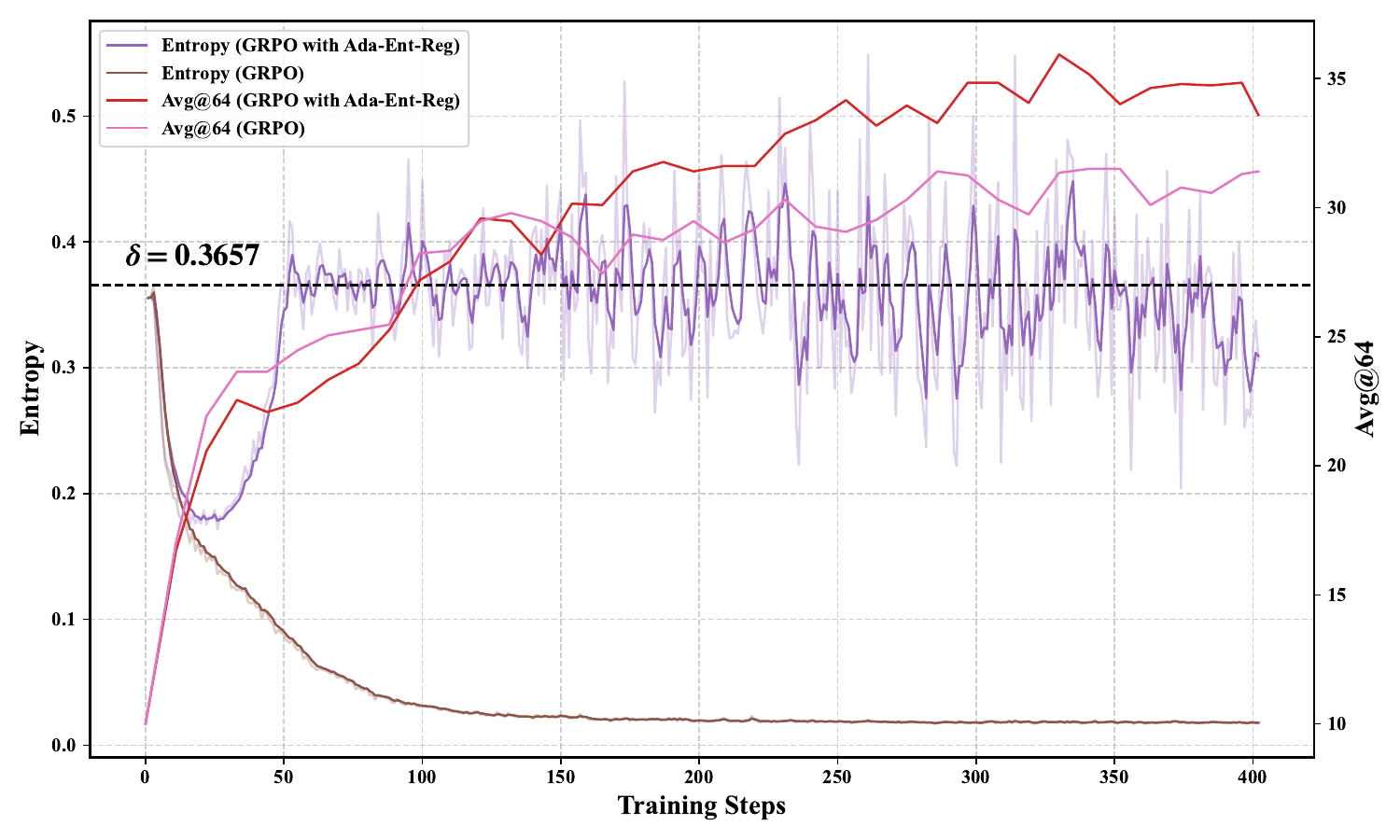}
    \caption{Evolution of LLM entropy and Avg@64 performance on AIME 2024 during RLVR training. ``Ada-Ent-Reg'' denotes adaptive entropy regularization.}
    \label{fig:fixed_entropy_valid_acc}
\end{figure}

\section{How Does the Entropy of LLMs Trained with RLVR Correlate with Their Performance?}
\label{sec:RQ_1}

For the analyses of entropy and response diversity, entropy dynamics on prompts, and the relationship between prompt entropy and accuracy, we summarize the main empirical findings below. Detailed empirical analyses can be found in Appendix~\ref{subsec:appendix_entropy_and_response_diversity}, Appendix~\ref{subsec:appendix_entropy_dynamics_on_prompts}, and Appendix~\ref{subsec:appendix_prompt_entropy_and_accuracy}, respectively.
\begin{findings}
\begin{itemize}[itemsep=0pt, leftmargin=*]
    \item The diversity of responses generated by LLMs is strongly and positively correlated with their entropy during training.
    \item During RLVR training, the entropy of LLMs decreases for both in-domain and out-of-domain prompts, with a more substantial reduction observed for in-domain prompts.
    \item The entropy of LLMs on prompts shows only a weak correlation with their accuracy.
\end{itemize}
\end{findings}

\subsection{Performance Gains Without Trading Off Entropy}
\label{subsec:performance_gains_without_trading_off_entropy}
\begin{findings}
The performance of LLMs continues to improve during training, even as their entropy fluctuates around the value observed before training, indicating that performance gains are not solely achieved by trading off entropy.
\end{findings}

To maintain LLM entropy at a level comparable to that observed prior to training, we apply adaptive entropy regularization during RLVR training. Specifically, we randomly sample 1,000 prompts from the training set and compute the model’s response entropy before training. The computed entropy value is set as the threshold $\delta$, ensuring that LLM entropy remains above $\delta$ throughout training.

Figure~\ref{fig:fixed_entropy_valid_acc} illustrates the evolution of entropy and accuracy on AIME 2024. As shown in Figure~\ref{fig:fixed_entropy_valid_acc}, under adaptive entropy regularization, entropy initially decreases sharply, then increases and fluctuates around the specified threshold. Meanwhile, accuracy on AIME 2024 exhibits an upward trend and even surpasses that of the model trained without adaptive entropy regularization, indicating that performance gains are not merely achieved by trading off entropy. In contrast, LLMs trained without entropy regularization experience rapid entropy collapse, wherein entropy declines to a low value. Correspondingly, the Avg@64 metric rises sharply early in training, it quickly plateaus and remains below that achieved with adaptive entropy regularization. These results suggest that entropy collapse may lead to performance degradation.

\subsection{Correlations Between Entropy and Model Performance}
\label{subsec:correlations_between_entropy_and_model_performance}
\begin{findings}
The correlation between LLM entropy and its performance on benchmarks depends on both the task and the evaluation metric used.
\end{findings}

The Spearman correlation coefficients between LLM entropy and model performance across benchmarks are summarized in Table~\ref{tab:spearman_correlation_coefficients}. As shown, when LLMs are trained with RLVR using only mathematical data, the Avg@64 scores on LiveCodeBench exhibit a strong negative correlation with model entropy during training. This relationship is further illustrated in Figure~\ref{fig:entropy_code_acc_corrcoef_plot_2} in Appendix~\ref{subsec:appendix_correlations_between_entropy_and_model_performance}, which shows a clear negative correlation between LLM entropy and Avg@64 scores on LiveCodeBench. In contrast, the correlations between entropy and performance on other benchmarks are relatively weak, suggesting that the correlation between LLM entropy and performance is highly dependent on both the task and the evaluation metric.

\begin{table}[t]
  \centering
  \small
    \begin{tabular}{lrr}
    \toprule
          & \multicolumn{1}{c}{\textbf{Avg@64}} & \multicolumn{1}{c}{\textbf{Pass@64}} \\
    \midrule
    \textbf{AIME 2024} & 0.41721 & 0.12799 \\
    \textbf{AIME 2025} & 0.04730 & 0.24753 \\
    \textbf{MATH500} & 0.08916 & 0.46253 \\
    \textbf{AMC 2023} & -0.13453 & \textbf{0.59091} \\
    \textbf{Minerva} & -0.14829 & 0.56942 \\
    \textbf{LiveCodeBench} & \textbf{-0.89371} & -0.18895 \\
    \textbf{IF-Eval} & 0.62660 & -0.32060 \\
    \bottomrule
    \end{tabular}
  \caption{Spearman correlation coefficients between LLM entropy and performance across benchmarks.}
  \label{tab:spearman_correlation_coefficients}
\end{table}

\subsection{Entropy Collapse and Miscalibration}
\label{subsec:entropy_collapse_and_miscalibration}
\begin{findings}
    While RLVR enhances the performance of LLMs, it can induce miscalibration, leading models to become overconfident in their responses. This miscalibration typically worsens as entropy collapse becomes more severe.
\end{findings}

We further investigate the relationship between entropy and the calibration of LLMs. For a well-calibrated model, correct responses should receive higher probabilities than incorrect ones. To assess this, we compute the distributions of average per-token log probabilities for correct and incorrect responses generated by LLMs.

\begin{figure}[t]
    \centering
    \includegraphics[width=\linewidth]{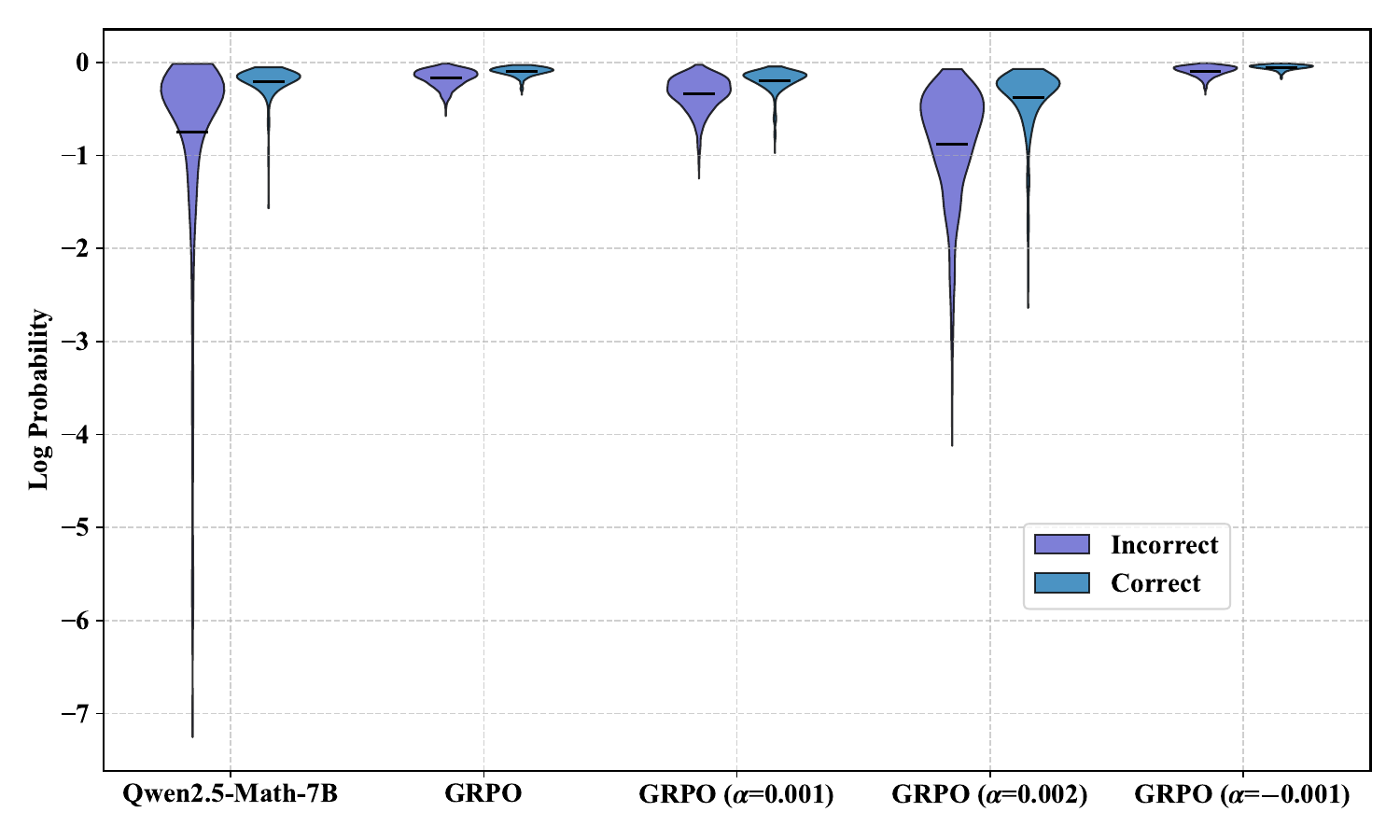}
    \caption{Distribution of log probabilities for correct and incorrect responses. Values in parentheses indicate the entropy regularization coefficients. In each violin plot, the black line denotes the mean.}
    \label{fig:response_acc_probs}
\end{figure}

As shown in Figure~\ref{fig:response_acc_probs}, prior to training, Qwen2.5-Math-7B generally assigns higher probabilities to correct responses than to incorrect ones. However, after GRPO training, the probabilities of both correct and incorrect responses increase, indicating that the model becomes more overconfident. Meanwhile, the probability gap between correct and incorrect responses narrows, suggesting reduced discriminability and poorer calibration, which aligns with the observations of \citet{DBLP:journals/corr/abs-2508-11800}. Furthermore, when the entropy regularization coefficient is negative, thereby promoting entropy collapse, both overconfidence and miscalibration become more pronounced. In contrast, employing a positive entropy regularization coefficient mitigates these effects. Considering Figures~\ref{fig:response_acc_probs} and \ref{fig:entropy_ngram_diversity_plot_aime24} in Appendix~\ref{subsec:appendix_entropy_and_response_diversity} jointly, we observe a consistent trend in overconfidence and miscalibration: GRPO ($\alpha=-0.001$) > GRPO > GRPO ($\alpha=0.001$) > GRPO ($\alpha=0.002$). A similar ordering is observed for entropy collapse, suggesting a potential correlation between miscalibration and entropy collapse.

\section{What Factors Govern Entropy Dynamics, Both Theoretically and Empirically?}
\label{sec:RQ_2}

To investigate the factors that shape the entropy dynamics of LLMs during GRPO training, we conduct a series of experiments across three dimensions: (1) the effect of varying clipping thresholds on model entropy and performance (\S\ref{subsec:clipping_threshold}); (2) the impact of off-policy updates on entropy dynamics and performance on both the training and test sets (\S\ref{subsec:off_policy_updates}); and (3) the role of training data diversity in shaping entropy dynamics (\S\ref{subsec:training_data_diversity}).

\subsection{Clipping Threshold}
\label{subsec:clipping_threshold}

\begin{findings}
Increasing the upper clipping threshold in GRPO alleviates entropy collapse, whereas decreasing it exacerbates this phenomenon. A similar trend is observed for the lower clipping threshold: higher values mitigate entropy collapse, while lower values intensify it.
\end{findings}

To examine how clipping thresholds affect the entropy and performance of LLMs, we trained LLMs with GRPO under various lower and upper clipping settings beyond the default $\varepsilon_{\text{low}} = \varepsilon_{\text{high}} = 0.2$. Specifically, we conducted the following experiments: (1) \textbf{Clip-Higher}, with $\varepsilon_{\text{high}} = 0.28$; (2) \textbf{Clip-Lower}, with $\varepsilon_{\text{low}} = 0.28$; (3) \textbf{Clip-Tighter}, with $\varepsilon_{\text{low}} = 0.12$; and (4) \textbf{Clip-Free}, which removes clipping from the GRPO objective. Detailed descriptions are provided in Appendix~\ref{subsec:appendix_clipping_threshold}.

\begin{figure}[t]
    \centering
    \includegraphics[width=\linewidth]{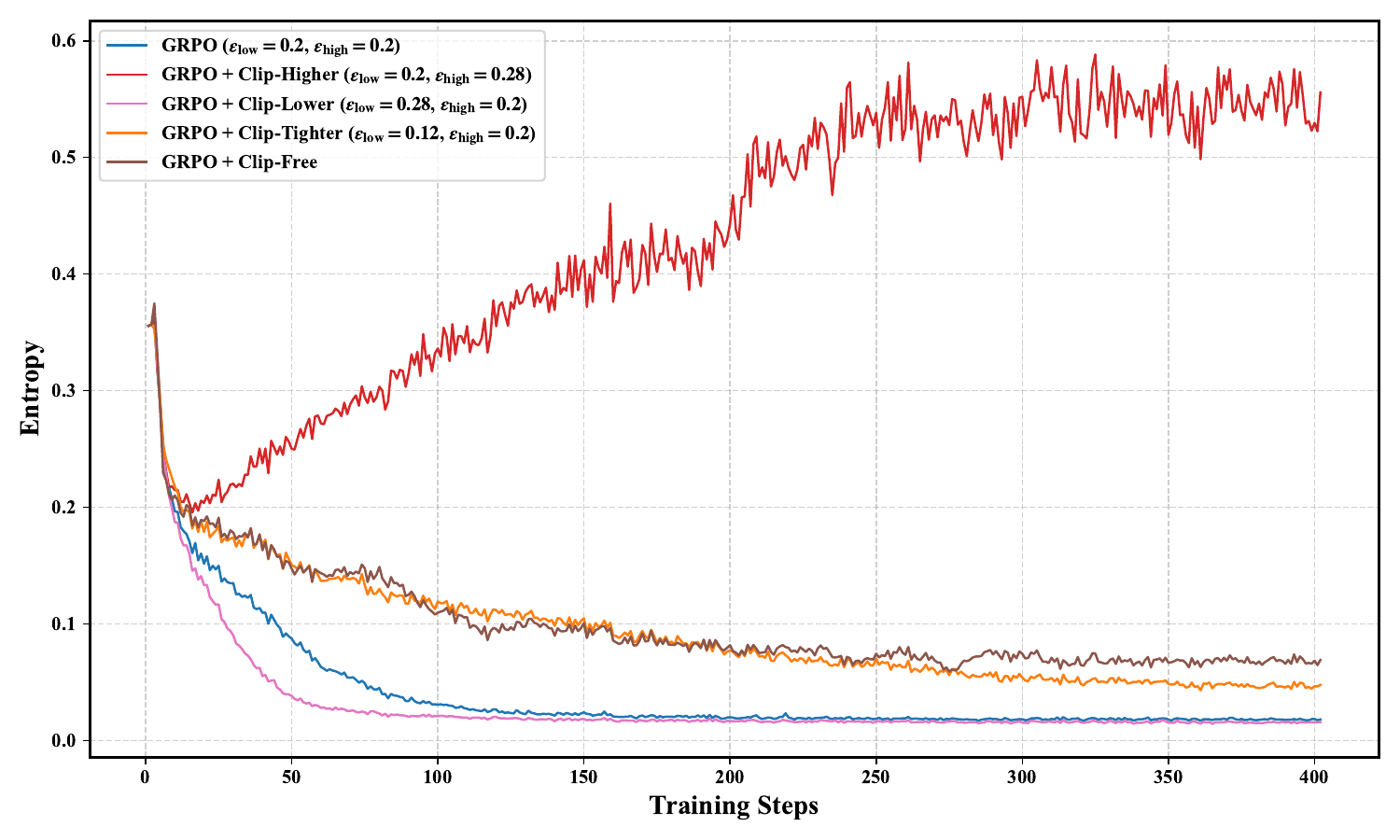}
    \caption{Evolution of LLM entropy during RLVR training under varying lower and upper clipping thresholds.}
    \label{fig:entropy_plot_clipping}
\end{figure}

The entropy dynamics of LLMs trained with GRPO under various clipping thresholds are illustrated in Figure~\ref{fig:entropy_plot_clipping}. As shown, \textbf{Clip-Higher} effectively prevents entropy collapse and even increases entropy during training. In contrast, the other clipping variants (\textbf{Clip-Lower}, \textbf{Clip-Tighter}, and \textbf{Clip-Free}) induce varying degrees of entropy collapse. Among them, \textbf{Clip-Lower} results in the most pronounced collapse, whereas \textbf{Clip-Tighter} mitigates entropy collapse and maintains higher entropy than the default clipping configuration.

These observations align with theoretical expectations. \textbf{Clip-Higher} expands the upper clipping bound, allowing low-probability tokens with positive advantages to avoid being clipped. Consequently, these tokens can increase their probabilities during training, mitigating the over-concentration of the probability distribution across the vocabulary and preventing entropy collapse. Conversely, \textbf{Clip-Lower} decreases the lower clipping bound, enabling low-probability tokens with negative advantages are excluded from clipping. This allows their probabilities to decrease further, thereby exacerbating the over-concentration of the distribution and intensifying entropy collapse. In comparison, \textbf{Clip-Tighter} raises the lower clipping bound, making low-probability tokens with negative advantages more likely to be clipped. As a result, their probabilities are preserved during training, which helps alleviate entropy collapse.

Notably, Figure~\ref{fig:entropy_plot_clipping} shows that \textbf{Clip-Free} achieves the highest entropy among all clipping variants, except for \textbf{Clip-Higher}. Moreover, as presented in Table~\ref{tab:all_results}, on in-domain test sets, \textbf{Clip-Free} yields the second-highest average Avg@64 and Pass@64 scores among all clipping variants. On out-of-domain test sets, \textbf{Clip-Free} similarly achieves the second-highest Avg@64 and the highest average Pass@64. These results indicate that, when the number of off-policy updates is small, removing clipping from the GRPO objective does not compromise the stability of GRPO training.

\subsection{Off-Policy Updates}
\label{subsec:off_policy_updates}
\begin{findings}
With the clipping hyperparameters held constant, increasing the number of off-policy updates amplifies changes in LLM entropy and enables the models to achieve higher rewards on the training set; however, the corresponding performance improvements on the test set are considerably less pronounced.
\end{findings}
In GRPO, a batch of prompts is sampled for rollout, advantage estimation, and log-probability computation using the rollout LLM. This batch is then divided into several mini-batches, with model parameters updated once per mini-batch. Since the parameters change after the first update, the data in the remaining mini-batches can be regarded as off-policy data with respect to the updated policy.

\begin{figure}[t]
    \centering
    \includegraphics[width=\linewidth]{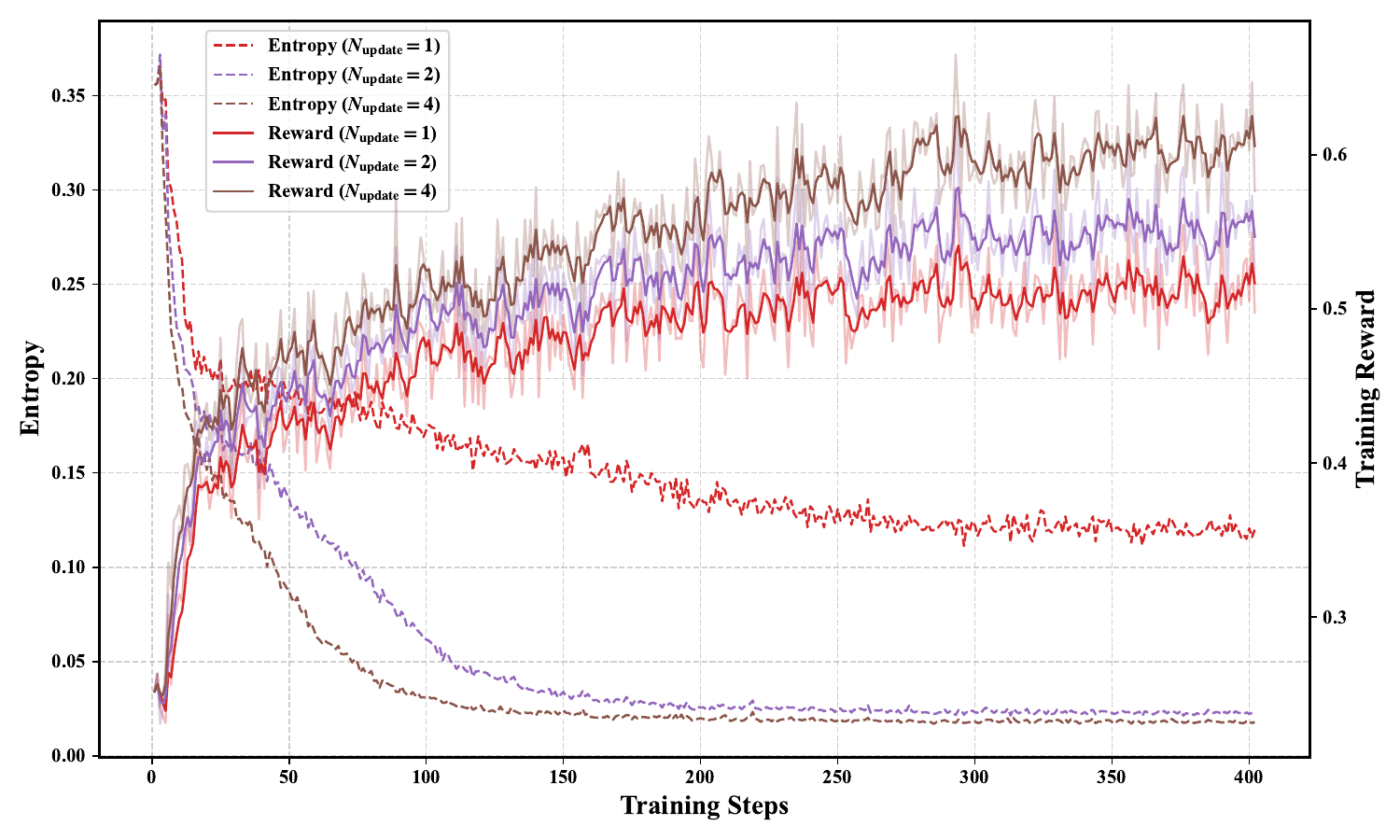}
    \caption{Evolution of entropy and training rewards under the default clipping hyperparameter setting.}
    \label{fig:entropy_train_reward}
\end{figure}

To study the effect of off-policy data on LLM entropy and reward during training, we conducted experiments under two clipping configurations: (1) the default setting with $\varepsilon_{\text{low}}=\varepsilon_{\text{high}}=0.2$, and (2) the Clip-Higher setting with $\varepsilon_{\text{low}}=0.2$ and $\varepsilon_{\text{high}}=0.28$. For each configuration, we varied the number of parameter updates per batch, $N_{\text{update}} \in \{1,2,4\}$, while keeping other hyperparameters fixed.

Figures~\ref{fig:entropy_train_reward} and~\ref{fig:entropy_train_reward_2} (Appendix~\ref{subsec:appendix_off_policy_updates}) depict the evolution of entropy and training reward under the default and Clip-Higher clipping settings, respectively. With clipping hyperparameters fixed, increasing the number of off-policy updates amplifies entropy changes. Under the default setting, where entropy decreases with few updates, additional updates accelerate this decline (Figure~\ref{fig:entropy_train_reward}). Conversely, under the Clip-Higher setting, where entropy increases with few updates, more off-policy updates lead to a faster entropy increase (Figure~\ref{fig:entropy_train_reward_2}). While more off-policy updates improve training rewards, Table~\ref{tab:all_results} shows that the corresponding gains in average Avg@64 on in-domain test sets remain below 1\%. Moreover, the average Pass@64 score decreases on both in-domain and out-of-domain benchmarks as the number of off-policy updates increases, suggesting that excessive off-policy updates may lead to overfitting.

\subsection{Training Data Diversity}
\label{subsec:training_data_diversity}

\begin{findings}
Lower data diversity intensifies entropy collapse during RLVR training. Moreover, training data size is not the only factor determining the performance of LLMs trained with RLVR, as an LLM trained on \textasciitilde
600 samples perform comparably to one trained on \textasciitilde
17k samples.
\end{findings}
To examine how training data diversity affects LLM entropy dynamics, we train models on datasets with identical sample sizes but varying diversity. Specifically, we construct training data subsets using K-means clustering and random sampling, with both methods selecting the same number of samples. Subsets produced by K-means clustering are expected to be less diverse than those obtained via random sampling, and overall diversity is expected to decline as dataset size decreases. Details on subset construction and entropy computation are provided in Appendix~\ref{subsec:appendix_training_data_diversity}.

Table~\ref{tab:data_diversity} in Appendix~\ref{subsec:appendix_training_data_diversity} summarizes the performance of LLMs trained on the constructed subsets and the full training dataset. As shown, entropy decreases as the size of the training data is reduced. Moreover, LLMs trained on subsets constructed via K-means clustering consistently exhibit lower entropy than those trained on randomly sampled subsets, except in the case of models trained on 5,031 samples. These results further suggest that training data diversity plays a critical role in entropy dynamics, with entropy tending to decline as data diversity diminishes. Notably, despite substantial reductions in training data, LLMs trained on substantially smaller subsets (e.g., the subset with 616 samples constructed via K-means clustering) can still achieve performance comparable to those trained on the full dataset. This finding indicates that data scale alone does not determine model performance, consistent with prior studies \citep{DBLP:journals/corr/abs-2502-11886,DBLP:journals/corr/abs-2502-03387,DBLP:journals/corr/abs-2501-19393}.

\section{How Can Entropy Be Effectively Regulated to Improve the Performance of LLMs?}
\label{sec:RQ_3}

To investigate how to regulate the entropy of LLMs for improved performance, we first present a theoretical analysis of how tokens with positive and negative advantages influence entropy dynamics during RLVR training (\S\ref{subsec:theoretical_analysis}). We then empirically validate the conclusions derived from this analysis (\S\ref{subsec:empirical_analysis}). Finally, building on these theoretical insights, we propose a \textbf{Positive-Advantage Reweighting} approach to effectively control the entropy of LLMs during RLVR training (\S\ref{subsec:positive-advantage_reweighting}).

\subsection{Theoretical Analysis}
\label{subsec:theoretical_analysis}

To effectively regulate entropy, we analyze which tokens drive entropy changes during training. Under GRPO, tokens can be grouped by advantage into those with positive, negative, or zero advantage. Since tokens with zero advantage do not contribute to the gradient, we exclude them from the analysis. Intuitively, high-probability tokens are more likely to be sampled during decoding and thus tend to dominate model responses. When such tokens also have positive advantages, their probabilities are further amplified after parameter updates, resulting in a more concentrated probability distribution and entropy collapse. We therefore hypothesize that entropy collapse in LLMs is primarily driven by tokens with positive advantages, which we further justify theoretically.

Let $z_v$ denote the logit of token $v$ produced by the LLM, and $r_{t}\left( \theta \right)$ denote the importance ratio $\frac{\pi_{\theta}\left( \bm{y}_{t} \vert \bm{x}, \bm{y}_{<t} \right)}{\pi_{\theta_\text{old}}\left( \bm{y}_{t} \vert \bm{x}, \bm{y}_{<t} \right)}$. When token $v$ \textbf{is not} sampled at step $t$, the gradient of the GRPO optimization objective with respect to $z_v$ can be approximated as follows:
\begin{equation}
\small
\frac{\partial \mathcal{J} \left(\theta \right)}{\partial z_v} = 
\begin{cases}
- r_{t} \left( \theta \right) \pi_{\theta} \left(v \mid \bm{x}, \bm{y}_{<t} \right) \hat{A}_{t}, \\
\quad \quad \;\;\; \text{if } \hat{A}_{t} > 0 \text{ and } r_{t} \left( \theta \right) < 1 + \varepsilon_{\text{high}} \\
0, \\
\quad \quad \;\;\; \text{if } \hat{A}_{t} > 0 \text{ and } r_{t} \left( \theta \right) > 1 + \varepsilon_{\text{high}} \\
- r_{t} \left( \theta \right) \pi_{\theta} \left(v \mid \bm{x}, \bm{y}_{<t} \right) \hat{A}_{t}, \\
\quad \quad \;\;\; \text{if } \hat{A}_{t} < 0 \text{ and } r_{t} \left( \theta \right) > 1 - \varepsilon_{\text{low}} \\
0,\\
\quad \quad \;\;\; \text{if } \hat{A}_{t} < 0 \text{ and } r_{t} \left( \theta \right) < 1 - \varepsilon_{\text{low}}
\end{cases}
\label{eq:logits_gradient_result_1}
\end{equation}

Similarly, when token $v$ \textbf{is} sampled at step $t$, the gradient can be approximated as follows:
\begin{equation}
\small
\frac{\partial \mathcal{J} \left(\theta \right)}{\partial z_v} = 
\begin{cases}
r_{t} \left( \theta \right) \left(1 - \pi_{\theta} \left(v \mid \bm{x}, \bm{y}_{<t} \right) \right) \hat{A}_{t}, \\
\quad \quad \;\;\; \text{if } \hat{A}_{t} > 0 \text{ and } r_{t} \left( \theta \right) < 1 + \varepsilon_{\text{high}} \\
0, \\
\quad \quad \;\;\; \text{if } \hat{A}_{t} > 0 \text{ and } r_{t} \left( \theta \right) > 1 + \varepsilon_{\text{high}} \\
r_{t} \left( \theta \right) \left(1 - \pi_{\theta} \left(v \mid \bm{x}, \bm{y}_{<t} \right) \right) \hat{A}_{t}, \\
\quad \quad \;\;\; \text{if } \hat{A}_{t} < 0 \text{ and } r_{t} \left( \theta \right) > 1 - \varepsilon_{\text{low}} \\
0, \\
\quad \quad \;\;\; \text{if } \hat{A}_{t} < 0 \text{ and } r_{t} \left( \theta \right) < 1 - \varepsilon_{\text{low}}
\end{cases}
\label{eq:logits_gradient_result_2}
\end{equation}

Detailed derivations are provided in Appendix~\ref{subsec:appendix_theoretical_analysis}. As shown in Eq.~(\ref{eq:logits_gradient_result_1}), because RLVR performs gradient ascent to maximize the objective, when token $v$ \textbf{is not} sampled at step $t$, a positive advantage decreases its probability, whereas a negative advantage increases it. Similarly, as shown in Eq.~(\ref{eq:logits_gradient_result_2}), when token $v$ \textbf{is} sampled at step $t$, a positive advantage increases its probability, while a negative advantage decreases it.

Taken together, a positive advantage leads to updates that increase the probabilities of sampled tokens while decreasing those of unsampled ones. Since high-probability tokens are more likely to be sampled, this mechanism further amplifies their probabilities while suppressing those of low-probability, unsampled tokens, thereby concentrating the probability mass and exacerbating entropy collapse. Conversely, when the advantage is negative, the update decreases the probabilities of sampled tokens and increases those of unsampled ones. In this case, because high-probability tokens are still more likely to be sampled, the update tends to reduce the probabilities of these high-probability sampled tokens while increasing those of low-probability, unsampled tokens. This effect counteracts over-concentration of the distribution and thus mitigates entropy collapse.

Furthermore, when the importance sampling ratio in the GRPO objective is clipped, the gradient with respect to $z_v$ becomes zero. This implies that clipping modulates the relative contributions of gradients from tokens with positive versus negative advantages, and thus also influences entropy, consistent with the empirical results in Section~\ref{subsec:clipping_threshold}.

\subsection{Empirical Analysis}
\label{subsec:empirical_analysis}

To empirically validate this hypothesis, we conducted two comparative experiments in which LLMs were trained exclusively on tokens with either non-negative advantages ($\text{Adv} \ge 0$) or non-positive advantages ($\text{Adv} \le 0$). We compared these settings with the following baselines: (1) \textbf{Adaptive Entropy Regularization}, (2) \textbf{Clip-Cov}, (3) \textbf{KL-Cov}, (4) \textbf{Entropy-Adv}, and (5) \textbf{Rand-Pos-Clip}. Details of these baselines and experimental settings are provided in Appendix~\ref{subsec:appendix_empirical_analysis}.

\begin{table*}[t]
  \centering
  \resizebox{\textwidth}{!}{
    \begin{tabular}{l|l|l|l|l|l|c|l|>{\columncolor{gray!15}}c|>{\columncolor{gray!15}}c|>{\columncolor{gray!15}}c}
    \toprule
          \multicolumn{1}{c|}{\textbf{Model}} & \multicolumn{1}{c|}{\textbf{AIME 2024}} & \multicolumn{1}{c|}{\textbf{AIME 2025}} & \multicolumn{1}{c|}{\textbf{MATH500}} & \multicolumn{1}{c|}{\textbf{AMC 2023}} & \multicolumn{1}{c|}{\textbf{Minerva}} & \multicolumn{1}{c|}{\textbf{LiveCodeBench}} & \multicolumn{1}{c|}{\textbf{ IF-Eval}} & \textbf{Average (ID)} & \textbf{Average (OOD)} & \textbf{Entropy} \\
    \midrule
    Qwen2.5-Math-7B & 10.00 / 60.00 & 3.80 / 33.33 & 43.76 / 95.60 & 30.04 / 92.50 & 14.41 / 60.29 & 3.62 / 30.15 & 22.67 / 80.46 & 20.40 / 68.35 & 13.15 / 55.30 & N/A \\
    \arrayrulecolor{black}\midrule
    + GRPO ($N_{\text{update}}=1$) & 28.75 / 63.33 & 14.69 / 50.00 & 78.14 / 96.80 & 64.38 / 97.50 & 34.64 / 64.34 & 7.85 / 33.46 & 30.17 / 72.90 & 44.12 / 74.39 & 19.01 / 53.18 & 0.11838 \\
    \arrayrulecolor{lightgray}\midrule
    + GRPO ($N_{\text{update}}=2$) & 29.58 / 70.00 & 16.98 / 46.67 & 76.56 / 94.40 & 67.42 / 92.50 & 33.28 / 62.50 & 10.66 / 34.56 & 28.72 / 71.10 & 44.76 / 73.21 & 19.69 / 52.83 & 0.02286 \\
    \midrule
    + DAPO ($N_{\text{update}}=4$) & 33.96 / 66.67 & 16.61 / 50.00 & 71.13 / 93.60 & 69.49 / 97.50 & 28.62 / 58.46 & 7.43 / 34.56 & 31.89 / 65.35 & 43.96 / 73.24 & 19.66 / 49.95 & 0.47900 \\
    \midrule
    + GRPO ($N_{\text{update}}=4$) & 31.41 / 63.33 & 14.90 / 56.67 & 72.09 / 90.80 & 75.43 / 90.00 & 31.14 / 55.51 & 12.37 / 34.56 & 29.83 / 70.38 & 44.99 / 71.26 & 21.10 / 52.47 & 0.01789 \\
    \arrayrulecolor{black}\midrule
    \: \: + Clip-Higher & 33.33 / 60.00 & 15.94 / 53.33 & 72.35 / 94.20 & 67.62 / 97.50 & 30.57 / 63.97 & 5.88 / 32.35 & 31.35 / 66.19 & 43.96 / 73.80 & 18.62 / 49.27 & 0.53910 \\
    \arrayrulecolor{lightgray}\midrule
    \: \: + Clip-Lower & 27.76 / 56.67 & 15.31 / 50.00 & 71.61 / 89.20 & 74.73 / 87.50 & 30.07 / 55.51 & 11.43 / 31.99 & 28.10 / 66.91 & 43.90 / 67.78 & 19.76 / 49.45 & 0.01577 \\
    \midrule
    \: \: + Clip-Tighter & 32.19 / 63.33 & 16.09 / 43.33 & 67.59 / 90.40 & 69.18 / 95.00 & 26.03 / 54.78 & 8.64 / 34.56 & 29.96 / 69.06 & 42.22 / 69.37 & 19.30 / 51.81 & 0.04681 \\
    \midrule
    \: \: + Clip-Free & 34.38 / 66.67 & 17.19 / 43.33 & 73.02 / 92.40 & 69.14 / 97.50 & 31.01 / 59.93 & 9.35 / 34.93 & 30.53 / 70.50 & 44.95 / 71.97 & 19.94 / 52.72 & 0.06745 \\
    \midrule
    \: \: + Ada-Ent-Reg ($\delta = 0.2$) & 32.92 / 66.67 & 16.30 / 50.00 & 69.05 / 90.40 & 69.34 / 92.50 & 27.63 / 61.03 & 7.16 / 33.09 & 31.26 / 68.94 & 43.05 / 72.12 & 19.21 / 51.02 & 0.19692 \\
    \midrule
    \: \: + Ada-Ent-Reg ($\delta = 0.3657$) & 33.96 / 66.67 & 18.65 / 50.00 & 73.98 / 92.80 & 68.52 / 97.50 & 31.66 / 61.76 & 6.31 / 32.35 & 29.66 / 69.78 & 45.35 / 73.75 & 17.98 / 51.07 & 0.30941 \\
    \midrule
    \: \: + Clip-Cov & 31.98 / 70.00 & 18.18 / 53.33 & 74.27 / 95.80 & 68.13 / 97.50 & 32.23 / 62.50 & 7.85 / 34.56 & 31.05 / 69.06 & 44.96 / 75.83 & 19.45 / 51.81 & 0.20899 \\
    \midrule
    \: \: + KL-Cov & 33.96 / 66.67 & 16.20 / 53.33 & 70.10 / 93.60 & 68.79 / 97.50 & 28.63 / 61.03 & 7.61 / 34.93 & 30.82 / 68.35 & 43.54 / 74.43 & 19.21 / 51.64 & 0.19695 \\
    \midrule
    \: \: + Entropy-Adv & 31.09 / 66.67 & 15.52 / 46.67 & 76.65 / 93.60 & 70.63 / 87.50 & 33.80 / 60.29 & 11.75 / 31.25 & 29.21 / 70.14 & 45.54 / 70.95 & 20.48 / 50.70 & 0.01669 \\
    \midrule
    \: \: + $\text{Adv} \leq 0$ & 29.79 / 63.33 & 11.04 / 46.67 & 76.55 / 96.00 & 63.40 / 97.50 & 32.73 / 63.97 & 3.64 / 26.47 & 33.70 / 62.11 & 42.70 / 73.49 & 18.67 / 44.29 & 0.88384 \\
    \midrule
    \: \: + $\text{Adv} \geq 0$ & 27.55 / 53.33 & 13.23 / 43.33 & 72.20 / 91.00 & 64.49 / 92.50 & 34.05 / 58.09 & 10.92 / 33.46 & 28.17 / 71.58 & 42.30 / 67.65 & 19.55 / 52.52 & 0.01460 \\
    \midrule
    \: \: + Rand-Pos-Clip & 34.27 / 66.67 & 16.93 / 46.67 & 73.21 / 93.80 & 68.13 / 97.50 & 31.86 / 61.03 & 8.84 / 34.19 & 30.74 / 69.54 & 44.88 / 73.13 & 19.79 / 51.87 & 0.05763 \\
    \midrule
    \: \: + Pos-Adv-Reweight (Stage-based) & 31.72 / 56.67 & 15.21 / 46.67 & 78.79 / 96.00 & 64.84 / 95.00 & 33.67 / 65.81 & 4.92 / 31.99 & 33.07 / 60.91 & 44.85 / 72.03 & 19.00 / 46.45 & 0.32983 \\
    \midrule
    \: \: + Pos-Adv-Reweight (Epoch-wise) & 32.34 / 66.67 & 17.45 / 43.33 & 75.75 / 95.40 & 66.17 / 95.00 & 33.51 / 60.29 & 6.93 / 33.46 & 31.63 / 66.43 & 45.05 / 72.14 & 19.28 / 49.94 & 0.05203 \\
    \midrule
    \: \: + Pos-Adv-Reweight (Entropy-guided) & 34.38 / 73.33 & 15.89 / 40.00 & 75.93 / 95.40 & 69.34 / 92.50 & 32.78 / 64.71 & 6.89 / 33.82 & 31.88 / 66.07 & 45.66 / 73.19 & 19.39 / 49.95 & 0.18746 \\
    \arrayrulecolor{black}\bottomrule
    \end{tabular}
  }
  \caption{Performance of Qwen2.5-Math-7B trained with GRPO and its variants. ``Ada-Ent-Reg'' denotes Adaptive Entropy Regularization. ``Average (ID)'' and ``Average (OOD)'' indicate the mean performance across in-domain and out-of-domain benchmarks, respectively. Results are presented as A / B, representing Avg@64 and Pass@64.}
  \label{tab:all_results}
\end{table*}

\begin{figure}[t]
    \centering
    \includegraphics[width=\linewidth]{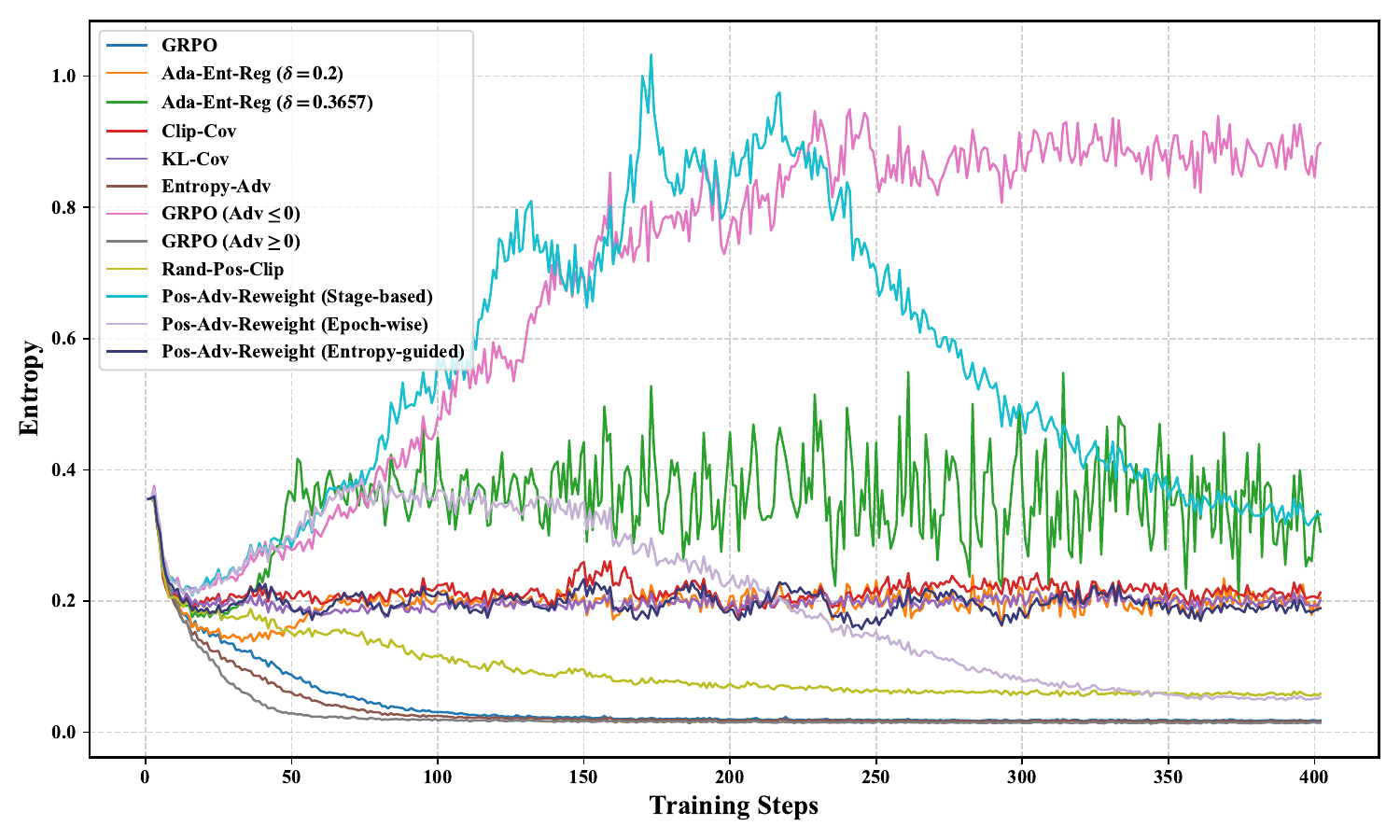}
    \caption{Evolution of the entropy of LLMs during RLVR training under different methods. ``Ada-Ent-Reg'' denotes Adaptive Entropy Regularization.}
    \label{fig:entropy_plot}
\end{figure}

Figure~\ref{fig:entropy_plot} shows the entropy evolution across various methods. As shown, \textbf{Clip-Cov}, \textbf{KL-Cov}, \textbf{Adaptive Entropy Regularization}, and \textbf{Rand-Pos-Clip} effectively alleviate entropy collapse, whereas \textbf{Ent-Adv} intensifies it. Furthermore, training LLMs exclusively on tokens with advantages $\ge 0$ leads to the most severe entropy collapse, while training on tokens with advantages $\le 0$ yields high entropy. These results empirically support our hypothesis that entropy collapse primarily stems from tokens with positive advantages, suggesting that adjusting the loss weights of tokens with advantages $\ge 0$ can regulate model entropy.

\subsection{Positive-Advantage Reweighting}
\label{subsec:positive-advantage_reweighting}

Building on this insight, we propose \textbf{Positive-Advantage Reweighting (Pos-Adv-Reweight)}, which controls entropy by dynamically reweighting the loss of tokens with positive advantages. Specifically, we introduce a hyperparameter $\lambda$ that determines the loss weights of such tokens. We consider three variants of \textbf{Pos-Adv-Reweight}:
\begin{itemize}[itemsep=0pt, leftmargin=*]
    \item \textbf{Pos-Adv-Reweight (Stage-based)} divides RLVR training into two equal stages. In the first stage, $\lambda=0$, so training uses only tokens with non-positive advantages. In the second stage, $\lambda$ increases linearly from 0 to 1, gradually incorporating tokens with positive advantages.
    \item \textbf{Pos-Adv-Reweight (Epoch-wise)} increases $\lambda$ linearly across epochs, from 0 in the first epoch to 1 in the final epoch. If the total number of epochs is $E$ and the current epoch is $e$, then $\lambda$ is defined as $\lambda = (e-1)/(E-1)$.
    \item \textbf{Pos-Adv-Reweight (Entropy-guided)} adaptively adjusts $\lambda$ based on the model entropy during training. Specifically, when the LLM entropy exceeds a predefined threshold $\delta$, $\lambda$ is increased by $\Delta$ to suppress entropy; otherwise, $\lambda$ is decreased by $\Delta$ to encourage higher entropy. In this way, the training process dynamically regulates the model entropy around the target threshold $\delta$. Formally, let $\delta$ denote the predefined entropy threshold, and let $\lambda_k$ represent the loss weight of tokens with positive advantages at training step $k$. To ensure comparability with Ada-Ent-Reg, we set $\delta = 0.2$, fix the step size $\Delta = 0.05$, and initialize $\lambda_0 = 0$. The update rule is given by:
\end{itemize}
\begin{equation}
\small
\lambda_{k+1} =
\begin{cases}
\text{clip}\left( \lambda_k - \Delta, 0, 1 \right), & \text{if } \mathcal{H}_k \left( \pi_{\theta} \right) < \delta \\
\text{clip}\left( \lambda_k + \Delta, 0, 1 \right), & \text{otherwise}
\end{cases}.
\label{eq:pos_adv_reweight}
\end{equation}

As shown in Figure~\ref{fig:entropy_plot}, both \textbf{Pos-Adv-Reweight (Stage-based)} and \textbf{Pos-Adv-Reweight (Epoch-wise)} induce an initial rise in entropy followed by a gradual decline, indicating that entropy decreases as the loss weights of tokens with positive advantages increase. Meanwhile, \textbf{Pos-Adv-Reweight (Entropy-guided)} effectively maintains the model entropy around the target value of 0.2. Moreover, \textbf{Rand-Pos-Clip}, which randomly sets the gradients of a small subset of tokens with positive advantages to zero, mitigates entropy collapse relative to the GRPO baseline. Collectively, these results demonstrate that dynamically adjusting the relative loss weights of tokens with positive and negative advantages effectively regulates entropy.

Table~\ref{tab:all_results} summarizes the performance of LLMs trained with RLVR under various entropy regularization approaches across multiple benchmarks. As shown in Table~\ref{tab:all_results}, although training exclusively on tokens with advantages $\le 0$ effectively mitigates entropy collapse, it results in lower average Avg@64 scores on both in-domain and out-of-domain benchmarks, as well as reduced average Pass@64 scores on out-of-domain benchmarks. In contrast, by dynamically adjusting the loss weights of tokens with non-negative advantages, both \textbf{Pos-Adv-Reweight (Stage-based)} and \textbf{Pos-Adv-Reweight (Epoch-wise)} further improve performance beyond the \textbf{$\text{Adv} \le 0$} setting and outperform the GRPO baseline on five of the seven benchmarks (AIME 2024, AIME 2025, MATH500, Minerva, and IF-Eval) in terms of Avg@64, while achieving average Avg@64 scores comparable to other entropy regularization methods.

Although \textbf{Clip-Higher} also alleviates entropy collapse, it does not explicitly regulate entropy. As a result, entropy may fluctuate unpredictably and drift without control. In contrast, \textbf{Pos-Adv-Reweight} explicitly regulates entropy toward a predefined target value, enabling precise and effective entropy control during training. Furthermore, all three variants of \textbf{Pos-Adv-Reweight}, namely \textbf{Pos-Adv-Reweight (Stage-based)}, \textbf{Pos-Adv-Reweight (Epoch-wise)}, and \textbf{Pos-Adv-Reweight (Entropy-guided)}, consistently outperform \textbf{Clip-Higher} in terms of average Avg@64 scores across both in-domain and out-of-domain benchmarks. Notably, \textbf{Pos-Adv-Reweight (Entropy-guided)} achieves higher average Avg@64 scores than \textbf{Clip-Higher} on six of the seven benchmarks (AIME 2024, MATH500, AMC 2023, Minerva, LiveCodeBench, and IF-Eval), and attains the best average Avg@64 scores among all entropy regularization approaches.

Overall, these results show that \textbf{Pos-Adv-Reweight} effectively mitigates entropy collapse while maintaining competitive performance. Moreover, despite its simplicity, \textbf{Rand-Pos-Clip}, which randomly zeros the gradients of a small subset of tokens with positive advantages, achieves average Avg@64 scores comparable to \textbf{Clip-Cov} on both in-domain and out-of-domain benchmarks, as well as comparable average Pass@64 scores on out-of-domain benchmarks. This suggests that adjusting the loss weights of tokens with positive advantages is an effective strategy for controlling model entropy while preserving strong performance.

To connect our study with theoretical analyses of LLM entropy dynamics during GRPO training \citep{zhihu2025entropy,DBLP:journals/corr/abs-2505-22617}, we visualize the covariance between token log-probabilities and advantages, with detailed results provided in Appendix~\ref{subsec:appendix_analysis_of_the_covariance}. Furthermore, we conduct experiments on Llama-3.1-8B-Instruct to demonstrate that our findings generalize beyond Qwen2.5-Math-7B. The corresponding results are reported in Appendix~\ref{sec:appendix_experiments_on_llama_3_1_8b_instruct}.

\section{Conclusion}
In this paper, we comprehensively investigate the entropy dynamics of LLMs trained with RLVR. Through extensive empirical analyses, we examine how entropy correlates with response diversity, model calibration, and performance across multiple benchmarks. Furthermore, we identify three factors influencing entropy dynamics: the clipping threshold, the number of off-policy updates, and training data diversity. Notably, we observe that an LLM trained on approximately 600 samples performs comparably to one trained on about 17k samples. Our theoretical and empirical analyses further reveal that entropy collapse primarily arises from tokens with positive advantages, and that entropy can be effectively regulated by adjusting the loss weights of tokens with positive advantages. Building on this insight, we propose \textbf{Positive-Advantage Reweighting (Pos-Adv-Reweight)}, a simple yet effective approach that dynamically adjusts the loss weights of positive-advantage tokens to control entropy, while maintaining competitive performance across benchmarks.

\section*{Acknowledgements}
The present research was supported by the National Key Research and Development Program of China (Grant No. 2024YFE0203000), the State Key Laboratory of Tibetan Intelligence (Grant No. 2025-ZJ-J08), the Postdoctoral Fellowship Program of CPSF (Grant No. GZC20251075), and the National Natural Science Foundation of China (Grant No. 62306213). We would like to thank the anonymous reviewers for their insightful comments.

\section*{Limitations}
A concurrent study introducing QwenLong-L1.5 \citep{shen2025qwenlongl15posttrainingrecipelongcontext} proposes AEPO to enhance the stability of RL training for LLMs in long-context settings and to improve their long-context reasoning performance. Specifically, AEPO dynamically sets the gradients of samples with negative advantages to zero based on the model entropy during training, which is equivalent to assigning zero loss weights to such samples or training exclusively on samples with non-negative advantages. This approach shares a similar core idea with our proposed \textbf{Positive-Advantage Reweighting}, suggesting that \textbf{Positive-Advantage Reweighting} has the potential to enhance both training stability and model performance for RL-trained LLMs beyond the mathematical domain. However, due to computational constraints, the RLVR experiments in our study were conducted exclusively on training data from the mathematical domain. This limitation may restrict the extent to which the proposed \textbf{Positive-Advantage Reweighting} fully demonstrates its effectiveness in stabilizing RLVR training and improving LLM performance in more dynamic environments, such as agentic RL \citep{DBLP:journals/corr/abs-2509-02547}. Nevertheless, we expect that \textbf{Positive-Advantage Reweighting} can effectively enhance both training stability and model performance for LLMs trained with RL beyond the mathematical domain. Furthermore, we hope that our empirical analysis of entropy dynamics in RLVR training provides valuable insights to the research community and motivates the development of more effective entropy regularization strategies in future work.

\bibliography{custom}

\clearpage

\appendix

\section{Preliminaries}
\subsection{Group Relative Policy Optimization (GRPO)}
\label{subsec:appendix_GRPO}

Let the prompt be $\bm{x}$ and the response generated by the LLM $\pi_{\theta}$ (parameterized by $\theta$) be $\bm{y}$. The reward for response $\bm{y}$ is denoted as $R(\bm{y})$. The RLVR objective is to maximize the expected reward of responses generated by $\pi_{\theta}$, formulated as:
\begin{equation}
\small
J\left( \theta \right) = \mathbb{E}_{\bm{y} \sim \pi_{\theta}\left(\cdot \vert \bm{x} \right)} R\left(\bm{y}\right).
\end{equation}

To optimize $\pi_{\theta}$ for maximizing the expected reward, GRPO \citep{DBLP:journals/corr/abs-2402-03300} adopts the surrogate objective of PPO \citep{DBLP:journals/corr/SchulmanWDRK17}, replacing critic-based advantage estimation with group-based reward normalization to reduce the memory and computational cost of training a critic. Instead of learning a state-value function, GRPO samples multiple responses per prompt and computes each response’s advantage by normalizing its reward with the group’s mean and standard deviation. Following DAPO, we employ a token-level loss to ensure equal token contribution across responses of varying lengths and remove the KL penalty term from the original GRPO objective. Formally, let $\pi_{\theta}$ sample $G$ responses $\{\bm{y}^{j}\}_{j=1}^{G}$ for each prompt $\bm{x}$, and let $\mathcal{D}$ denote the prompt dataset. The optimization objective of GRPO with token-level loss and without the KL penalty term is:
\begin{equation}
\small
\begin{aligned}
    \mathcal{J} \left( \theta \right) = \:&\mathbb{E}_{\bm{x} \sim \mathcal{D},\left\{ \bm{y}^{i} \right\}_{i=1}^{G} \sim \pi_{\theta_\text{old}}\left(\cdot \vert \bm{x}\right)} 
    \\
    & \Biggl[ \frac{1}{\sum_{i=1}^{G}\left| \bm{y}^i \right|} \sum_{i=1}^{G}\sum_{t=1}^{\left| \bm{y}^i \right|} \min \Bigl( r_{i,t}(\theta) \hat{A}_{i,t}, 
    \\
    & \: \text{clip} \left( r_{i,t} \left( \theta \right), 1 - \varepsilon_{\text{low}}, 1 + \varepsilon_{\text{high}} \right) \hat{A}_{i,t} \Bigr) \Biggr],
\end{aligned}
\end{equation}
where $r_{i,t} \left( \theta \right) = \frac{\pi_{\theta}\left( \bm{y}_{t}^{i} \vert \bm{x}, \bm{y}_{<t}^i \right)}{\pi_{\theta_\text{old}}\left( \bm{y}_{t}^{i} \vert \bm{x}, \bm{y}_{<t}^i \right)}$, the advantage is given by $\hat{A}_{i,t} = \frac{ R\left( \bm{y}^{i} \right) - \text{mean}\left( \left\{ R\left( \bm{y}^{j} \right) \right\}_{j=1}^{G} \right)}{\text{std}\left( \left\{ R\left(\bm{y}^{j} \right) \right\}_{j=1}^{G} \right)}$, and $\varepsilon_{\text{low}}$ and $\varepsilon_{\text{high}}$ are clipping hyperparameters controlling the lower and upper bounds, respectively.

\section{Experimental Setup}
\label{sec:appendix_experimental_setup}
We trained Qwen2.5-Math-7B \citep{DBLP:journals/corr/abs-2409-12122} with GRPO using the veRL framework \citep{DBLP:conf/eurosys/ShengZYWZZPL025} on the DAPO-Math-17K dataset \citep{DBLP:journals/corr/abs-2503-14476}. Training employed a rollout batch size of 256, generating 16 responses per prompt. The AdamW optimizer was applied with a cosine learning rate schedule and a peak rate of $1 \times 10^{-6}$. During rollouts, decoding parameters were fixed as: top-$p=1.0$, temperature $=1.0$, and a maximum generation length of 4096 tokens.

We evaluated both in-domain and out-of-domain performance. The in-domain benchmarks included AIME 2024/2025 \citep{aime2024,aime2025}, MATH500 \citep{DBLP:conf/nips/HendrycksBKABTS21}, AMC 2023 \citep{amc2023}, and Minerva Math \citep{DBLP:conf/nips/LewkowyczADDMRS22}, while out-of-domain evaluation covered LiveCodeBench \citep{DBLP:conf/iclr/JainHGLYZWSSS25} for coding and IF-Eval \citep{DBLP:journals/corr/abs-2311-07911} for instruction following. For each question, we sampled 64 responses and reported both Avg@64 and Pass@64. All evaluations used a decoding temperature of 1.0 and top-$p$ of 1.0. To mitigate potential data contamination in the Qwen2.5 series of LLMs \citep{DBLP:journals/corr/abs-2507-10532}, AIME 2025 served as the validation set, and the best checkpoint was evaluated on all other benchmarks.

\section{How Does the Entropy of LLMs Trained with RLVR Correlate with Their Performance?}
\label{sec:appendix_RQ_1}

\subsection{Entropy and Response Diversity}
\label{subsec:appendix_entropy_and_response_diversity}

\begin{findings}
The diversity of responses generated by LLMs is strongly and positively correlated with their entropy during training.
\end{findings}

To examine the relationship between response diversity and model entropy, we control LLM entropy using vanilla entropy regularization by varying the regularization coefficient. Experiments are conducted with coefficients $\{0.001, 0.002\}$, without entropy regularization, and with a negative coefficient $-0.001$ that explicitly promotes entropy minimization. Response diversity is evaluated using the N-gram Diversity \citep{DBLP:conf/naacl/LiGBGD16} and SelfBLEU \citep{DBLP:conf/sigir/ZhuLZGZWY18} metrics. 

The N-gram Diversity metric \citep{DBLP:conf/naacl/LiGBGD16} measures the proportion of unique n-grams relative to the total number of n-grams in the generated responses. Let $U_i$ denote the number of unique n-grams and $C_i$ the total number of n-grams of order $i$. The metric is defined as follows:
\begin{equation}
\small
\text{N-gram Diversity} = \prod_{i=1}^{N} \frac{U_i}{C_i}.
\end{equation}
In our experiments, we set $N=5$ to assess response diversity.

SelfBLEU \citep{DBLP:conf/sigir/ZhuLZGZWY18} provides a complementary measure of diversity. For each response, the BLEU score is computed by treating it as the hypothesis and all other responses as references. The resulting BLEU scores, averaged over 1- to 4-gram, are then averaged across all responses to obtain the final SelfBLEU score.

\begin{figure}[t]
    \centering
    \includegraphics[width=\linewidth]{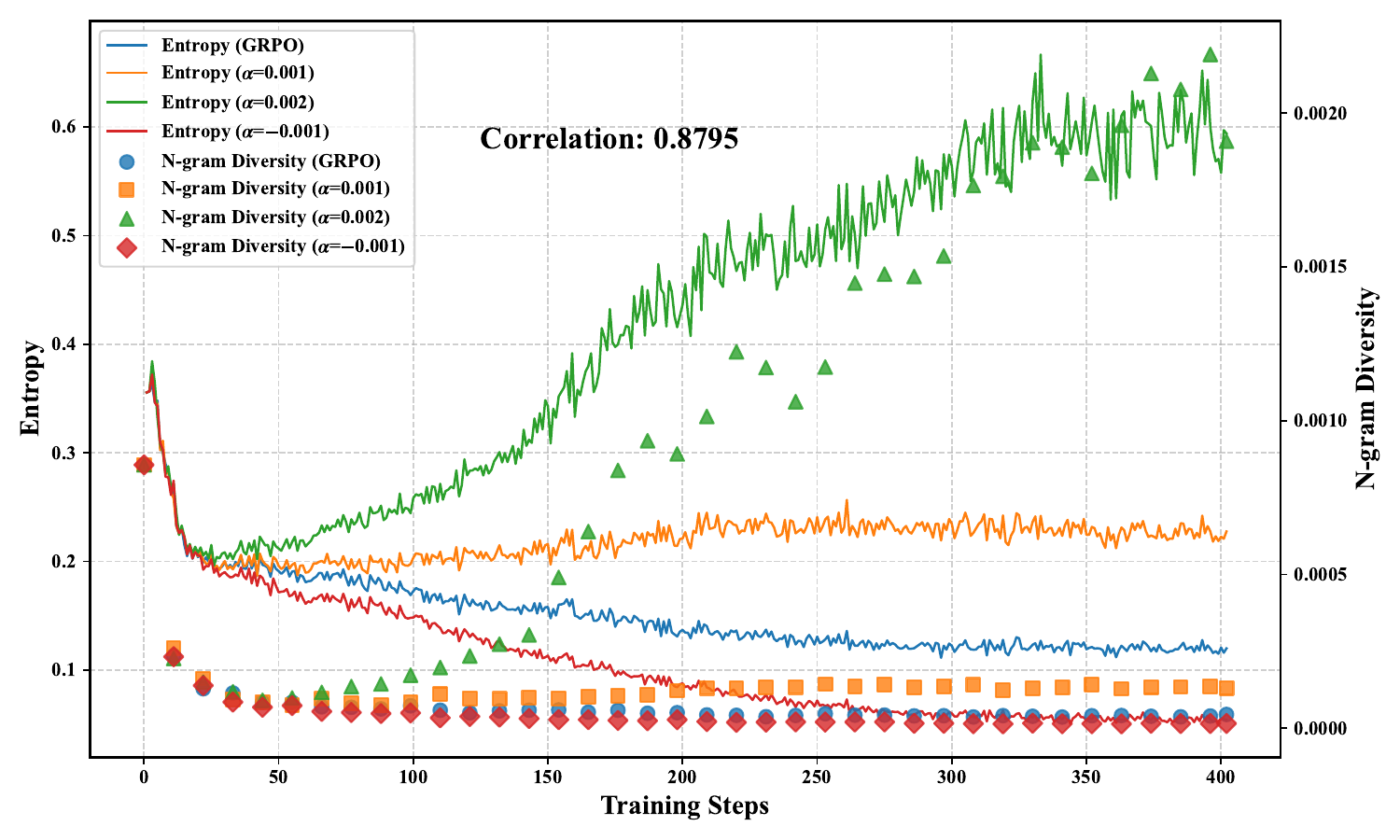}
    \caption{Evolution of entropy (solid lines) and N-gram diversity (markers) across training steps under different entropy regularization settings for Qwen2.5-Math-7B trained with GRPO.}
    \label{fig:entropy_ngram_diversity_plot_aime24}
\end{figure}

\begin{figure}[t]
    \centering
    \includegraphics[width=\linewidth]{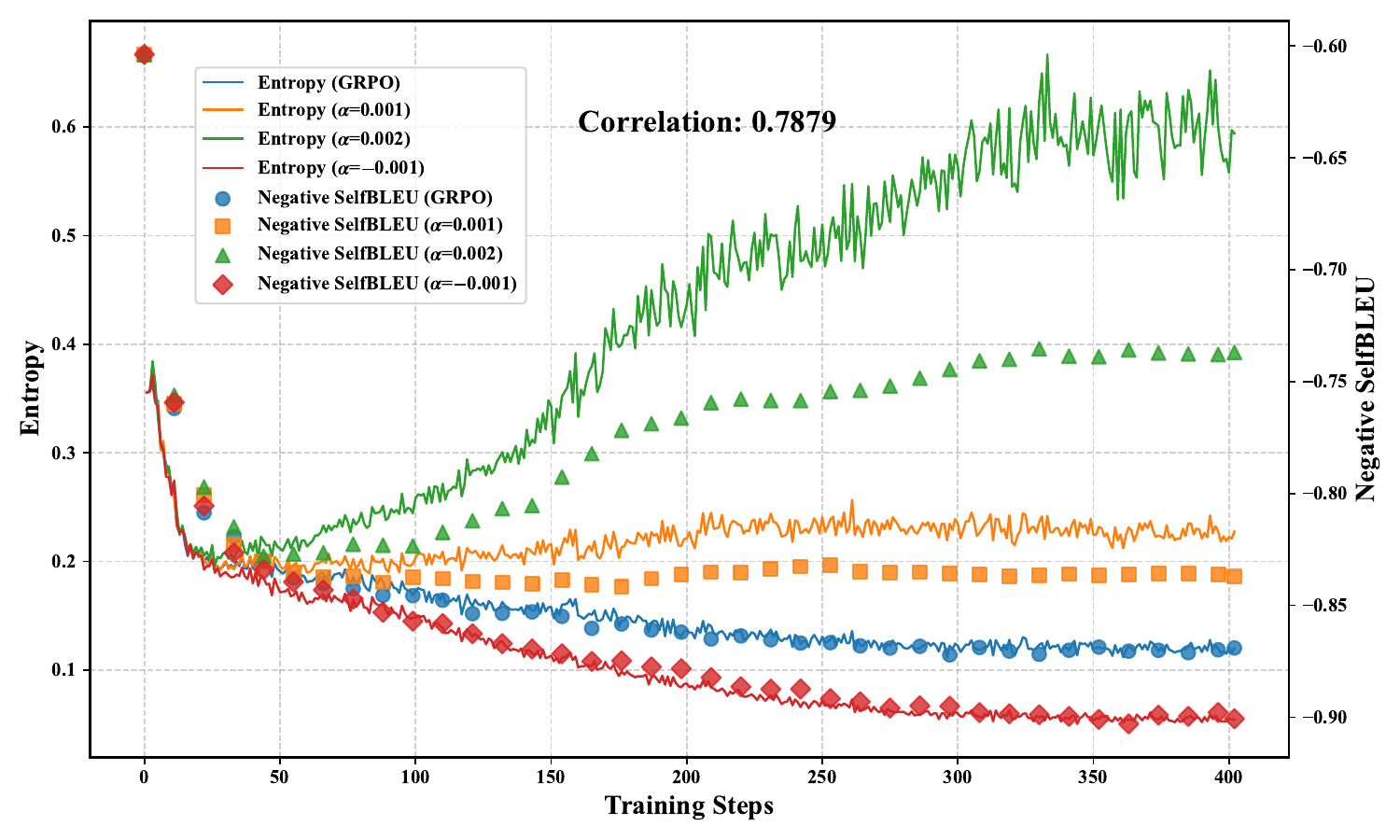}
    \caption{Evolution of entropy (solid lines) and negative SelfBLEU (markers) across training steps under different entropy regularization settings for Qwen2.5-Math-7B trained with GRPO.}
    \label{fig:entropy_selfbleu_plot_aime24}
\end{figure}

Figures~\ref{fig:entropy_ngram_diversity_plot_aime24} and~\ref{fig:entropy_selfbleu_plot_aime24} illustrate the entropy dynamics during RLVR training, together with the diversity of the generated responses to AIME 2024 prompts, as measured by N-gram Diversity and SelfBLEU, respectively. As shown in Figure~\ref{fig:entropy_ngram_diversity_plot_aime24}, LLMs trained with GRPO exhibit entropy collapse during training, which is alleviated by vanilla entropy regularization with $\alpha=0.001$, while a larger coefficient ($\alpha=0.002$) causes entropy to increase continuously in later stages. Conversely, applying a negative coefficient ($\alpha=-0.001$) further exacerbates entropy collapse. Notably, N-gram Diversity follows a similar trajectory to entropy, with an average Spearman correlation of 0.8795 across the four training settings, indicating a strong positive correlation between response diversity and model entropy during training.

\subsection{Entropy Dynamics on Prompts}
\label{subsec:appendix_entropy_dynamics_on_prompts}
\begin{figure}[t]
    \centering
    \includegraphics[width=\linewidth]{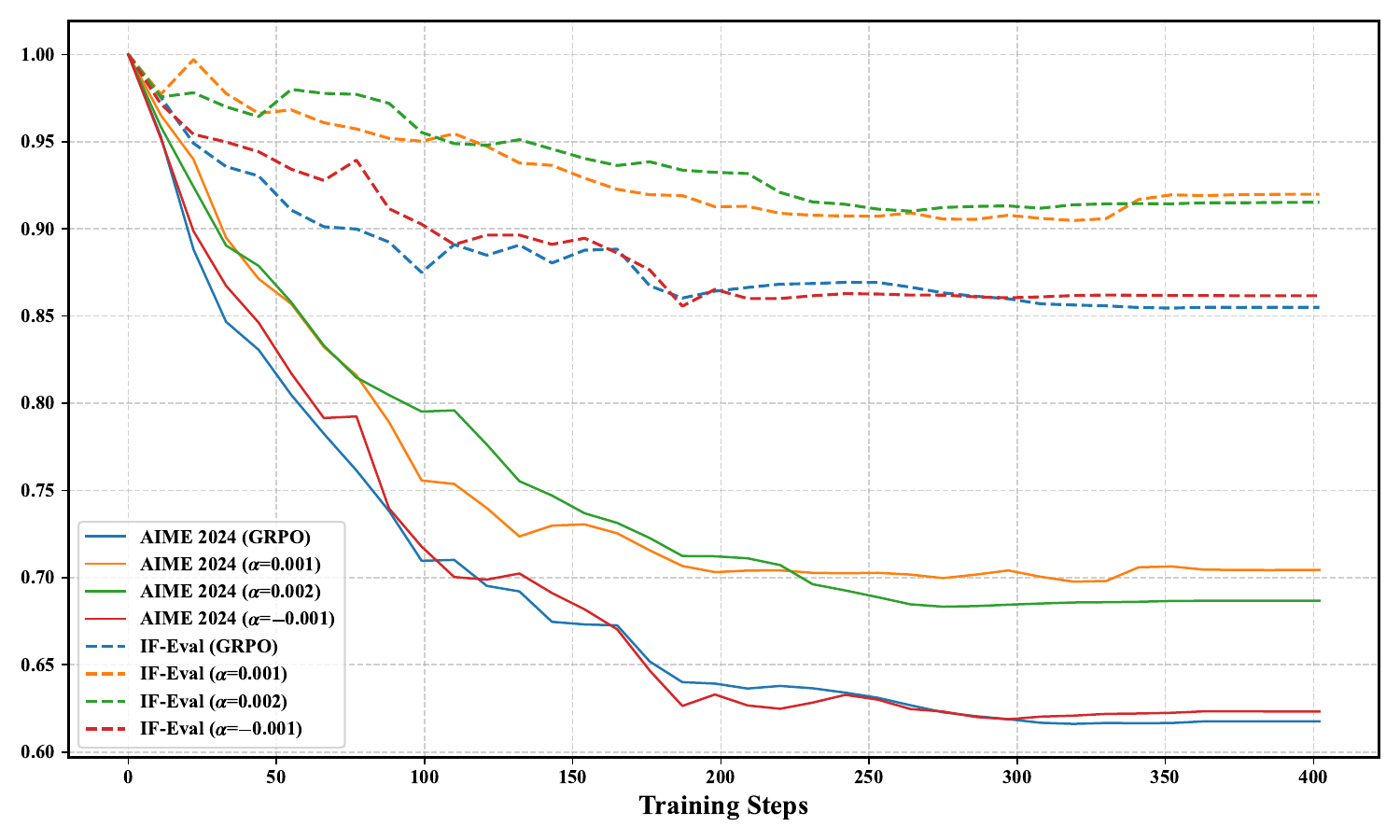}
    \caption{Ratio of the prompt entropy of Qwen2.5-Math-7B at various training steps relative to its initial entropy.}
    \label{fig:entropy_prompts_aime24}
\end{figure}

\begin{findings}
During RLVR training, the entropy of LLMs decreases for both in-domain and out-of-domain prompts, with a more substantial reduction observed for in-domain prompts.
\end{findings}
In addition to examining the entropy dynamics of responses generated by LLMs trained with RLVR, we also analyze the entropy dynamics of LLMs on prompts during RLVR training. Specifically, we compute the ratio between the entropy of LLMs on prompts at different training steps and their entropy on the same prompts before training, as shown in Figure~\ref{fig:entropy_prompts_aime24}. The entropy of LLMs on prompts decreases over training and eventually stabilizes. This reduction is more pronounced for prompts from the in-domain AIME 2024 benchmark than for those from the out-of-domain IF-Eval benchmark. Moreover, similar to preventing entropy collapse in generated responses, applying a positive entropy regularization coefficient helps maintain the entropy of LLMs on prompts.

\subsection{Prompt Entropy and Accuracy}
\label{subsec:appendix_prompt_entropy_and_accuracy}

\begin{findings}
The entropy of LLMs on prompts shows only a weak correlation with their accuracy.
\end{findings}

\begin{figure}[t]
    \centering
    \includegraphics[width=\linewidth]{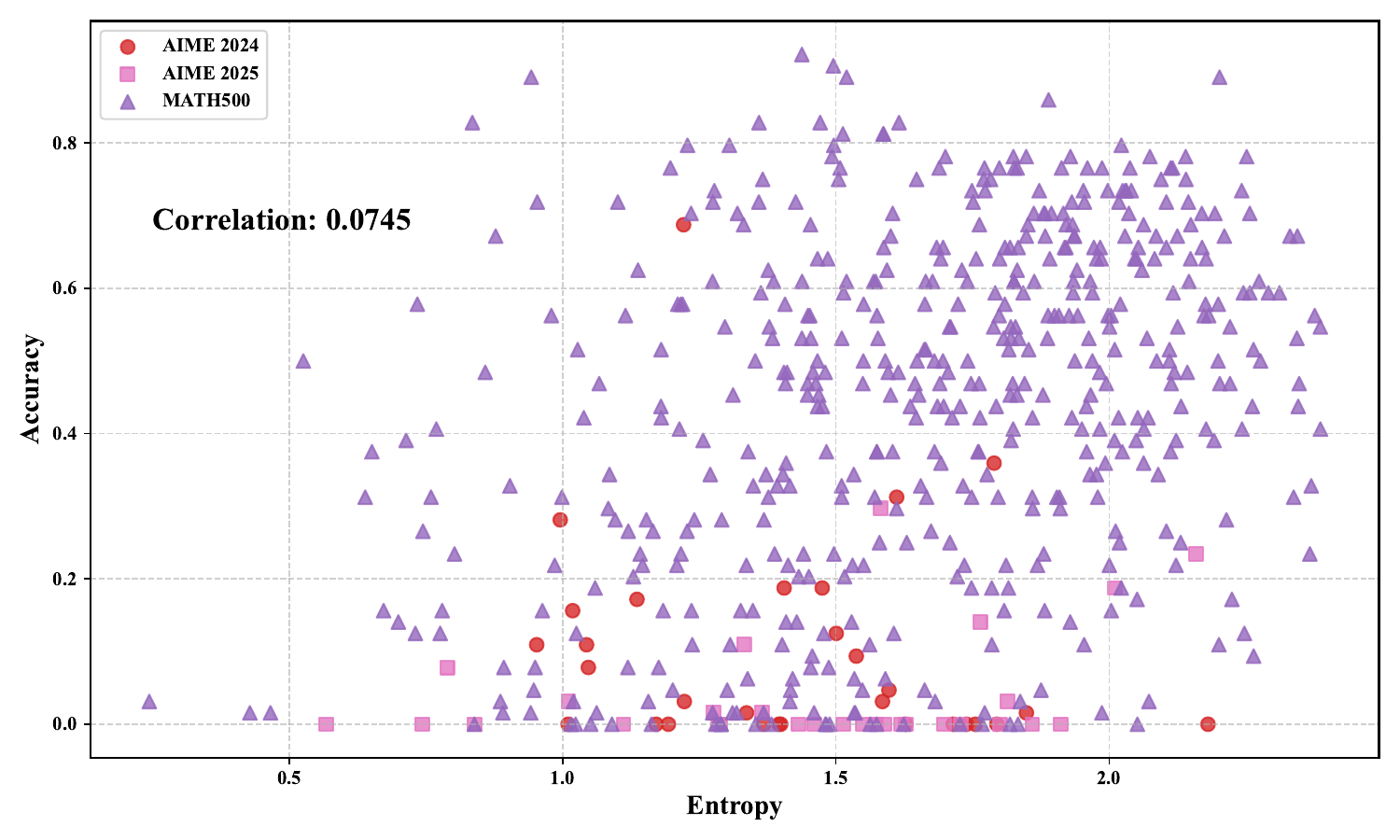}
    \caption{Scatter plot of prompt entropy versus accuracy across AIME 2024, AIME 2025, and MATH500.}
    \label{fig:entropy_acc_prompts}
\end{figure}

To examine the correlation between prompt entropy and response accuracy in LLMs, we quantify the accuracy of each prompt as the proportion of correct responses among 64 generated responses. Figure~\ref{fig:entropy_acc_prompts} shows scatter plots of prompt entropy versus accuracy on AIME 2024, AIME 2025, and MATH500 for Qwen2.5-Math-7B. The results show only a weak correlation, with an average Spearman's rank coefficient of 0.0745 across the three benchmarks. We further compute the Spearman correlation at different training steps. As shown in Figure~\ref{fig:entropy_acc_prompts_corrcoef_plot}, although the coefficient fluctuates during training, it remains consistently small, further confirming the weak correlation between prompt entropy and response accuracy.

\subsection{Correlations Between Entropy and Model Performance}
\label{subsec:appendix_correlations_between_entropy_and_model_performance}

\begin{figure}[t]
    \centering
    \includegraphics[width=\linewidth]{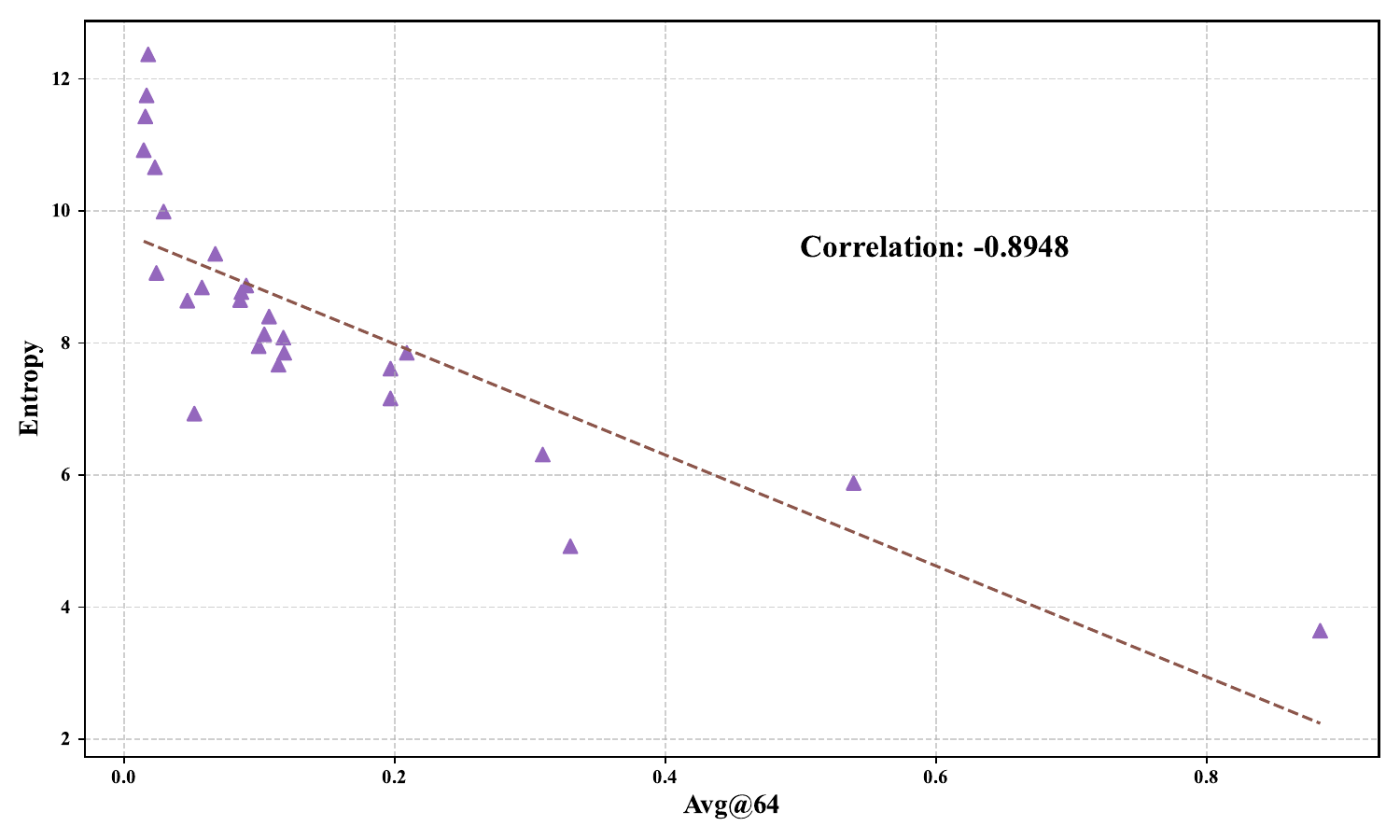}
    \caption{Scatter plot illustrating the relationship between LLM entropy during training and Avg@64 scores on LiveCodeBench. The brown dashed line represents the least-squares regression fit to the data points.}
    \label{fig:entropy_code_acc_corrcoef_plot_2}
\end{figure}

The scatter plot depicting the relationship between LLM training entropy and model performance on LiveCodeBench, as measured by Avg@64, is presented in Figure~\ref{fig:entropy_code_acc_corrcoef_plot_2}.

\section{What Factors Govern Entropy Dynamics, Both Theoretically and Empirically?}
\subsection{Clipping Threshold}
\label{subsec:appendix_clipping_threshold}

A detailed description of the GRPO clipping variants explored in our study is presented below:

\begin{figure}[t]
    \centering
    \includegraphics[width=\linewidth]{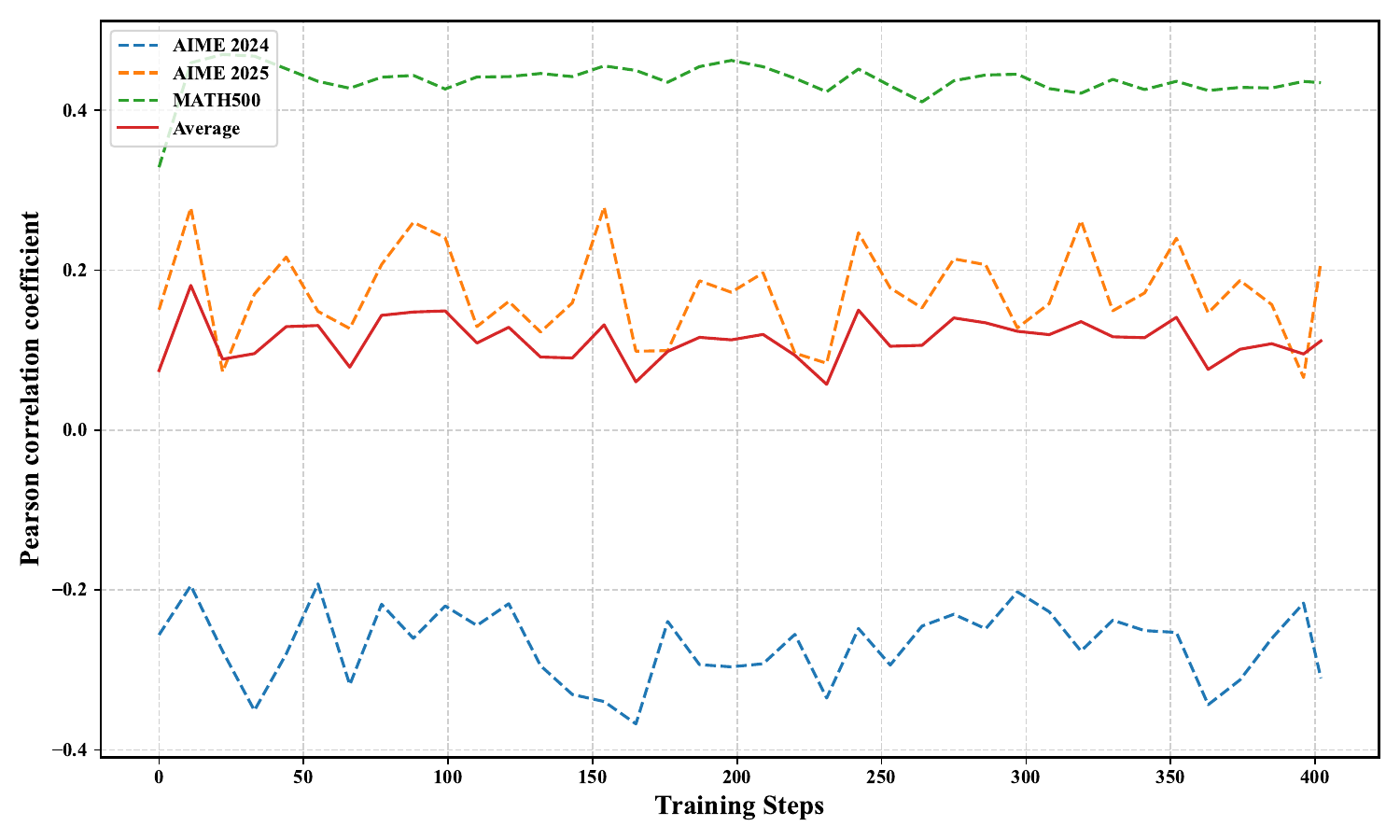}
    \caption{Spearman's rank correlation coefficients between the entropy of LLMs on prompts and their corresponding accuracy across different training steps. ``Average'' indicates the mean Spearman's rank correlation coefficient computed over AIME 2024, AIME 2025, and MATH500.}
    \label{fig:entropy_acc_prompts_corrcoef_plot}
\end{figure}

\begin{table*}[t]
  \centering
  \resizebox{\textwidth}{!}{
    \begin{tabular}{l|l|l|l|l|l|c|l|>{\columncolor{gray!15}}c|>{\columncolor{gray!15}}c|>{\columncolor{gray!15}}c}
    \toprule
          \multicolumn{1}{c|}{\textbf{Model}} & \multicolumn{1}{c|}{\textbf{AIME 2024}} & \multicolumn{1}{c|}{\textbf{AIME 2025}} & \multicolumn{1}{c|}{\textbf{MATH500}} & \multicolumn{1}{c|}{\textbf{AMC 2023}} & \multicolumn{1}{c|}{\textbf{Minerva}} & \multicolumn{1}{c|}{\textbf{LiveCodeBench}} & \multicolumn{1}{c|}{\textbf{IF-Eval}} & \textbf{Average (ID)} & \textbf{Average (OOD)} & \textbf{Entropy} \\
    \midrule
    Qwen2.5-Math-7B & 10.00 / 60.00 & 3.80 / 33.33 & 43.76 / 95.60 & 30.04 / 92.50 & 14.41 / 60.29 & 3.62 / 30.15 & 22.67 / 80.46 & 20.40 / 68.35 & 13.15 / 55.30 & N/A \\
    \arrayrulecolor{black}\midrule
    \quad + GRPO (Full-data) & 28.75 / 63.33 & 14.69 / 50.00 & 78.14 / 96.80 & 64.38 / 97.50 & 34.64 / 64.34 & 7.85 / 33.46 & 30.17 / 72.90 & 44.12 / 74.39 & 19.01 / 53.18 & 0.11838 \\
    \arrayrulecolor{lightgray}\midrule
    \quad + GRPO (10,001{\tiny K-means}) & 31.41 / 63.33 & 14.22 / 46.67 & 76.69 / 95.60 & 64.69 / 95.00 & 34.38 / 65.81 & 7.67 / 33.82 & 29.71 / 73.74 & 44.28 / 73.28 & 18.69 / 53.78 & 0.11419 \\
    \midrule
    \quad + GRPO (10,001{\tiny random}) & 30.00 / 63.33 & 15.10 / 50.00 & 77.29 / 96.00 & 63.48 / 92.50 & 34.48 / 65.81 & 8.08 / 35.29 & 30.56 / 71.46 & 44.07 / 73.53 & 19.32 / 53.38 & 0.11777 \\
    \midrule
    \quad + GRPO (5,031{\tiny K-means}) & 30.16 / 66.67 & 13.75 / 46.67 & 74.86 / 95.40 & 62.85 / 95.00 & 33.99 / 65.81 & 8.40 / 34.19 & 30.90 / 71.58 & 43.12 / 73.91 & 19.65 / 52.89 & 0.10722 \\
    \midrule
    \quad + GRPO (5,031{\tiny random}) & 30.47 / 70.00 & 14.53 / 50.00 & 74.80 / 94.80 & 63.36 / 95.00 & 33.32 / 64.34 & 7.95 / 34.93 & 30.29 / 72.78 & 43.30 / 74.83 & 19.12 / 53.85 & 0.09953 \\
    \midrule
    \quad + GRPO (2,538{\tiny K-means}) & 30.94 / 70.00 & 13.85 / 46.67 & 75.29 / 95.40 & 62.23 / 100.00 & 34.09 / 63.24 & 8.77 / 35.66 & 30.41 / 73.14 & 43.28 / 75.06 & 19.59 / 54.40 & 0.08674 \\
    \midrule
    \quad + GRPO (2,538{\tiny random}) & 30.00 / 70.00 & 14.32 / 53.33 & 74.58 / 95.00 & 63.55 / 95.00 & 33.67 / 63.24 & 8.13 / 31.62 & 29.49 / 72.90 & 43.22 / 75.31 & 18.81 / 52.26 & 0.10369 \\
    \midrule
    \quad + GRPO (1,246{\tiny K-means}) & 30.31 / 70.00 & 14.01 / 36.67 & 76.83 / 95.80 & 63.98 / 95.00 & 34.71 / 62.50 & 8.65 / 33.09 & 30.45 / 71.70 & 43.97 / 71.99 & 19.55 / 52.40 & 0.08588 \\
    \midrule
    \quad + GRPO (1,246{\tiny random}) & 30.47 / 63.33 & 14.69 / 56.67 & 75.99 / 95.20 & 65.27 / 95.00 & 34.29 / 65.44 & 8.87 / 34.56 & 30.47 / 70.50 & 44.14 / 75.13 & 19.67 / 52.53 & 0.09030 \\
    \midrule
    \quad + GRPO (616{\tiny K-means}) & 29.32 / 60.00 & 17.60 / 46.67 & 78.57 / 94.80 & 64.10 / 87.50 & 36.05 / 59.19 & 9.06 / 34.93 & 30.70 / 67.75 & 45.13 / 69.63 & 19.88 / 51.34 & 0.02398 \\
    \midrule
    \quad + GRPO (616{\tiny random}) & 27.55 / 63.33 & 15.00 / 53.33 & 68.54 / 89.40 & 65.27 / 95.00 & 29.62 / 58.09 & 9.99 / 36.03 & 31.23 / 70.26 & 41.20 / 71.83 & 20.61 / 53.15 & 0.02926 \\
    \arrayrulecolor{black}\bottomrule
    \end{tabular}
  }
  \caption{Performance of Qwen2.5-Math-7B trained with GRPO under varying data scales. ``Entropy'' denotes the EMA-smoothed entropy at the final training step. Numbers in parentheses indicate the number of training samples, with subscripts ``K-means'' and ``random'' referring to datasets selected via K-means clustering and random sampling, respectively. ``Average (ID)'' and ``Average (OOD)'' denote the mean performance across in-domain and out-of-domain benchmarks. Results are reported as A / B, corresponding to Avg@64 and Pass@64.}
  \label{tab:data_diversity}
\end{table*}

\begin{itemize}
    \item \textbf{Clip-Higher} \citep{DBLP:journals/corr/abs-2503-14476} raises the upper clipping bound in the GRPO objective to reduce the proportion of low-probability tokens being clipped. This relaxation allows these tokens to increase their likelihoods more freely, thereby enhancing exploration and mitigating entropy collapse in LLMs. Following \citet{DBLP:journals/corr/abs-2503-14476}, we set $\varepsilon_{\text{high}} = 0.28$.
    \item \textbf{Clip-Lower} adopts a similar design to \textbf{Clip-Higher} but increases $\varepsilon_{\text{low}}$, thereby lowering the clipping lower bound in the GRPO objective. Increasing $\varepsilon_{\text{low}}$ makes low-probability tokens with negative advantages less susceptible to clipping, allowing their probabilities to decrease more rapidly. Consequently, \textbf{Clip-Lower} is expected to intensify entropy collapse in LLMs. To align the setup with \textbf{Clip-Higher}, we set $\varepsilon_{\text{low}}$ to 0.28.
    \item \textbf{Clip-Tighter}, in contrast to \textbf{Clip-Lower}, decreases $\varepsilon_{\text{low}}$, thereby raising the lower clipping bound in the GRPO objective. Consequently, low-probability tokens with negative advantages are more likely to be clipped, preventing excessive suppression of their probabilities and mitigating entropy collapse in LLMs. Concretely, while \textbf{Clip-Lower} increases $\varepsilon_{\text{low}}$ from 0.2 to 0.28 (+0.08), \textbf{Clip-Tighter} symmetrically decreases it by 0.08, yielding a final value of 0.12.
    \item \textbf{Clip-Free} removes the clipping operation from the GRPO objective. The clipping mechanism, inherited from PPO, serves to penalize updates that deviate substantially from the current policy, thereby stabilizing training. Removing it allows us to examine how clipping influences the entropy dynamics of LLMs and training stability.
\end{itemize}

\subsection{Off-Policy Updates}
\label{subsec:appendix_off_policy_updates}
\begin{figure}[t]
    \centering
    \includegraphics[width=\linewidth]{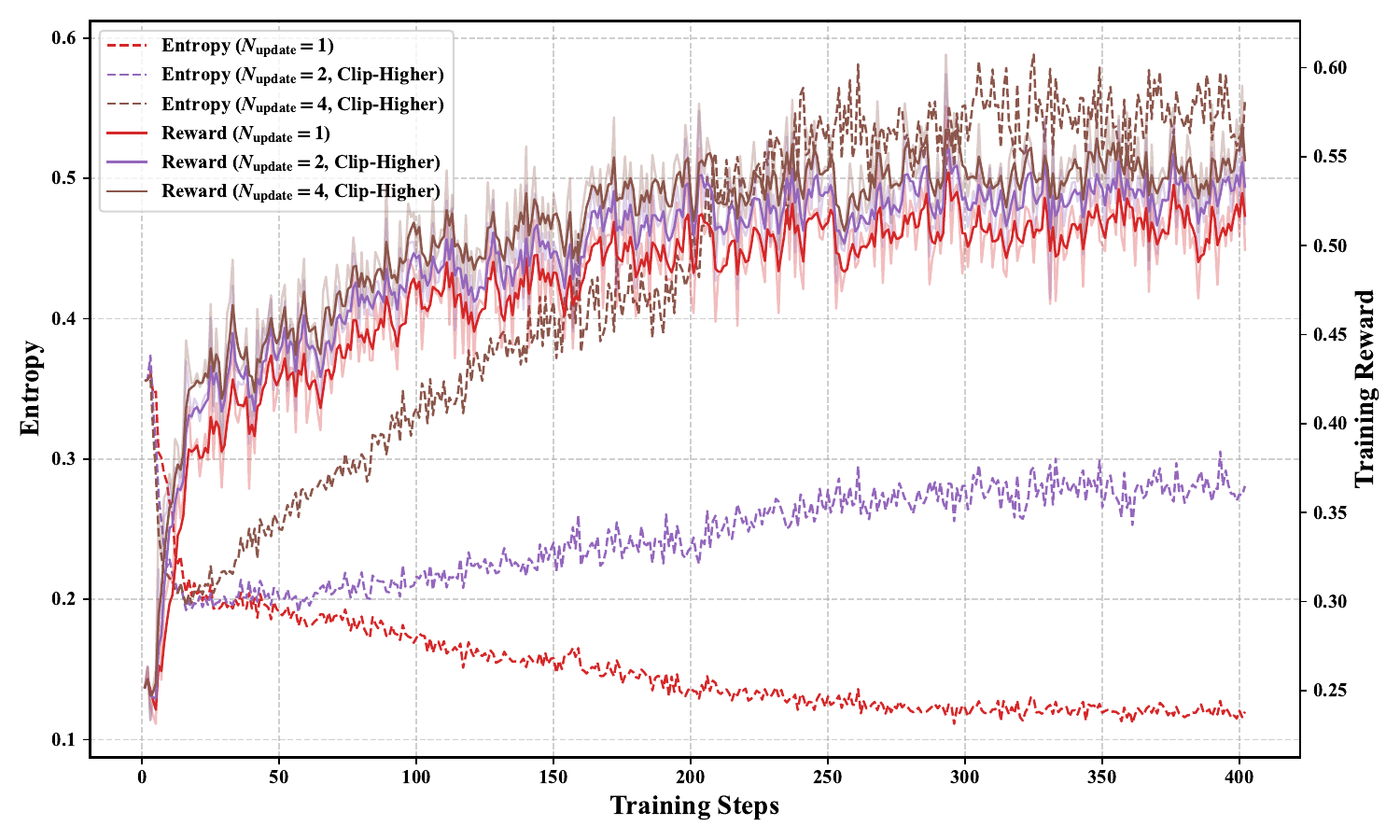}
    \caption{Evolution of entropy and training rewards under the Clip-Higher setting.}
    \label{fig:entropy_train_reward_2}
\end{figure}

The evolution of entropy and reward on the training set during LLM training under the Clip-Higher setting is presented in Figure~\ref{fig:entropy_train_reward_2}.

\subsection{Training Data Diversity}
\label{subsec:appendix_training_data_diversity}

For K-means clustering, each prompt is represented by a mean-pooled token embedding from the final layer of Qwen2.5-Math-7B. We perform K-means clustering with $K = 1,000$ and sort the resulting clusters in descending order according to their sample counts. Subsets are constructed by selecting samples from the top $M \in \{281, 112, 49, 21, 9\}$ clusters, yielding subsets of 10,001, 5,031, 2,538, 1,246, and 616 samples, respectively. For random sampling, we select the same number of samples to ensure a fair comparison. Qwen2.5-Math-7B is then trained with RLVR on all subsets using identical hyperparameters.

Since the entropy of LLMs evolves dynamically during training, we employ an Exponential Moving Average (EMA) to mitigate short-term fluctuations and report the smoothed entropy at the final training step. Formally, at step $k$, the EMA-smoothed entropy $\mathcal{H}^{\text{EMA}}_k$ is defined as:
\begin{equation}
\small
\mathcal{H}^{\text{EMA}}_k = \mathcal{H}_k (\pi_\theta^{k}) \left( 1 - \varphi \right) + \varphi \mathcal{H}^{\text{EMA}}_{k-1},
\end{equation}
where the smoothing coefficient $\varphi$ is set to 0.6. The experimental results are summarized in Table~\ref{tab:data_diversity}.

\section{How Can Entropy Be Effectively Regulated to Improve the Performance of LLMs?}
\label{sec:appendix_RQ_3}

\subsection{Theoretical Analysis}
\label{subsec:appendix_theoretical_analysis}

To clearly illustrate the derivation of the gradient of the GRPO optimization objective with respect to the logit $z_v$ of token $v$, we first present the derivation without clipping, and then incorporate the clipping operation. Specifically, the GRPO optimization objective can be formulated as follows:
\begin{equation}
\small
\mathcal{J} \left(\theta \right) = E_{\bm{y}_t \sim \pi_{\theta_\text{old}} \left(\cdot \mid \bm{x}, \bm{y}_{<t} \right)} \frac{\pi_{\theta}(\bm{y}_t \mid \bm{x}, \bm{y}_{<t})}{\pi_{\theta_\text{old}} \left(\bm{y}_t \mid \bm{x}, \bm{y}_{<t} \right)} \hat{A}_{t}
\end{equation}
The gradient of the GRPO optimization objective with respect to the logit $z_v$ of token $v$ is given by:
\begin{equation}
\small
\begin{aligned}
\frac{\partial \mathcal{J} \left(\theta \right)}{\partial z_v} &= \frac{\partial E_{\bm{y}_t \sim \pi_{\theta_\text{old}} \left(\cdot \mid \bm{x}, \bm{y}_{<t} \right)} \frac{\pi_{\theta} \left(\bm{y}_t \mid \bm{x}, \bm{y}_{<t} \right)}{\pi_{\theta_\text{old}} \left(\bm{y}_t \mid \bm{x}, \bm{y}_{<t} \right)} \hat{A}_{t}}{\partial z_v} \\
&= \frac{\partial \sum_{\bm{y}_t} \pi_{\theta_\text{old}} \left(\bm{y}_t \mid \bm{x}, \bm{y}_{<t} \right) \frac{\pi_{\theta} \left(\bm{y}_t \mid \bm{x}, \bm{y}_{<t} \right)}{\pi_{\theta_\text{old}} \left(\bm{y}_t \mid \bm{x}, \bm{y}_{<t} \right)} \hat{A}_{t}}{\partial z_v} \\
&= \frac{\partial \sum_{\bm{y}_t} \pi_{\theta} \left(\bm{y}_t \mid \bm{x}, \bm{y}_{<t} \right) \hat{A}_{t}}{\partial z_v} \\
&= \frac{\sum_{\bm{y}_t} \partial \pi_{\theta} \left(\bm{y}_t \mid \bm{x}, \bm{y}_{<t} \right) \hat{A}_{t}}{\partial z_v} \\
&= \frac{\sum_{\bm{y}_t} \pi_{\theta_\text{old}} \left(\bm{y}_t \mid \bm{x}, \bm{y}_{<t} \right) \partial \pi_{\theta} \left(\bm{y}_t \mid \bm{x}, \bm{y}_{<t} \right) \hat{A}_{t}}{\pi_{\theta_\text{old}} \left(\bm{y}_t \mid \bm{x}, \bm{y}_{<t} \right) \partial z_v} \\
&= \frac{E_{\bm{y}_t \sim \pi_{\theta_\text{old}} \left(\cdot \mid \bm{x}, \bm{y}_{<t} \right)} \partial \pi_{\theta} \left(\bm{y}_t \mid \bm{x}, \bm{y}_{<t} \right) \hat{A}_{t}}{\pi_{\theta_\text{old}} \left(\bm{y}_t \mid \bm{x}, \bm{y}_{<t} \right) \partial z_v}
\end{aligned}
\label{eq:logits_gradient_1}
\end{equation}
When token $v$ \textbf{is not} sampled during the generation of $\bm{y}_t$, the gradient of $\pi_{\theta}\left(\bm{y}_t \vert \bm{x}, \bm{y}_{<t} \right)$ with respect to $z_v$ is:
\begin{equation}
\small
\begin{aligned}
\frac{\partial \pi_{\theta}\left(\bm{y}_t \vert \bm{x}, \bm{y}_{<t} \right)}{\partial z_v} &= \frac{\partial \frac{\text{exp}\left( z_{\bm{y}_t} \right)}{\sum_{v' \in \mathcal{V}}{\text{exp}\left( z_{v'} \right)}}}{\partial z_v} \\
& = \frac{ -\text{exp}\left( z_{\bm{y}_t} \right) \text{exp} \left ( z_v \right) }{ \left( \sum_{v' \in \mathcal{V}}{\text{exp}\left( z_{v'} \right)} \right)^2 } \\
& = - \pi_{\theta}\left(\bm{y}_t \vert \bm{x}, \bm{y}_{<t} \right) \pi_{\theta}\left(v \vert \bm{x}, \bm{y}_{<t} \right)
\end{aligned}
\label{eq:logits_gradient_2}
\end{equation}
Conversely, when token $v$ \textbf{is} sampled during the generation of $\bm{y}_t$, the gradient becomes:
\begin{equation}
\small
\begin{aligned}
\frac{\partial \pi_{\theta}\left(\bm{y}_t \vert \bm{x}, \bm{y}_{<t} \right)}{\partial z_v} &= \frac{\partial \frac{\text{exp}\left( z_{\bm{y}_t} \right)}{\sum_{v' \in \mathcal{V}}{\text{exp}\left( z_{v'} \right)}}}{\partial z_v} \\
& = \frac{\text{exp}\left( z_{\bm{y}_t} \right) \sum_{v' \in \mathcal{V}}{\text{exp}\left( z_{v'} \right)} - \text{exp}\left( z_{\bm{y}_t} \right)^2}{\left( \sum_{v' \in \mathcal{V}}{\text{exp}\left( z_{v'} \right)} \right)^2} \\
& = \frac{\text{exp}\left( z_{\bm{y}_t} \right)}{\left( \sum_{v' \in \mathcal{V}}{\text{exp}\left( z_{v'} \right)} \right)} - \frac{\text{exp}\left( z_{\bm{y}_t} \right)^2}{\left( \sum_{v' \in \mathcal{V}}{\text{exp}\left( z_{v'} \right)} \right)^2} \\
& = \pi_{\theta}\left(\bm{y}_t \vert \bm{x}, \bm{y}_{<t} \right)  - \pi_{\theta}\left(\bm{y}_t \vert \bm{x}, \bm{y}_{<t} \right) ^ 2 \\
& = \pi_{\theta}\left(\bm{y}_t \vert \bm{x}, \bm{y}_{<t} \right) \left( 1 - \pi_{\theta}\left(\bm{y}_t \vert \bm{x}, \bm{y}_{<t} \right) \right)
\end{aligned}
\label{eq:logits_gradient_3}
\end{equation}

By substituting Eq.~(\ref{eq:logits_gradient_2}) into Eq.~(\ref{eq:logits_gradient_1}), we obtain the gradient of the GRPO optimization objective with respect to $z_v$ when token $v$ \textbf{is not} sampled:
\begin{equation}
\small
\begin{aligned}
\frac{\partial \mathcal{J} \left(\theta \right)}{\partial z_v} &= \frac{E_{\bm{y}_t \sim \pi_{\theta_\text{old}} \left(\cdot \mid \bm{x}, \bm{y}_{<t} \right)} \partial \pi_{\theta} \left(\bm{y}_t \mid \bm{x}, \bm{y}_{<t} \right) \hat{A}_{t}}{\pi_{\theta_\text{old}}\left(\bm{y}_t \mid \bm{x}, \bm{y}_{<t} \right) \partial z_v} \\
&= \frac{- E_{\bm{y}_t \sim \pi_{\theta_\text{old}} \left(\cdot \mid \bm{x}, \bm{y}_{<t} \right)} \pi_{\theta} \left(\bm{y}_t \mid \bm{x}, \bm{y}_{<t} \right) \pi_{\theta} \left(v \mid \bm{x}, \bm{y}_{<t} \right) \hat{A}_{t}}{\pi_{\theta_\text{old}} \left(\bm{y}_t \mid \bm{x}, \bm{y}_{<t} \right)} \\
& \approx \frac{- \pi_{\theta} \left(\bm{y}_t \mid \bm{x}, \bm{y}_{<t} \right) \pi_{\theta} \left(v \mid \bm{x}, \bm{y}_{<t} \right) \hat{A}_{t}}{\pi_{\theta_\text{old}} \left(\bm{y}_t \mid \bm{x}, \bm{y}_{<t} \right)}
\end{aligned}
\label{eq:logits_gradient_4}
\end{equation}
Similarly, substituting Eq.~(\ref{eq:logits_gradient_3}) into Eq.~(\ref{eq:logits_gradient_1}) yields the gradient when token $v$ \textbf{is} sampled:
\begin{equation}
\small
\begin{aligned}
\frac{\partial \mathcal{J} \left(\theta \right)}{\partial z_v} &= \frac{E_{\bm{y}_t \sim \pi_{\theta_\text{old}} \left(\cdot \mid \bm{x}, \bm{y}_{<t} \right)} \partial \pi_{\theta} \left(\bm{y}_t \mid \bm{x}, \bm{y}_{<t} \right) \hat{A}_{t}}{\pi_{\theta_\text{old}}\left(\bm{y}_t \mid \bm{x}, \bm{y}_{<t} \right) \partial z_v} \\
&= \frac{E_{\bm{y}_t \sim \pi_{\theta_\text{old}} \left(\cdot \mid \bm{x}, \bm{y}_{<t} \right)} \pi_{\theta} \left(\bm{y}_t \mid \bm{x}, \bm{y}_{<t} \right) \left(1 - \pi_{\theta} \left(\bm{y}_t \mid \bm{x}, \bm{y}_{<t} \right) \right) \hat{A}_{t}}{\pi_{\theta_\text{old}} \left(\bm{y}_t \mid \bm{x}, \bm{y}_{<t} \right)} \\
&\approx \frac{\pi_{\theta} \left(\bm{y}_t \mid \bm{x}, \bm{y}_{<t} \right) \left(1 - \pi_{\theta} \left(\bm{y}_t \mid \bm{x}, \bm{y}_{<t} \right) \right) \hat{A}_{t}}{\pi_{\theta_\text{old}}\left(\bm{y}_t \mid \bm{x}, \bm{y}_{<t} \right)} \\
&\approx \frac{\pi_{\theta} \left(v \mid \bm{x}, \bm{y}_{<t} \right) \left(1 - \pi_{\theta} \left(v \mid \bm{x}, \bm{y}_{<t} \right) \right) \hat{A}_{t}}{\pi_{\theta_\text{old}} \left(v \mid \bm{x}, \bm{y}_{<t} \right)}
\end{aligned}
\end{equation}

When the importance sampling ratio in the GRPO optimization objective falls into the clipped region, the gradient with respect to $z_v$ becomes zero. Consequently, when token $v$ \textbf{is not} sampled at step $t$, the gradient of the GRPO objective with respect to $z_v$ can be approximated as follows:
\begin{equation}
\small
\frac{\partial \mathcal{J} \left(\theta \right)}{\partial z_v} = 
\begin{cases}
- r_{t} \left( \theta \right) \pi_{\theta} \left(v \mid \bm{x}, \bm{y}_{<t} \right) \hat{A}_{t}, \\
\quad \quad \;\;\; \text{if } \hat{A}_{t} > 0 \text{ and } r_{t} \left( \theta \right) < 1 + \varepsilon_{\text{high}} \\
0, \\
\quad \quad \;\;\; \text{if } \hat{A}_{t} > 0 \text{ and } r_{t} \left( \theta \right) > 1 + \varepsilon_{\text{high}} \\
- r_{t} \left( \theta \right) \pi_{\theta} \left(v \mid \bm{x}, \bm{y}_{<t} \right) \hat{A}_{t}, \\
\quad \quad \;\;\; \text{if } \hat{A}_{t} < 0 \text{ and } r_{t} \left( \theta \right) > 1 - \varepsilon_{\text{low}} \\
0,\\
\quad \quad \;\;\; \text{if } \hat{A}_{t} < 0 \text{ and } r_{t} \left( \theta \right) < 1 - \varepsilon_{\text{low}}
\end{cases}
\end{equation}

Similarly, when token $v$ \textbf{is} sampled at step $t$, the gradient can be approximated as follows:
\begin{equation}
\small
\frac{\partial \mathcal{J} \left(\theta \right)}{\partial z_v} = 
\begin{cases}
r_{t} \left( \theta \right) \left(1 - \pi_{\theta} \left(v \mid \bm{x}, \bm{y}_{<t} \right) \right) \hat{A}_{t}, \\
\quad \quad \;\;\; \text{if } \hat{A}_{t} > 0 \text{ and } r_{t} \left( \theta \right) < 1 + \varepsilon_{\text{high}} \\
0, \\
\quad \quad \;\;\; \text{if } \hat{A}_{t} > 0 \text{ and } r_{t} \left( \theta \right) > 1 + \varepsilon_{\text{high}} \\
r_{t} \left( \theta \right) \left(1 - \pi_{\theta} \left(v \mid \bm{x}, \bm{y}_{<t} \right) \right) \hat{A}_{t}, \\
\quad \quad \;\;\; \text{if } \hat{A}_{t} < 0 \text{ and } r_{t} \left( \theta \right) > 1 - \varepsilon_{\text{low}} \\
0, \\
\quad \quad \;\;\; \text{if } \hat{A}_{t} < 0 \text{ and } r_{t} \left( \theta \right) < 1 - \varepsilon_{\text{low}}
\end{cases}
\end{equation}

\subsection{Empirical Analysis}
\label{subsec:appendix_empirical_analysis}

The baseline methods against which we compare are described as follows:

\begin{itemize}
\item \textbf{Adaptive Entropy Regularization} dynamically adjusts the entropy regularization coefficient to keep model entropy above a predefined threshold $\delta$. We consider two settings for $\delta$: (1) $\delta = 0.2$, following \citet{DBLP:journals/corr/abs-2505-22312}, and (2) $\delta$ set to the entropy of the LLM on responses to 1,000 randomly sampled training prompts before training, which equals 0.3657.
\item \textbf{Clip-Cov} \citep{DBLP:journals/corr/abs-2505-22617} mitigates entropy collapse by zeroing the gradients of a small subset of tokens with high covariance between log probabilities and advantages.
\item \textbf{KL-Cov} \citep{DBLP:journals/corr/abs-2505-22617} adopts a similar approach to Clip-Cov but applies a KL penalty to high-covariance tokens.
\item \textbf{Entropy-Adv} \citep{DBLP:journals/corr/abs-2506-14758} augments the original advantage with an entropy term to encourage exploratory reasoning tokens and enhance LLM performance.
\item \textbf{Rand-Pos-Clip} serves as a counterpart to \textbf{Clip-Cov}. Unlike \textbf{Clip-Cov}, which sets the gradients of a small subset of tokens exhibiting high covariance between log probabilities and advantages to zero, \textbf{Rand-Pos-Clip} randomly zeroes the gradients of a subset of tokens with positive advantages. To ensure a fair comparison between the two methods, we maintain the same proportion of tokens whose gradients are set to zero in both \textbf{Rand-Pos-Clip} and \textbf{Clip-Cov}.
\end{itemize}

We conduct experiments with $N_{\text{update}}$ fixed at 4 to evaluate the effectiveness of different approaches in controlling the entropy of LLMs.

\subsection{Analysis of the Covariance between Log-Probability and Advantage}
\label{subsec:appendix_analysis_of_the_covariance}

\begin{figure*}[t]
    \centering
    \begin{subfigure}{0.495\textwidth}
        \includegraphics[width=\linewidth]{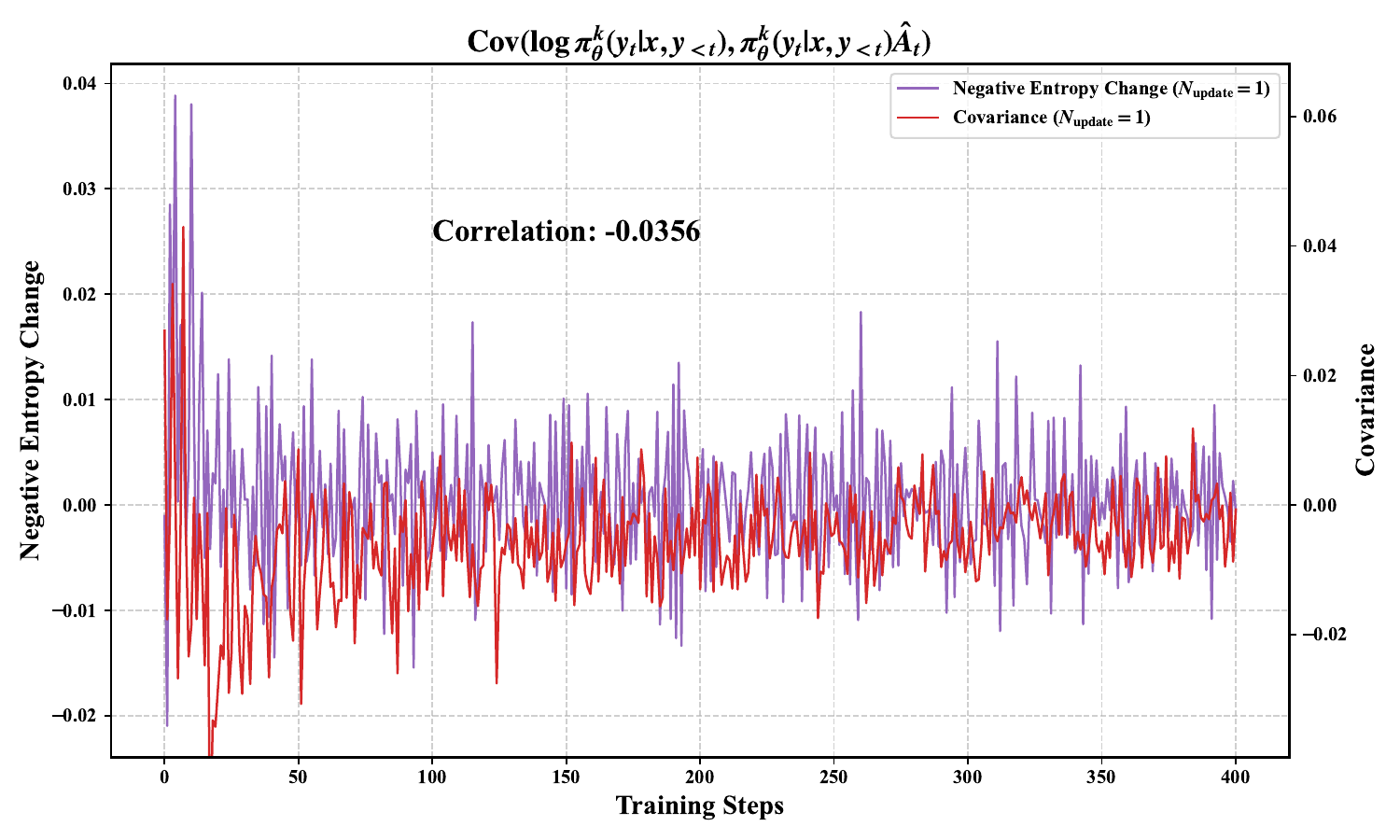}
    \end{subfigure}
    \begin{subfigure}{0.495\textwidth}
        \includegraphics[width=\linewidth]{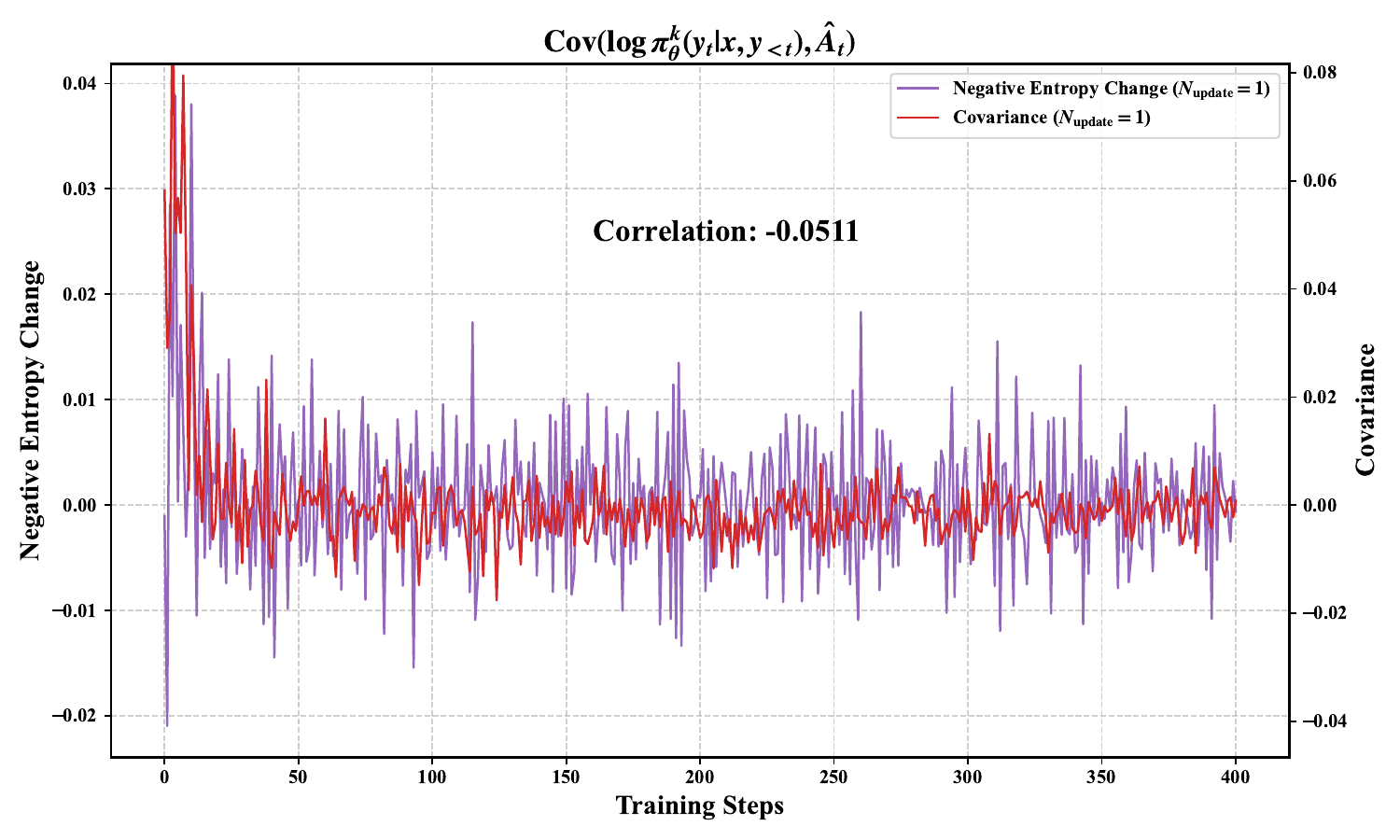}
    \end{subfigure}
    \caption{Evolution of the negative entropy change and the covariance terms during GRPO training of Qwen2.5-Math-7B with $N_{\text{update}} = 1$. The left figure reports the covariance between $\log \pi_\theta^k \left(\bm{y}_t \vert \bm{x}, \bm{y}_{<t} \right)$ and $\pi_\theta^k \left(\bm{y}_t \vert \bm{x}, \bm{y}_{<t} \right) \hat{A}_t$, while the right panel reports the covariance between $\log \pi_\theta^k \left(\bm{y}_t \vert \bm{x}, \bm{y}_{<t} \right)$ and $\hat{A}_t$.}
    \label{fig:entropy_change_cov_plot_001}
\end{figure*}

\begin{figure*}[t]
    \centering
    \begin{subfigure}{0.495\textwidth}
        \includegraphics[width=\linewidth]{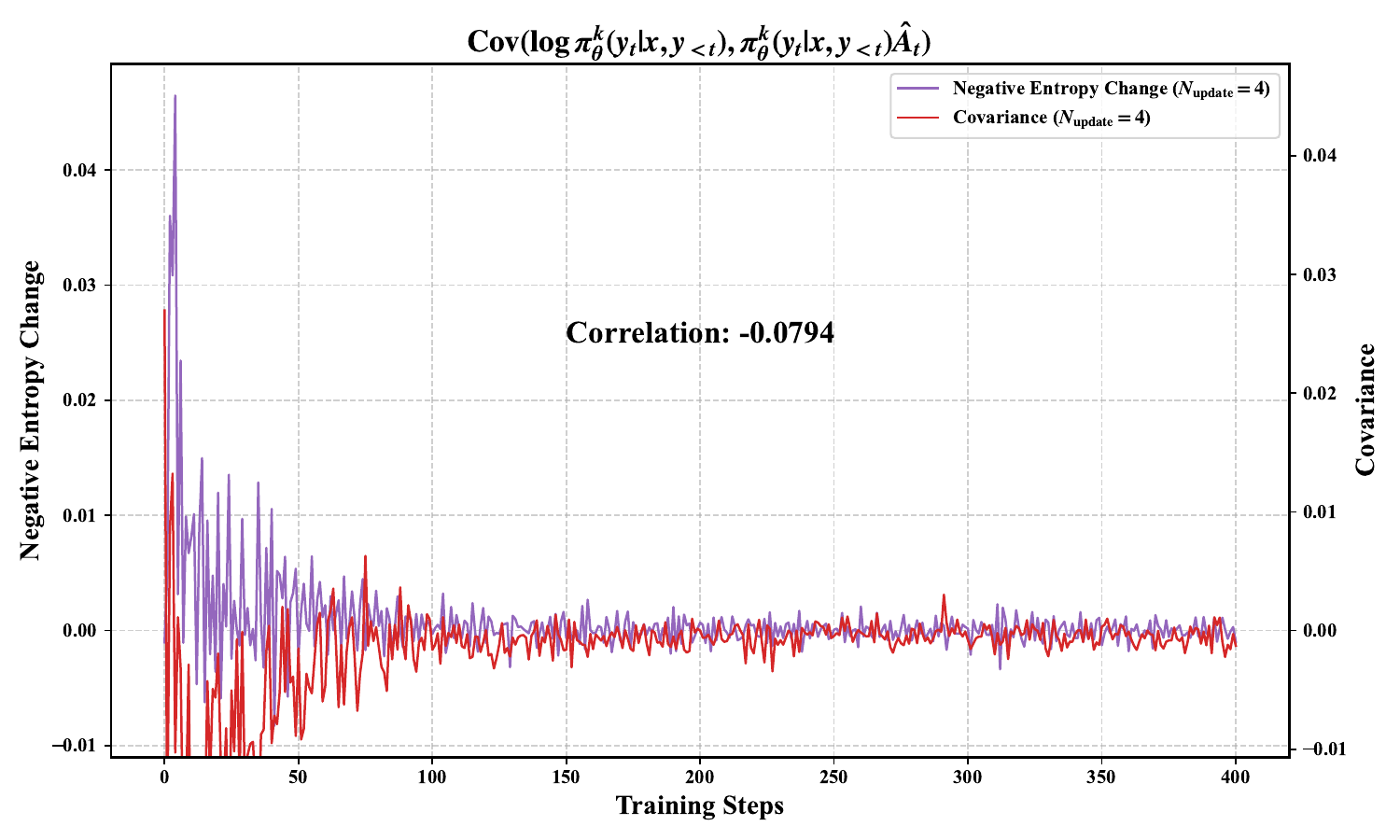}
    \end{subfigure}
    \begin{subfigure}{0.495\textwidth}
        \includegraphics[width=\linewidth]{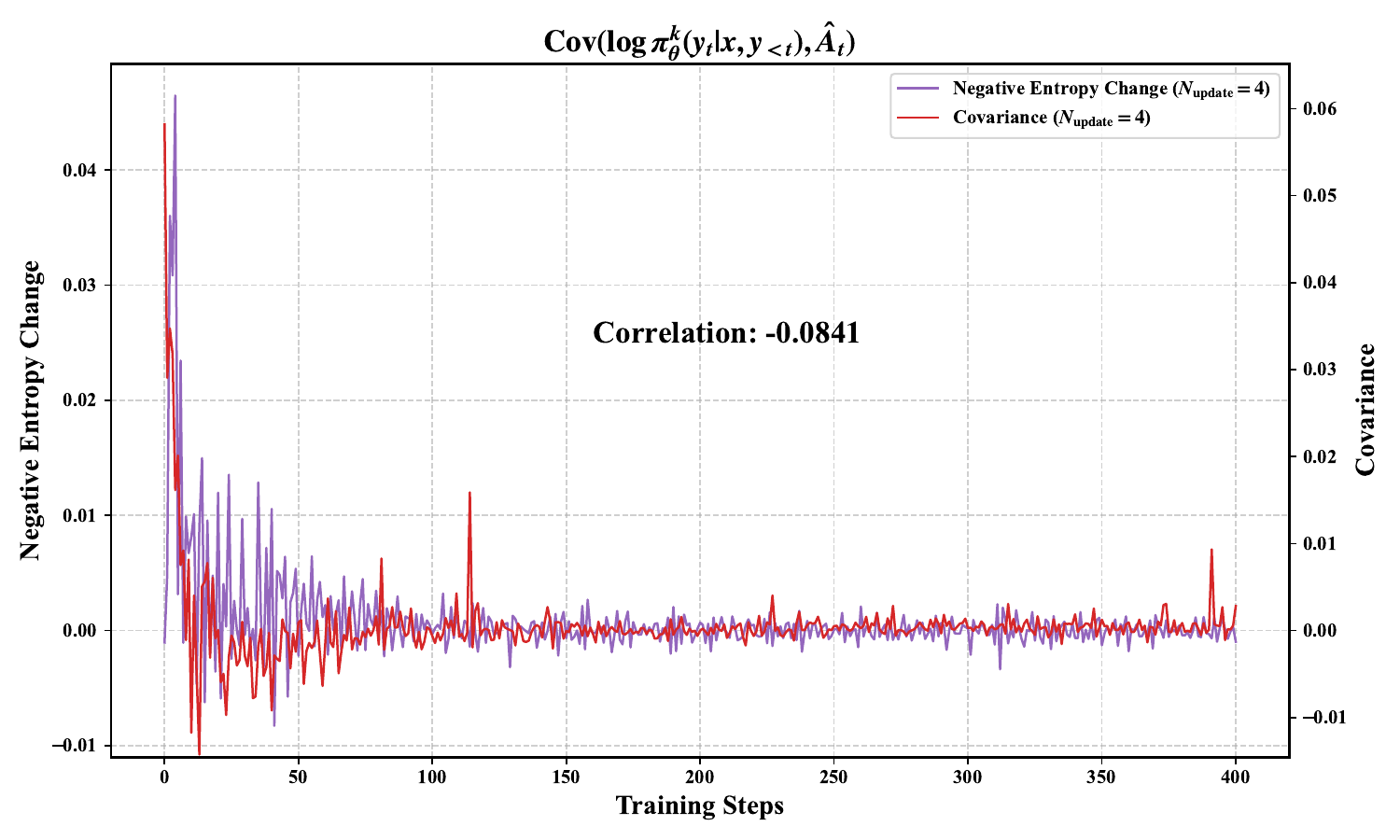}
    \end{subfigure}
    \caption{Evolution of the negative entropy change and the covariance terms during GRPO training of Qwen2.5-Math-7B with $N_{\text{update}} = 4$. The left figure reports the covariance between $\log \pi_\theta^k \left(\bm{y}_t \vert \bm{x}, \bm{y}_{<t} \right)$ and $\pi_\theta^k \left(\bm{y}_t \vert \bm{x}, \bm{y}_{<t} \right) \hat{A}_t$, while the right panel reports the covariance between $\log \pi_\theta^k \left(\bm{y}_t \vert \bm{x}, \bm{y}_{<t} \right)$ and $\hat{A}_t$.}
    \label{fig:entropy_change_cov_plot_012}
\end{figure*}

\citet{zhihu2025entropy} and \citet{DBLP:journals/corr/abs-2505-22617} provide a theoretical analysis showing that the change in entropy of an LLM between two consecutive training steps is governed by the covariance between token log-probabilities and advantages. Formally, let $\eta$ denote the learning rate at training step $k$, and let $\bm{y}_{t}$ be a token sampled from the policy $\pi_\theta^k \left ( \bm{y}_t \vert \bm{x}, \bm{y}_{<t} \right)$. Under the policy gradient, the entropy change of the policy $\pi_\theta$ from step $k+1$ to $k$ can be approximated as follows:
\begin{equation}
\small
\begin{aligned}
\mathcal{H} \left (\pi_\theta^{k+1} \left (\bm{y_t} \vert \bm{x}, \bm{y}_{<t} \right) \right) - \mathcal{H}\left (\pi_\theta^k \left ( \bm{y}_t \vert \bm{x}, \bm{y}_{<t} \right) \right) \approx \\
-\eta \cdot \text{Cov} \left(\log \pi_\theta^k \left(\bm{y}_t \vert \bm{x}, \bm{y}_{<t} \right), \pi_\theta^k \left(\bm{y}_t \vert \bm{x}, \bm{y}_{<t} \right) \hat{A}_t \right).
\end{aligned}
\end{equation}

\begin{figure*}[t]
    \centering
    \begin{subfigure}{0.495\textwidth}
        \includegraphics[width=\linewidth]{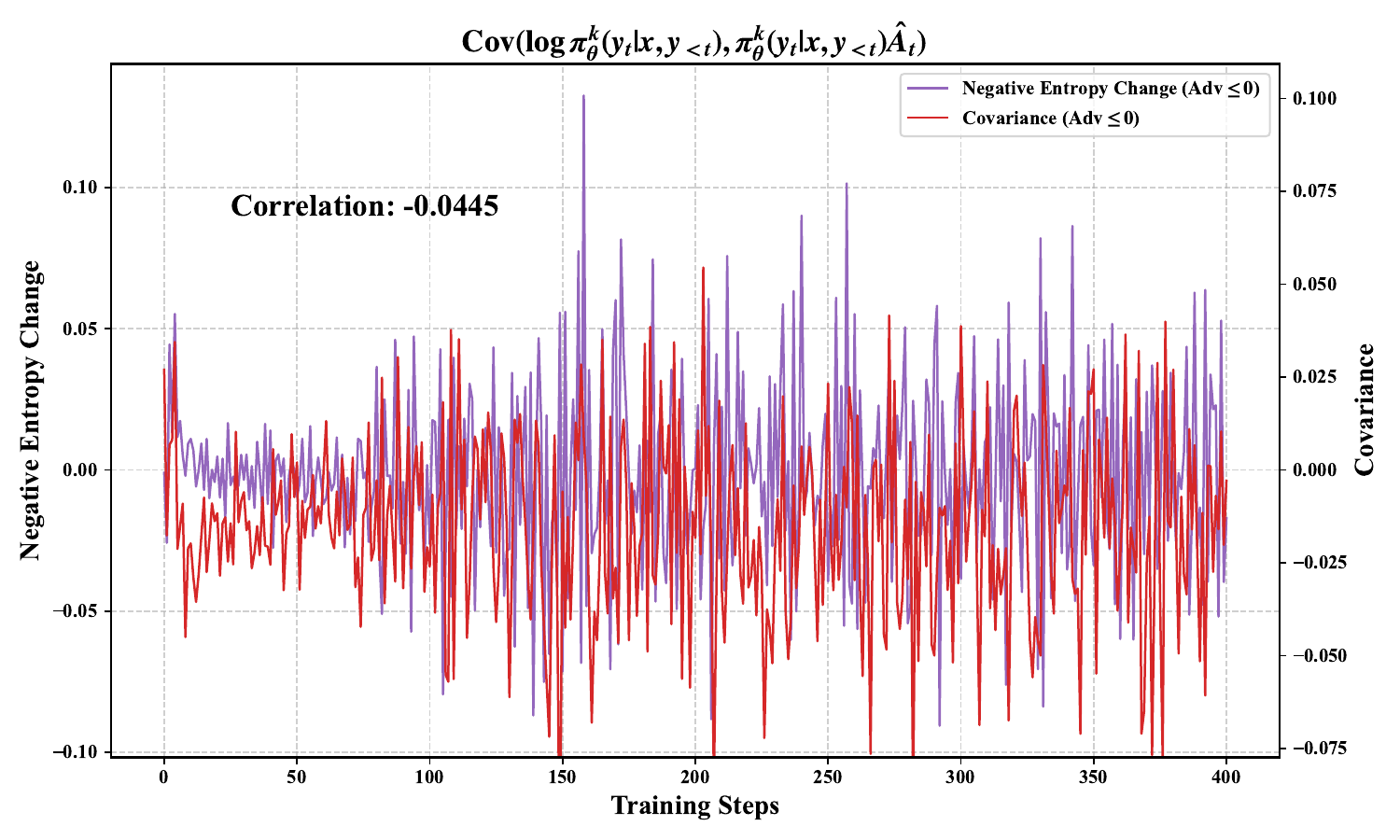}
    \end{subfigure}
    \begin{subfigure}{0.495\textwidth}
        \includegraphics[width=\linewidth]{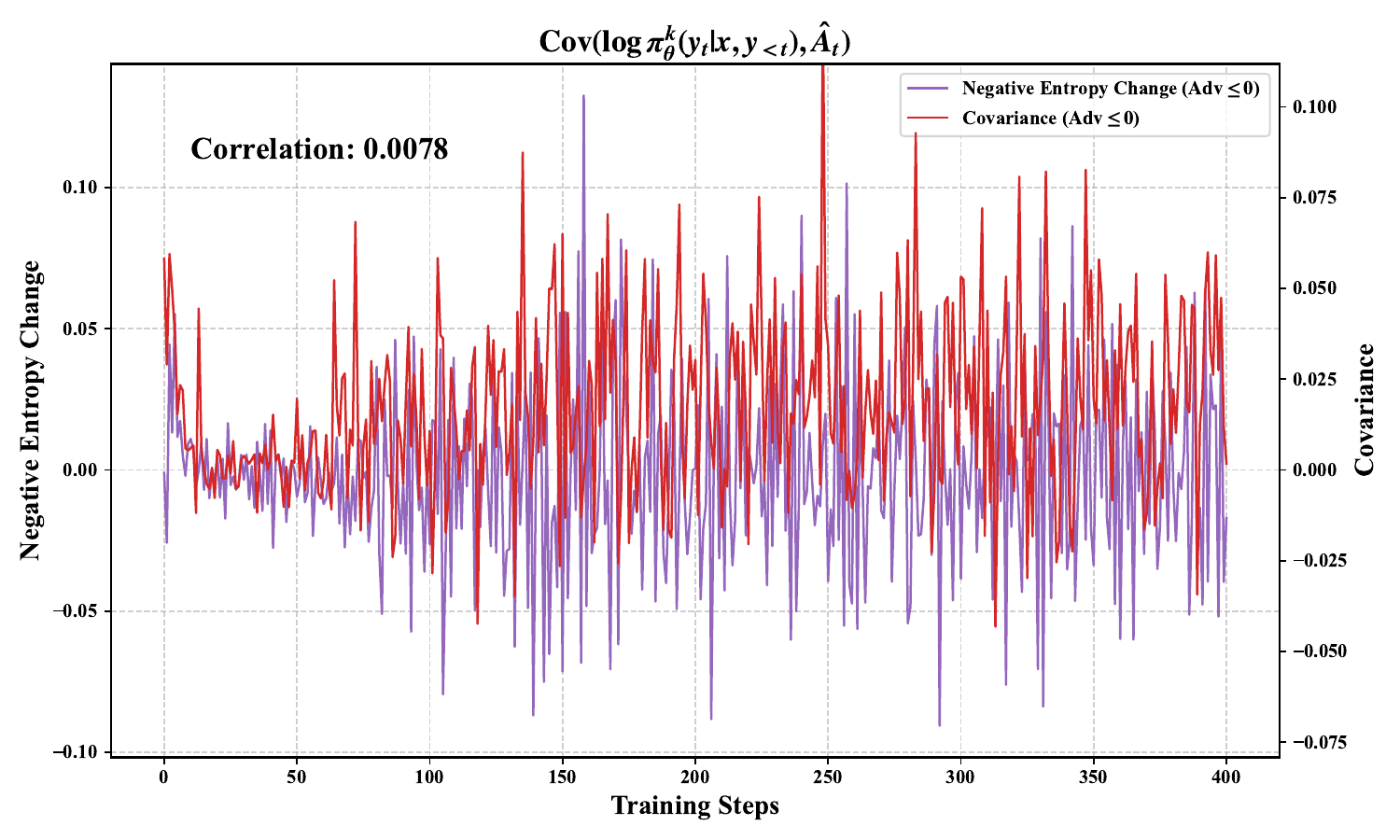}
    \end{subfigure}
    \caption{Evolution of the negative entropy change and the covariance terms during GRPO training of Qwen2.5-Math-7B with $N_{\text{update}} = 4$, where updates are performed exclusively on tokens with $\text{Adv} \le 0$. The left figure reports the covariance between $\log \pi_\theta^k \left(\bm{y}_t \vert \bm{x}, \bm{y}_{<t} \right)$ and $\pi_\theta^k \left(\bm{y}_t \vert \bm{x}, \bm{y}_{<t} \right) \hat{A}_t$, while the right panel reports the covariance between $\log \pi_\theta^k \left(\bm{y}_t \vert \bm{x}, \bm{y}_{<t} \right)$ and $\hat{A}_t$.}
    \label{fig:entropy_change_cov_plot_037}
\end{figure*}

\begin{figure*}[t]
    \centering
    \begin{subfigure}{0.495\textwidth}
        \includegraphics[width=\linewidth]{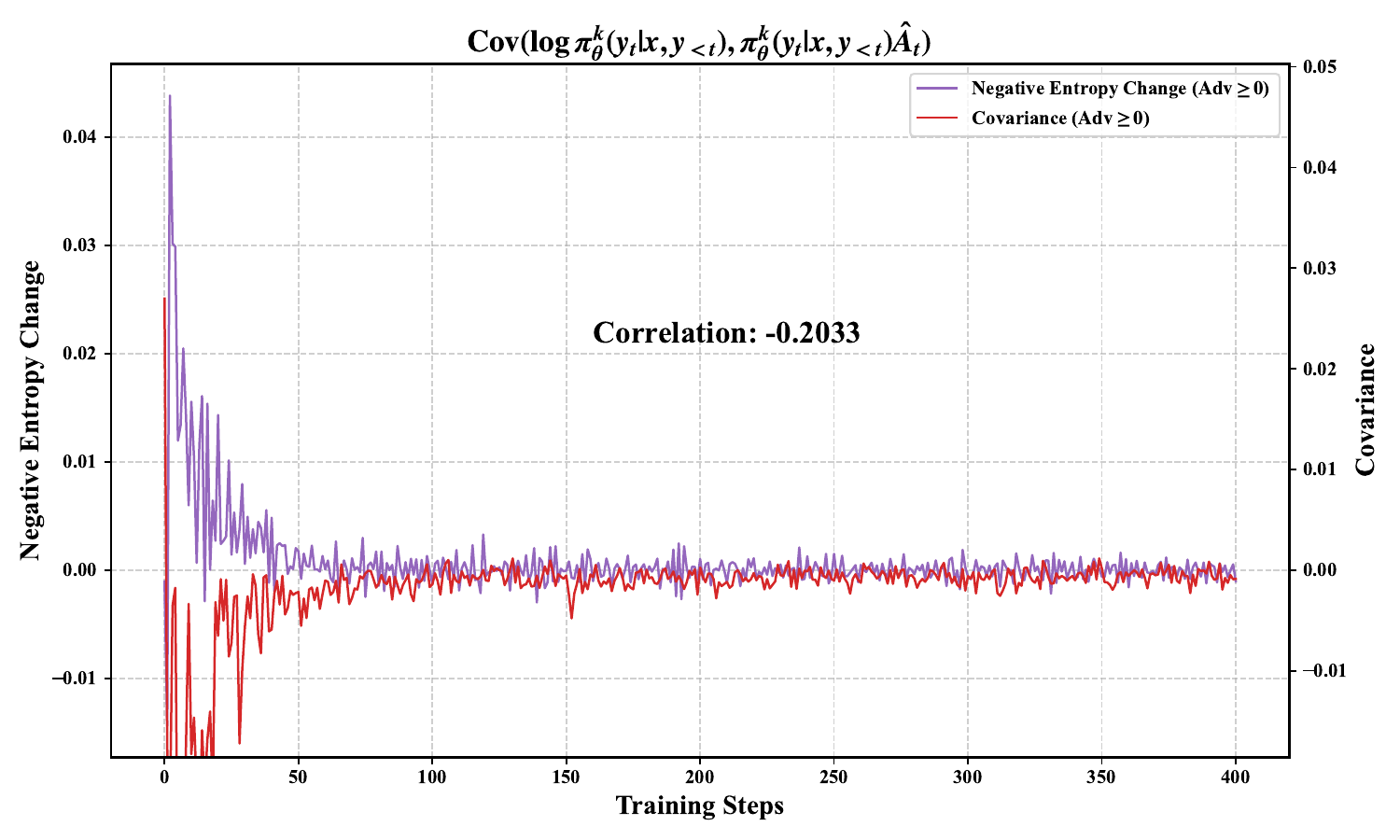}
    \end{subfigure}
    \begin{subfigure}{0.495\textwidth}
        \includegraphics[width=\linewidth]{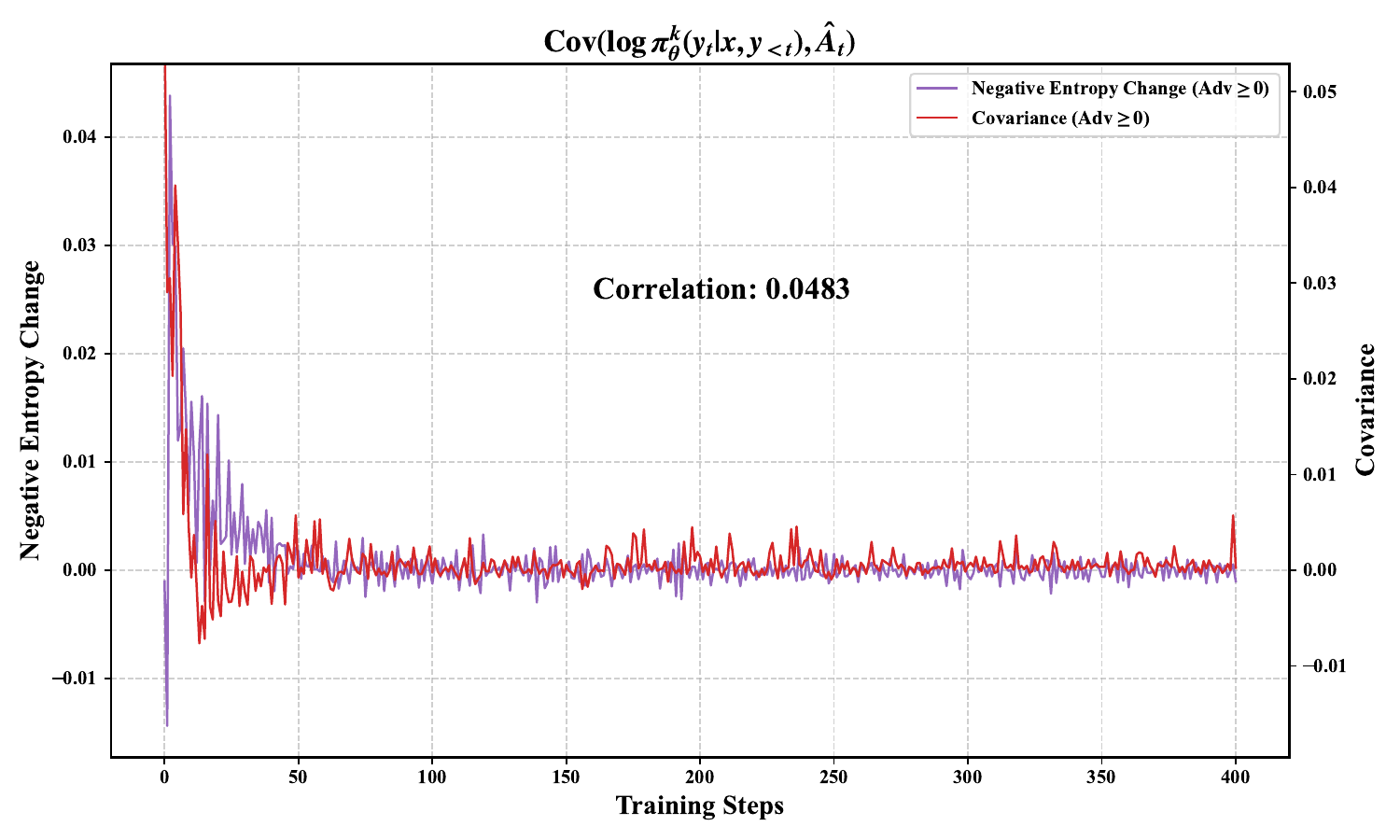}
    \end{subfigure}
    \caption{Evolution of the negative entropy change and the covariance terms during GRPO training of Qwen2.5-Math-7B with $N_{\text{update}} = 4$, where updates are performed exclusively on tokens with $\text{Adv} \ge 0$. The left figure reports the covariance between $\log \pi_\theta^k \left(\bm{y}_t \vert \bm{x}, \bm{y}_{<t} \right)$ and $\pi_\theta^k \left(\bm{y}_t \vert \bm{x}, \bm{y}_{<t} \right) \hat{A}_t$, while the right panel reports the covariance between $\log \pi_\theta^k \left(\bm{y}_t \vert \bm{x}, \bm{y}_{<t} \right)$ and $\hat{A}_t$.}
    \label{fig:entropy_change_cov_plot_038}
\end{figure*}

Similarly, under the natural policy gradient, the entropy change between steps $k$ and $k+1$ can be expressed as follows:
\begin{equation}
\small
\begin{aligned}
& \mathcal{H} \left (\pi_\theta^{k+1} \left (\bm{y_t} \vert \bm{x}, \bm{y}_{<t} \right) \right) - \mathcal{H}\left (\pi_\theta^k \left ( \bm{y}_t \vert \bm{x}, \bm{y}_{<t} \right) \right) \approx \\
& -\eta \cdot \text{Cov} \left(\log \pi_\theta^k \left(\bm{y}_t \vert \bm{x}, \bm{y}_{<t} \right), \hat{A}_t \right).
\end{aligned}
\end{equation}

To connect our empirical analysis with these theoretical derivations, we reran the experiments on Qwen2.5-Math-7B to record both $\text{Cov} (\log \pi_\theta^k (\bm{y}_t \vert \bm{x}, \bm{y}_{<t} ), \pi_\theta^k (\bm{y}_t \vert \bm{x}, \bm{y}_{<t}) \hat{A}_t )$ and $\text{Cov} (\log \pi_\theta^k (\bm{y}_t \vert \bm{x}, \bm{y}_{<t}), \hat{A}_t)$ throughout training. Specifically, we considered the following settings:
\begin{itemize}
    \item GRPO with $N_{\text{update}} = 1$;
    \item GRPO with $N_{\text{update}} = 4$;
    \item GRPO with $N_{\text{update}} = 4$ trained exclusively on tokens with $\text{Adv} \le 0$;
    \item GRPO with $N_{\text{update}} = 4$ trained exclusively on tokens with $\text{Adv} \ge 0$;
    \item GRPO with $N_{\text{update}} = 4$ augmented with \textbf{Pos-Adv-Reweight (Entropy-guided)}.
\end{itemize}

\begin{figure*}[t]
    \centering
    \begin{subfigure}{0.495\textwidth}
        \includegraphics[width=\linewidth]{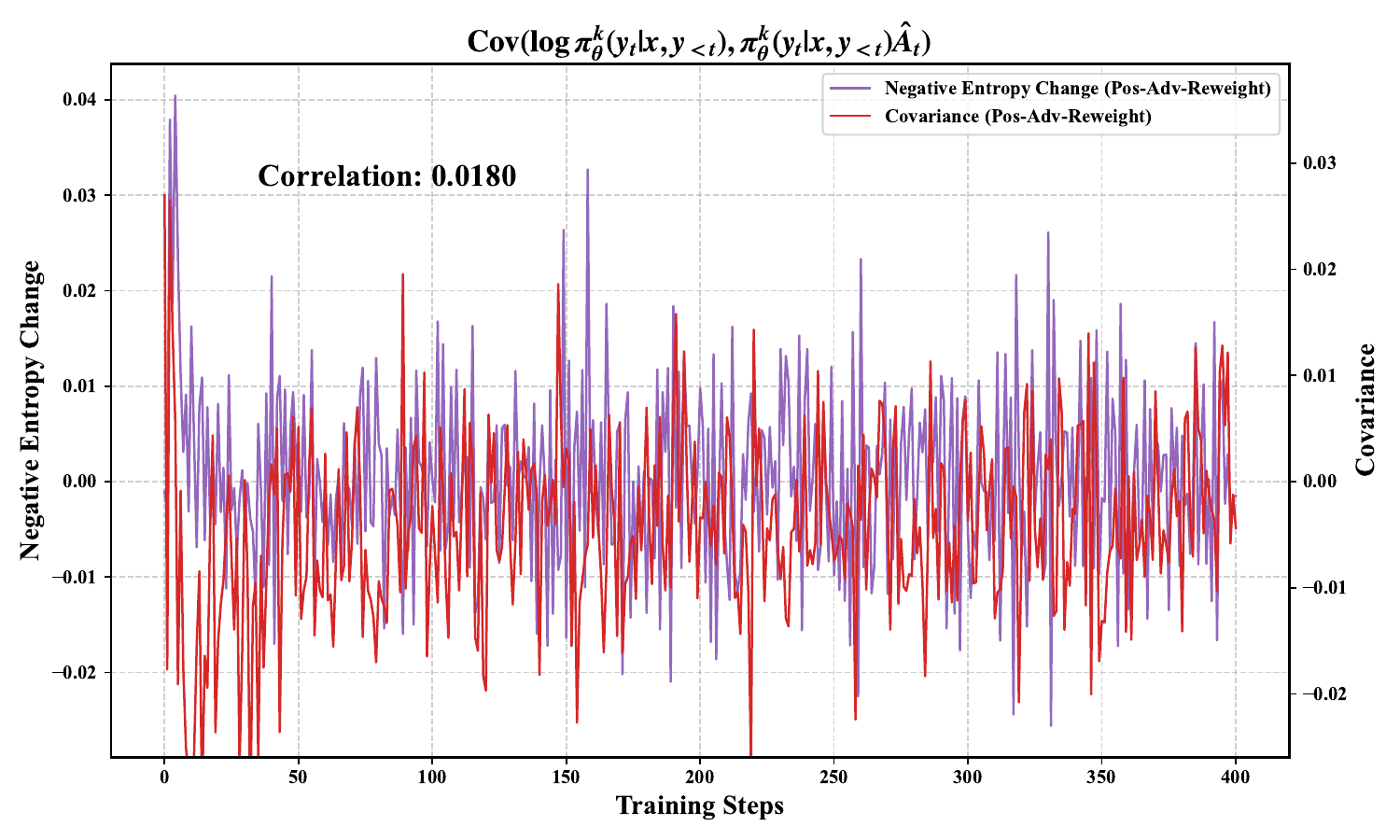}
    \end{subfigure}
    \begin{subfigure}{0.495\textwidth}
        \includegraphics[width=\linewidth]{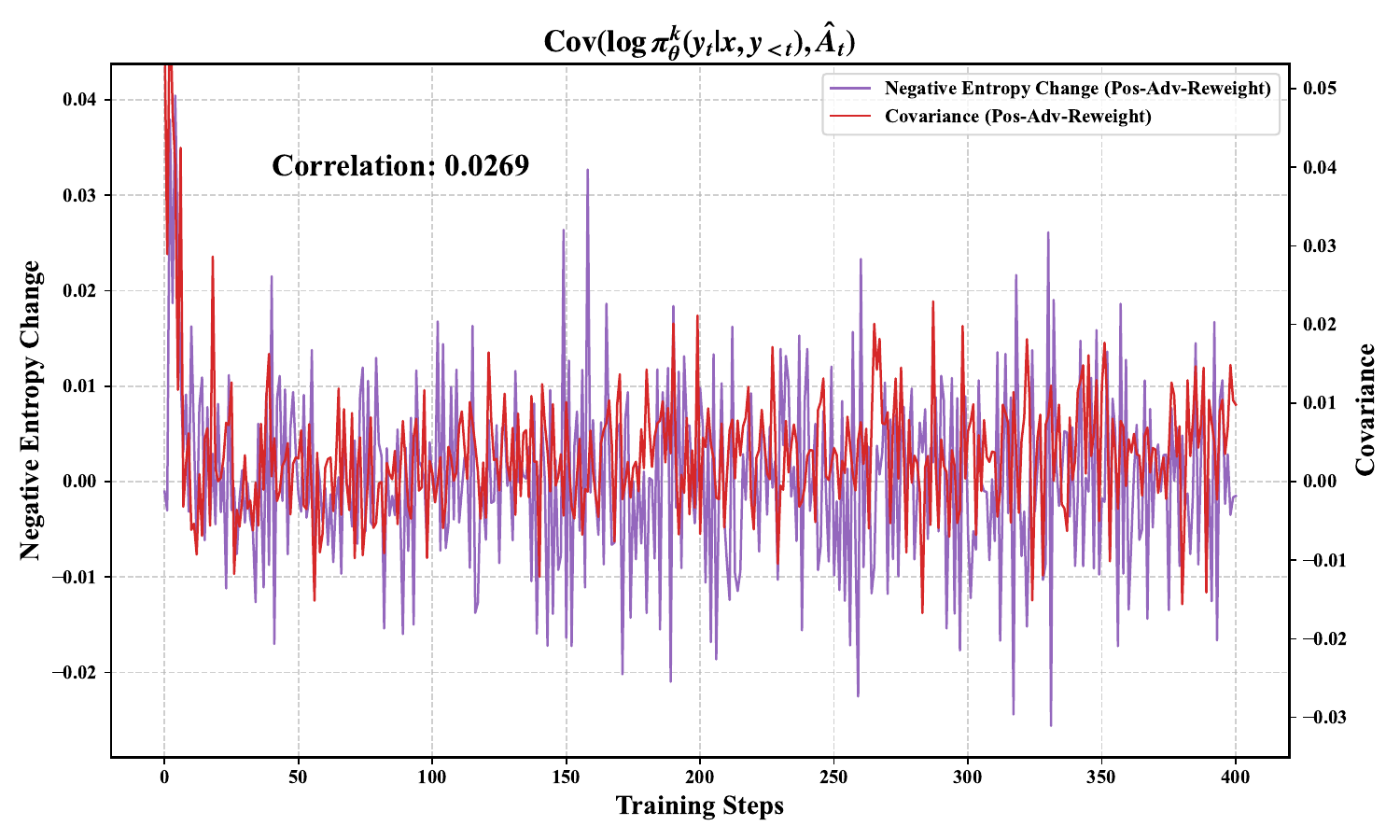}
    \end{subfigure}
    \caption{Evolution of the negative entropy change and the covariance terms during GRPO training augmented with \textbf{Pos-Adv-Reweight (Entropy-guided)} on Qwen2.5-Math-7B with $N_{\text{update}} = 4$. The left figure reports the covariance between $\log \pi_\theta^k \left(\bm{y}_t \vert \bm{x}, \bm{y}_{<t} \right)$ and $\pi_\theta^k \left(\bm{y}_t \vert \bm{x}, \bm{y}_{<t} \right) \hat{A}_t$, while the right panel reports the covariance between $\log \pi_\theta^k \left(\bm{y}_t \vert \bm{x}, \bm{y}_{<t} \right)$ and $\hat{A}_t$.}
    \label{fig:entropy_change_cov_plot_056}
\end{figure*}

Figure~\ref{fig:entropy_change_cov_plot_001} to \ref{fig:entropy_change_cov_plot_056} illustrate the evolution of the negative entropy change and the corresponding covariance terms under the above settings, as well as the Spearman's rank correlation coefficients between the negative entropy change and each covariance term. As shown in these figures, both the negative entropy change and the covariance terms exhibit noticeable fluctuations during training, while the absolute values of the Spearman's rank correlation coefficients remain generally small.

We hypothesize that the weak empirical correlations between the covariance terms and the corresponding entropy changes arise from a mismatch between the optimizer assumed in the theoretical analysis and that used in practice. Specifically, the theoretical derivation of the relationship between covariance terms and entropy change assumes SGD as the optimizer when deriving the exact updates of token logits \citep{zhihu2025entropy,DBLP:journals/corr/abs-2505-22617}. In contrast, LLMs are typically trained with the AdamW optimizer \citep{DBLP:journals/corr/KingmaB14,DBLP:conf/iclr/LoshchilovH19}, which employs adaptive learning rates based on first- and second-moment estimates of the gradients. This discrepancy complicates the computation of exact token logit updates and may lead to deviations between theoretical predictions and the observed entropy dynamics.

\section{Experiments on Llama-3.1-8B-Instruct}
\label{sec:appendix_experiments_on_llama_3_1_8b_instruct}

\subsection{Experimental Setup}
To assess whether our empirical findings generalize beyond Qwen2.5-Math-7B, we further trained Llama-3.1-8B-Instruct using GRPO.\footnote{We initially intended to train Llama-3.1-8B with GRPO, as it is a pretrained model that has not undergone instruction tuning, similar to Qwen2.5-Math-7B. However, during preliminary experiments, we observed that Llama-3.1-8B frequently generated endlessly repetitive responses during training, which substantially slowed down the training process and degraded training stability. Consequently, we selected Llama-3.1-8B-Instruct, which has undergone instruction tuning, to ensure more efficient and stable training.} The training configuration for Llama-3.1-8B-Instruct followed that of Qwen2.5-Math-7B, as described in Section~\ref{sec:experimental_setup}. Specifically, Llama-3.1-8B-Instruct was trained on the DAPO-Math-17K dataset \citep{DBLP:journals/corr/abs-2503-14476} with the AdamW optimizer, employing a cosine learning rate schedule and a peak learning rate of $1 \times 10^{-6}$. During rollout, the batch size was set to 256, and 16 responses were generated per prompt using top-$p$ sampling with $p=1.0$, a temperature of 1.0, and a maximum generation length of 4096 tokens. For evaluation, we largely followed the setup used for Qwen2.5-Math-7B, with one exception: MATH500 was used as the validation set. This choice was motivated by the generally poor performance of GRPO-trained Llama-3.1-8B on AIME 2025, which makes AIME 2025 an unreliable validation benchmark in this setting.

\begin{figure}[t]
    \centering
    \includegraphics[width=\linewidth]{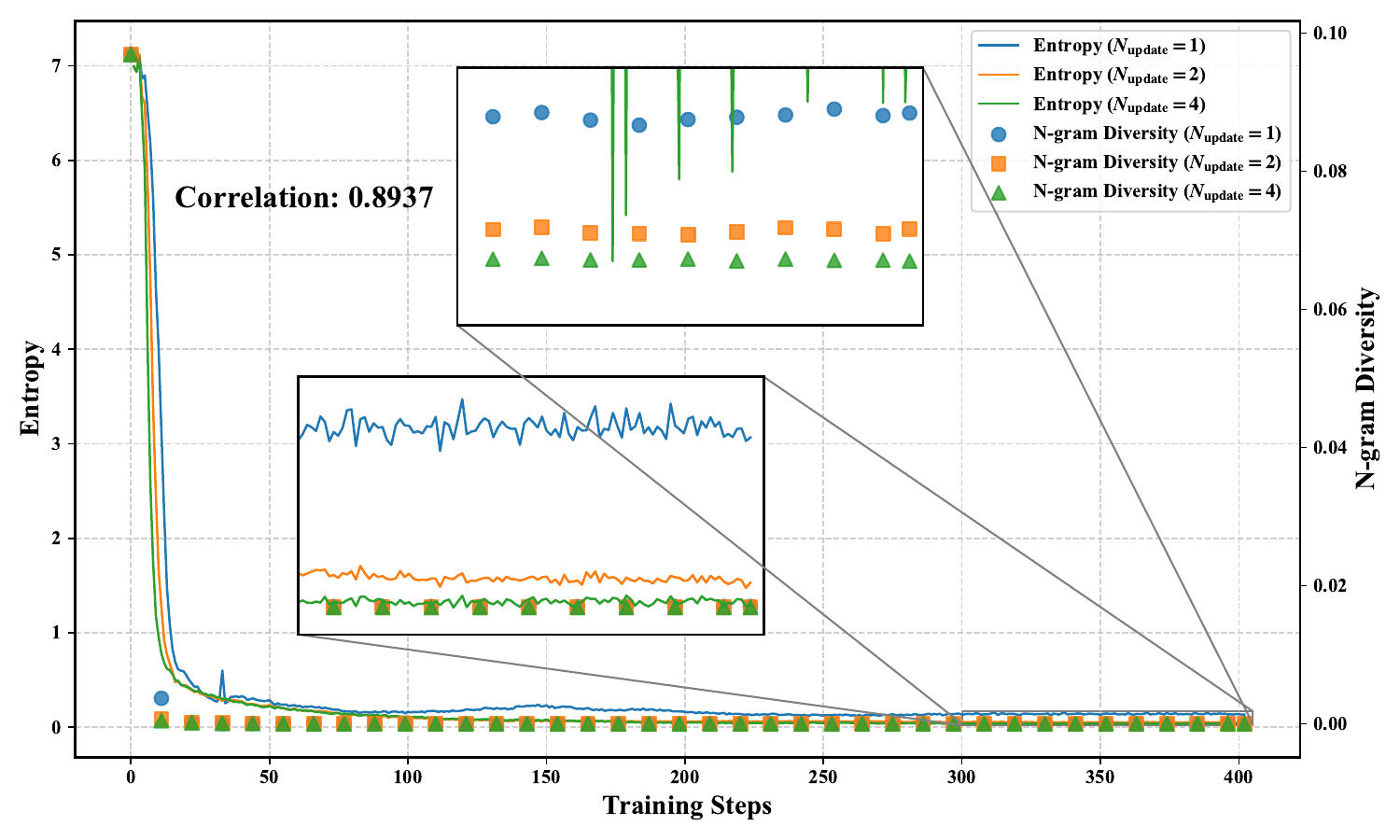}
    \caption{Evolution of entropy (solid lines) and N-gram diversity (markers) across training steps under varying numbers of off-policy updates for Llama-3.1-8B-Instruct trained with GRPO.}
    \label{fig:entropy_ngram_diversity_plot_aime24_llama}
\end{figure}

\begin{figure}[t]
    \centering
    \includegraphics[width=\linewidth]{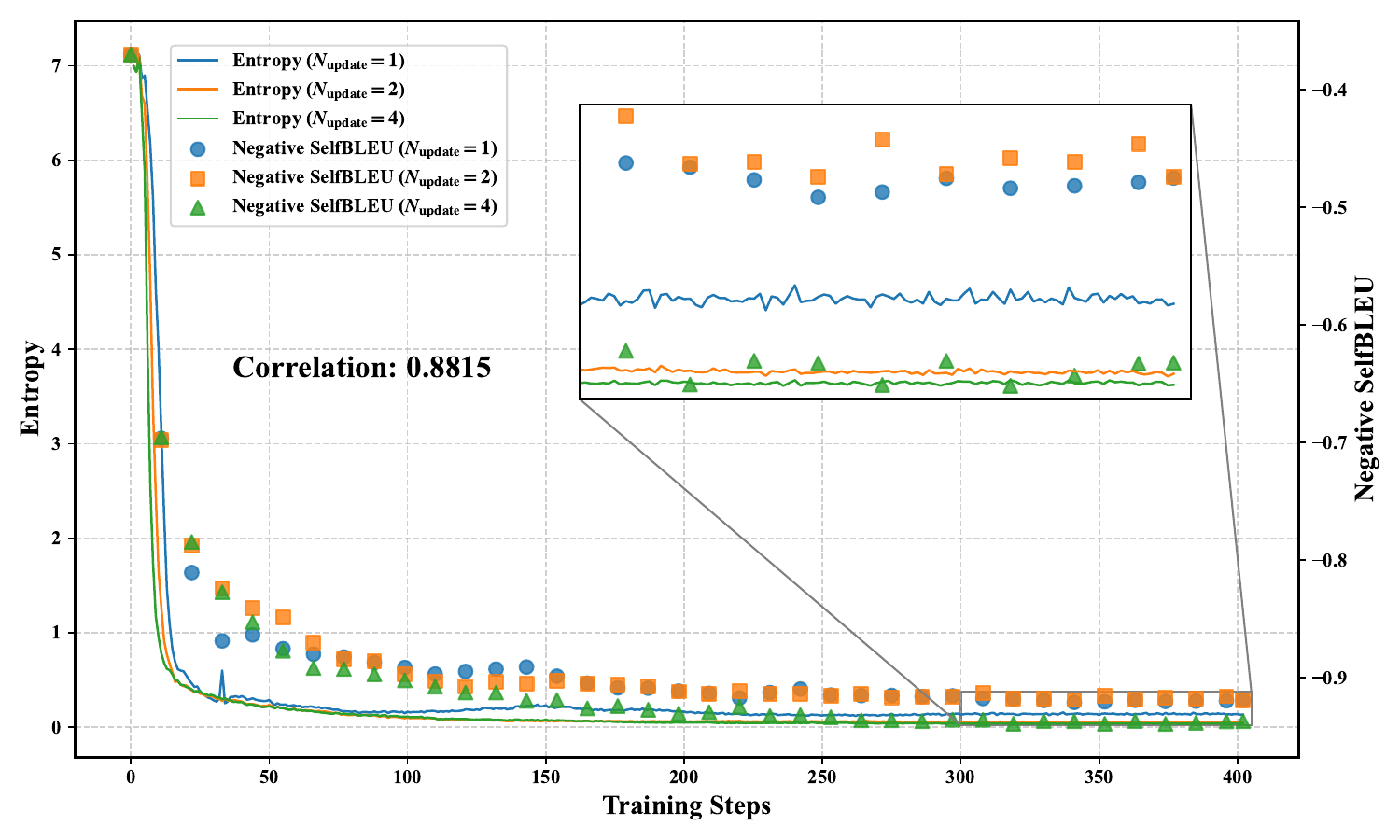}
    \caption{Evolution of entropy (solid lines) and negative SelfBLEU scores (markers) across training steps under varying numbers of off-policy updates for Llama-3.1-8B-Instruct trained with GRPO.}
    \label{fig:entropy_selfbleu_plot_aime24_llama}
\end{figure}

\subsection{Entropy and Response Diversity}
Figure~\ref{fig:entropy_ngram_diversity_plot_aime24_llama} and Figure~\ref{fig:entropy_selfbleu_plot_aime24_llama} illustrate the entropy dynamics of Llama-3.1-8B-Instruct trained with GRPO under varying numbers of off-policy updates, together with the diversity of generated responses across training steps, measured by N-gram Diversity and SelfBLEU, respectively. As shown in Figure~\ref{fig:entropy_ngram_diversity_plot_aime24_llama}, increasing the number of off-policy updates leads to a reduction in model entropy during training, which is accompanied by a corresponding decline in the diversity of the generated responses. Notably, a strong positive correlation is observed between entropy and N-gram Diversity, with a Spearman's rank correlation coefficient of 0.8937. This result indicates that the empirical relationship between entropy and response diversity identified for Qwen2.5-Math-7B in Section~\ref{subsec:appendix_entropy_and_response_diversity} also holds for Llama-3.1-8B-Instruct.

\subsection{Entropy Dynamics on Prompts}
\begin{figure}[t]
    \centering
    \includegraphics[width=\linewidth]{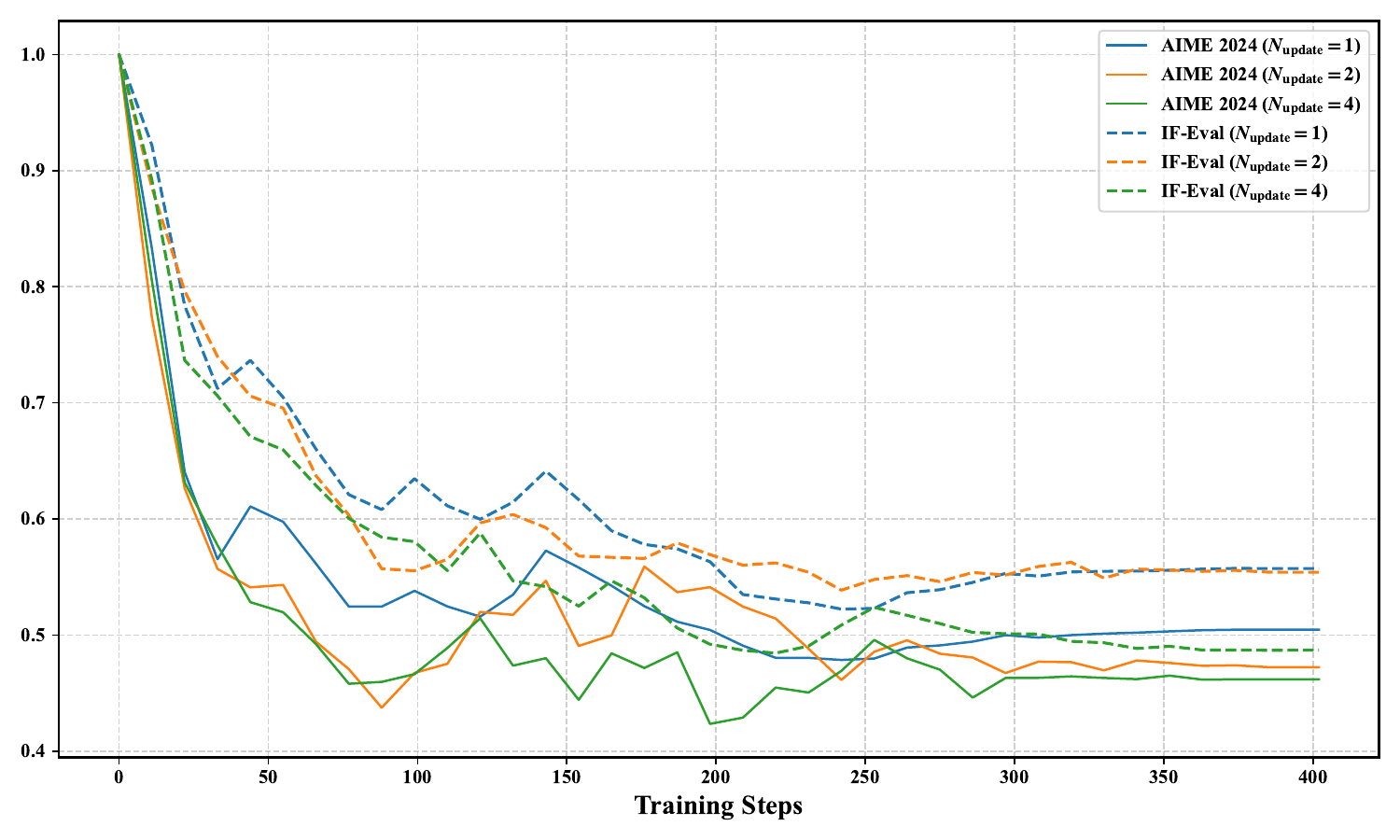}
    \caption{Ratio of the entropy of Llama-3.1-8B-Instruct at different training steps to its initial entropy prior to training, under varying numbers of off-policy updates.}
    \label{fig:entropy_prompts_aime24_llama}
\end{figure}

Figure~\ref{fig:entropy_prompts_aime24_llama} presents the ratio between the entropy of Llama-3.1-8B-Instruct measured at different training steps and its initial entropy prior to training, across different numbers of off-policy updates. As illustrated in Figure~\ref{fig:entropy_prompts_aime24_llama}, the entropy measured on prompts drawn from the same domain as the training data decreases more rapidly than that measured on prompts from different domains. A similar trend is observed for Qwen2.5-Math-7B, indicating that the empirical findings reported in Section~\ref{subsec:appendix_entropy_dynamics_on_prompts} also hold for Llama-3.1-8B-Instruct.

\subsection{Prompt Entropy and Accuracy}
\begin{figure}[t]
    \centering
    \includegraphics[width=\linewidth]{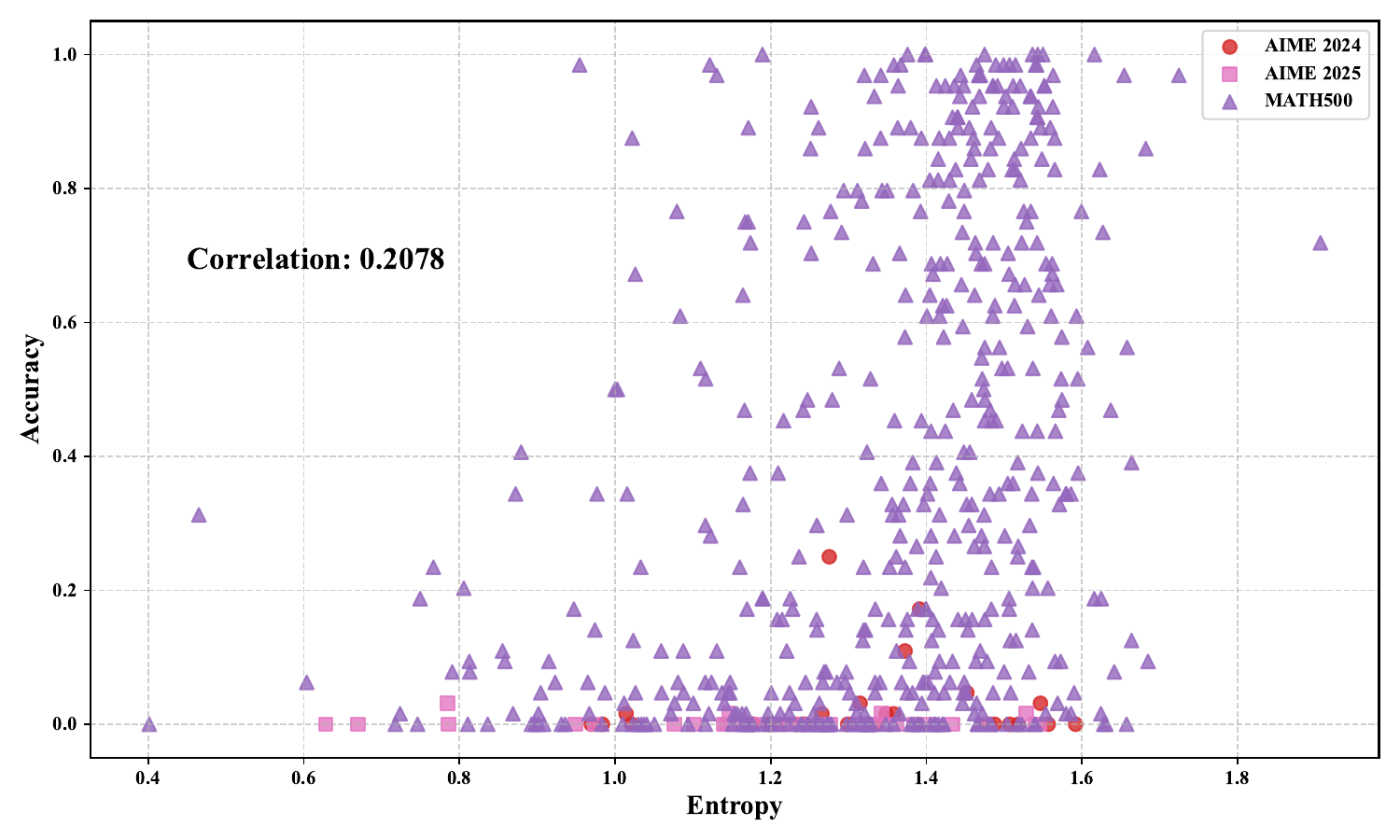}
    \caption{Scatter plot of accuracy versus entropy for Llama-3.1-8B-Instruct evaluated on AIME 2024, AIME 2025, and MATH500.}
    \label{fig:entropy_acc_prompts_llama}
\end{figure}

\begin{figure}[t]
    \centering
    \includegraphics[width=\linewidth]{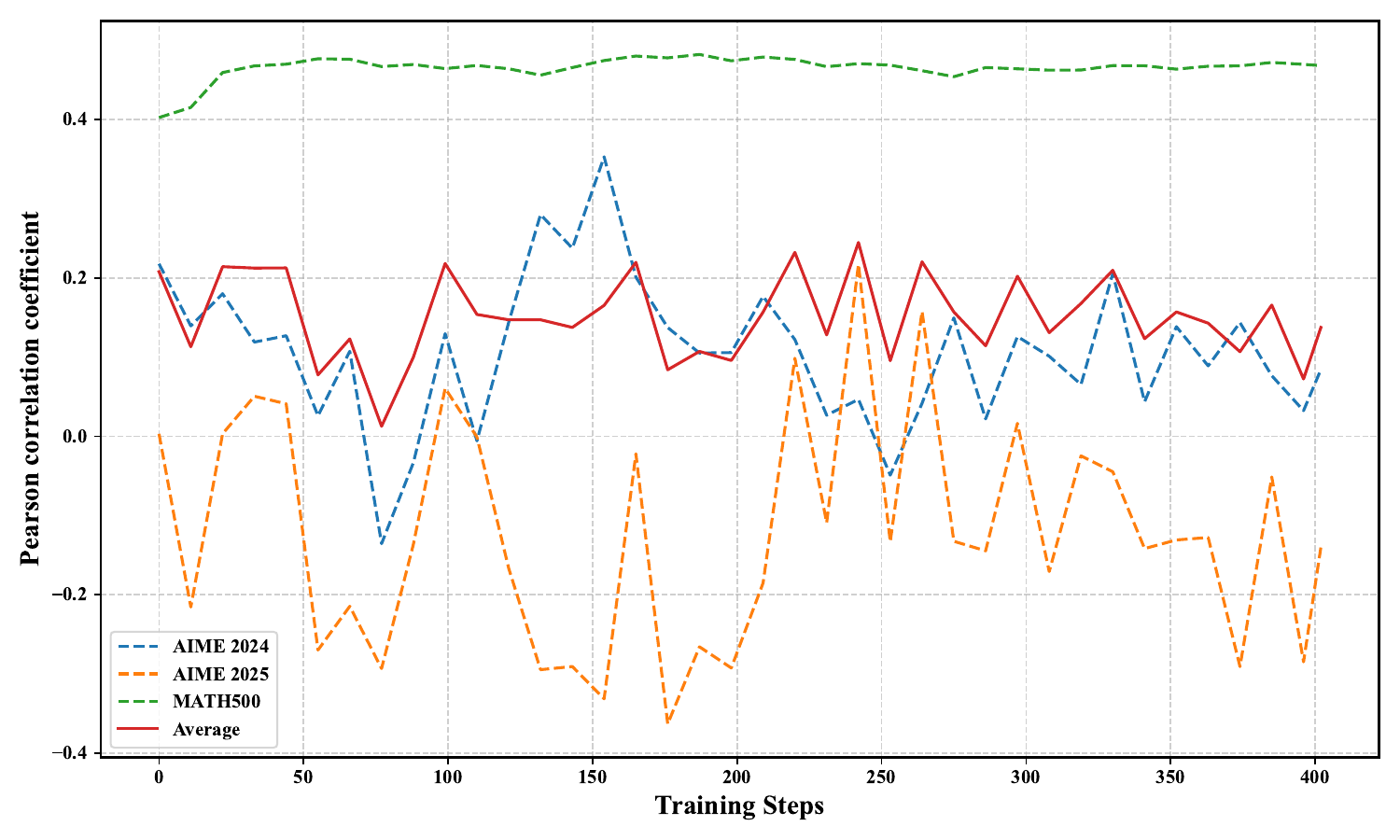}
    \caption{Spearman's rank correlation coefficients between entropy and accuracy for Llama-3.1-8B-Instruct, measured at checkpoints saved at different training steps. ``Average'' indicates the mean Spearman's rank correlation coefficient computed over AIME 2024, AIME 2025, and MATH500.}
    \label{fig:entropy_acc_prompts_corrcoef_plot_llama}
\end{figure}

Figure~\ref{fig:entropy_acc_prompts_llama} illustrates the relationship between accuracy and entropy for Llama-3.1-8B-Instruct across AIME 2024, AIME 2025, and MATH500. As shown in the figure, the entropy measured on prompts exhibits only a weak correlation with the accuracy of the corresponding responses. Quantitatively, the average Spearman's rank correlation coefficient between prompt entropy and response accuracy across the three benchmarks is 0.2078. Furthermore, Figure~\ref{fig:entropy_acc_prompts_corrcoef_plot_llama} shows that the Spearman's rank correlation coefficient remains consistently small throughout GRPO training. These results indicate that the empirical finding reported in Section~\ref{subsec:appendix_prompt_entropy_and_accuracy}, namely that prompt entropy is only weakly correlated with model accuracy, also holds for Llama-3.1-8B-Instruct.

\begin{table}[t]
  \centering
  \small
    \begin{tabular}{lrr}
    \toprule
          & \multicolumn{1}{c}{\textbf{Avg@64}} & \multicolumn{1}{c}{\textbf{Pass@64}} \\
    \midrule
    \textbf{AIME 2024} & -0.08571 & 0.39466 \\
    \textbf{AIME 2025} & -0.34786 & -0.46291 \\
    \textbf{MATH500} & -0.02857 & -0.31429 \\
    \textbf{AMC 2023} & 0.02857 & -0.26482 \\
    \textbf{Minerva} & -0.48571 & -0.11595 \\
    \textbf{LiveCodeBench} & -0.54286 & \textbf{-0.88571} \\
    \textbf{IF-Eval} & \textbf{-0.82857} & 0.46382 \\
    \bottomrule
    \end{tabular}
  \caption{Spearman's rank correlation coefficients between the entropy of LLMs and their performance across different benchmarks. Coefficients with the largest absolute values are highlighted in bold.}
  \label{tab:spearman_correlation_coefficients_llama}
\end{table}

\subsection{Correlations Between Entropy and Model Performance}
Table~\ref{tab:spearman_correlation_coefficients_llama} reports the Spearman's rank correlation coefficients between model entropy and the Avg@64 and Pass@64 performance metrics for Llama-3.1-8B-Instruct trained with GRPO and its variants. As shown in Table~\ref{tab:spearman_correlation_coefficients_llama}, entropy exhibits a strong negative correlation with Avg@64 on IF-Eval and a strong negative correlation with Pass@64 on LiveCodeBench. These results indicate that the empirical finding obtained with Qwen2.5-Math-7B in Section~\ref{subsec:correlations_between_entropy_and_model_performance}, namely that the relationship between LLM entropy and benchmark performance depends on both the task and the evaluation metric, also holds for Llama-3.1-8B-Instruct.

\begin{figure}[t]
    \centering
    \includegraphics[width=\linewidth]{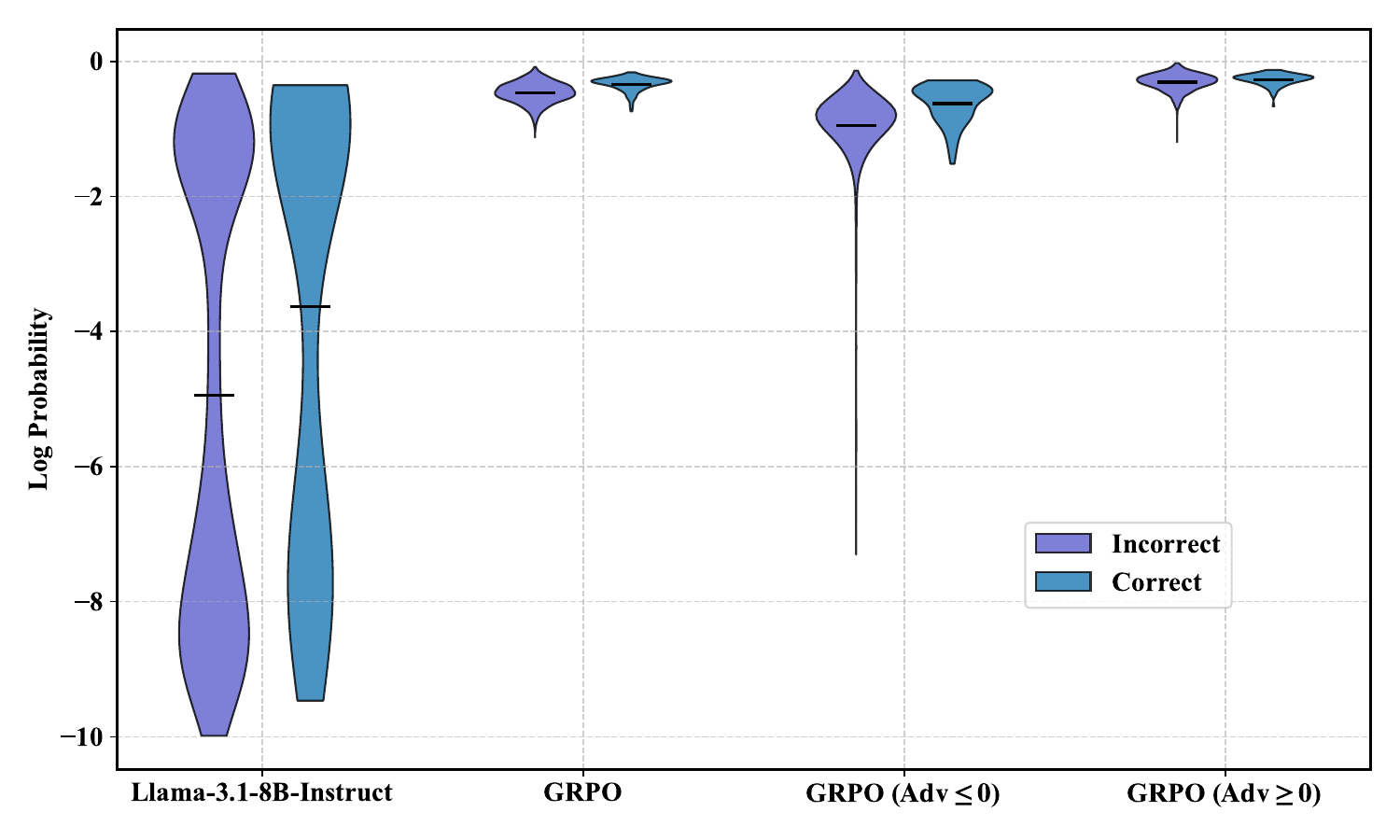}
    \caption{Distribution of log probabilities for correct and incorrect responses produced by Llama-3.1-8B-Instruct and its GRPO-trained variants. In each violin plot, the black line denotes the mean value.}
    \label{fig:response_acc_probs_llama}
\end{figure}

\subsection{Entropy Collapse and Miscalibration}
Figure~\ref{fig:response_acc_probs_llama} presents the distributions of log probabilities, as computed by the corresponding LLMs, for correct and incorrect responses generated by Llama-3.1-8B-Instruct and its GRPO-trained variants. As illustrated in Figure~\ref{fig:response_acc_probs_llama}, responses produced by Llama-3.1-8B-Instruct trained with GRPO and its variants are assigned higher probabilities than those produced by Llama-3.1-8B-Instruct. This observation suggests that GRPO training increases the model’s confidence in the responses it generates.

Moreover, consistent with the empirical findings for Qwen2.5-Math-7B, Llama-3.1-8B-Instruct assigns higher average probabilities to correct responses than to incorrect ones, and this pattern also holds for its GRPO-trained variants. However, after GRPO training, the gap in average probability between correct and incorrect responses becomes smaller, indicating that Llama-3.1-8B-Instruct becomes increasingly miscalibrated.

\begin{table*}[t]
  \centering
  \resizebox{\textwidth}{!}{
    \begin{tabular}{l|l|l|l|l|l|c|l|>{\columncolor{gray!15}}c|>{\columncolor{gray!15}}c|>{\columncolor{gray!15}}c}
    \toprule
          \multicolumn{1}{c|}{\textbf{Model}} & \multicolumn{1}{c|}{\textbf{AIME 2024}} & \multicolumn{1}{c|}{\textbf{AIME 2025}} & \multicolumn{1}{c|}{\textbf{MATH500}} & \multicolumn{1}{c|}{\textbf{AMC 2023}} & \multicolumn{1}{c|}{\textbf{Minerva}} & \multicolumn{1}{c|}{\textbf{LiveCodeBench}} & \multicolumn{1}{c|}{\textbf{IF-Eval}} & \textbf{Average (ID)} & \textbf{Average (OOD)} & \textbf{Entropy} \\
    \midrule
    Llama-3.1-8B-Instruct & 2.34 / 33.33 & 0.26 / 13.33 & 38.50 / 84.60 & 15.98 / 87.50 & 16.06 / 53.31 & 8.36 / 30.88 & 74.42 / 96.16 & 14.63 / 54.42 & 41.39 / 63.52 & N/A \\
    \arrayrulecolor{black}\midrule
    + GRPO ($N_{\text{update}}=1$) & 8.85 / 36.67 & 0.57 / 16.67 & 55.08 / 87.00 & 39.65 / 72.50 & 25.99 / 55.51 & 10.54 / 29.41 & 77.90 / 93.88 & 26.03 / 53.67 & 44.22 / 61.65 & 0.13570 \\
    \arrayrulecolor{lightgray}\midrule
    + GRPO ($N_{\text{update}}=2$) & 7.29 / 43.33 & 0.47 / 13.33 & 52.33 / 90.60 & 25.55 / 80.00 & 25.61 / 59.93 & 9.61 / 31.62 & 78.06 / 95.08 & 22.25 / 57.44 & 43.84 / 63.35 & 0.04994 \\
    \midrule
    + GRPO ($N_{\text{update}}=4$) & 5.52 / 30.00 & 0.47 / 16.67 & 49.82 / 91.00 & 26.64 / 87.50 & 24.33 / 61.76 & 9.77 / 30.51 & 77.62 / 94.84 & 21.35 / 57.39 & 43.69 / 62.68 & 0.03781 \\
    \arrayrulecolor{black}\midrule
    \: \: + $\text{Adv} \leq 0$ & 3.13 / 40.00 & 0.52 / 16.67 & 43.17 / 89.80 & 18.52 / 80.00 & 20.19 / 58.46 & 8.46 / 29.04 & 76.00 / 95.80 & 17.10 / 56.98 & 42.23 / 62.42 & 2.18019 \\
    \arrayrulecolor{lightgray}\midrule
    \: \: + $\text{Adv} \geq 0$ & 5.99 / 36.67 & 0.73 / 26.67 & 49.40 / 89.60 & 23.63 / 85.00 & 25.14 / 55.88 & 10.85 / 31.99 & 79.43 / 94.48 & 20.98 / 58.76 & 45.14 / 63.23 & 0.02846 \\
    \midrule
    \: \: + Pos-Adv-Reweight (Entropy-guided) & 7.03 / 36.67 & 0.31 / 13.33 & 50.16 / 89.40 & 26.88 / 87.50 & 23.78 / 58.46 & 9.90 / 30.15 & 77.27 / 94.84 & 21.63 / 57.07 & 43.59 / 62.50 & 0.19455 \\
    \arrayrulecolor{black}\bottomrule
    \end{tabular}
  }
  \caption{Performance of Llama-3.1-8B-Instruct trained with GRPO and its variants.}
  \label{tab:all_results_llama}
\end{table*}

Notably, when Llama-3.1-8B-Instruct is trained exclusively on tokens with no-positive advantages, which are expected to reduce the probabilities of responses generated during the rollout stage, the average probabilities of both correct and incorrect responses nonetheless increase. In this setting, the difference in average probability between correct and incorrect responses is larger than that observed under GRPO training, indicating that miscalibration is alleviated relative to GRPO. In contrast, when Llama-3.1-8B-Instruct is trained exclusively on tokens with non-negative advantages, the difference in average probability between correct and incorrect responses becomes even smaller than that observed under GRPO training, thereby further exacerbating miscalibration.

Taken together, these results demonstrate that the empirical finding reported in Section~\ref{subsec:entropy_collapse_and_miscalibration}, namely that training LLMs with GRPO can induce miscalibration, also holds for Llama-3.1-8B-Instruct.

\subsection{Effect of Off-Policy Updates}
\begin{figure}[t]
    \centering
    \includegraphics[width=\linewidth]{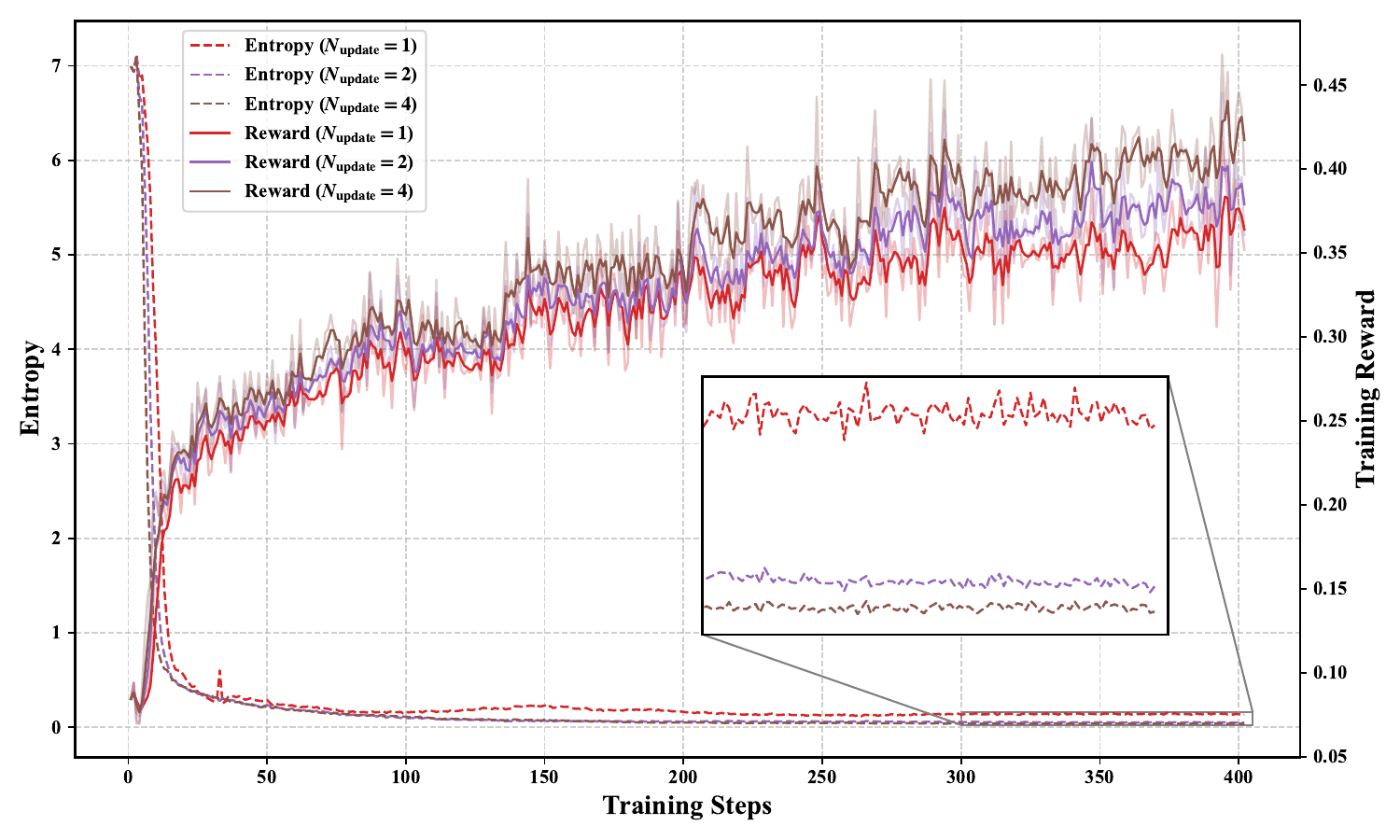}
    \caption{Evolution of entropy and reward on the training set during GRPO training of Llama-3.1-8B-Instruct.}
    \label{fig:entropy_train_reward_llama}
\end{figure}

Figure~\ref{fig:entropy_train_reward_llama} illustrates the entropy dynamics and the evolution of training reward when Llama-3.1-8B-Instruct is trained with GRPO under different numbers of off-policy updates. As shown in Figure~\ref{fig:entropy_train_reward_llama}, increasing the number of off-policy updates from 1 to 2 results in a more rapid decrease in entropy, and further increasing the number to 4 accelerates the entropy reduction even further. In addition, a larger number of off-policy updates leads to higher rewards on the training set for Llama-3.1-8B-Instruct trained with GRPO. In contrast, the test performance measured by the Avg@64 score deteriorates as the number of off-policy updates increases. These results indicate that the empirical findings discussed in Section~\ref{subsec:off_policy_updates} also apply to Llama-3.1-8B-Instruct.

\subsection{Experimental Results of Pos-Adv-Reweight}

\begin{figure}[t]
    \centering
    \includegraphics[width=\linewidth]{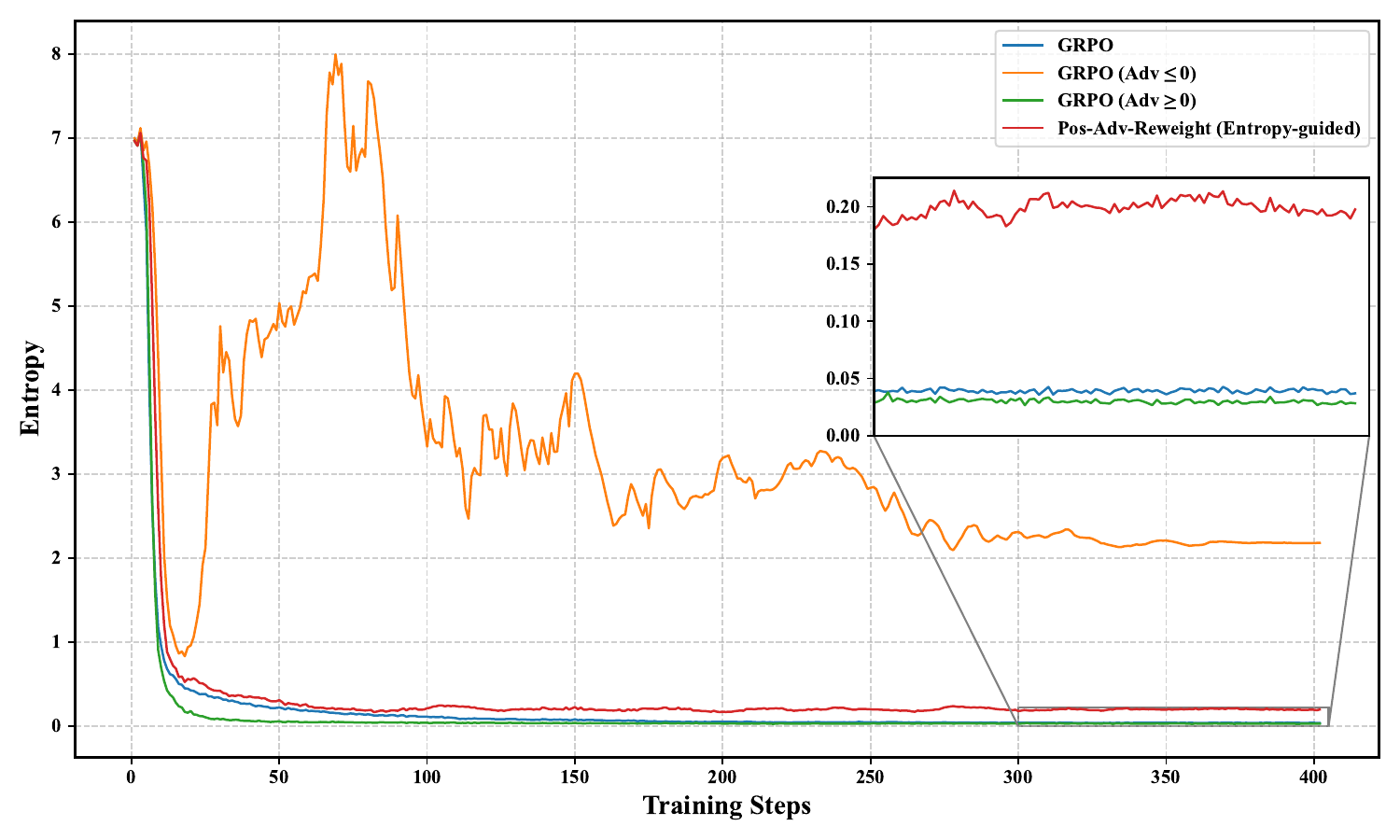}
    \caption{Evolution of entropy for Llama-3.1-8B-Instruct trained with GRPO and its variants.}
    \label{fig:entropy_plot_llama}
\end{figure}

To evaluate the effectiveness of \textbf{Pos-Adv-Reweight} on Llama-3.1-8B-Instruct, we train the model using GRPO with four off-policy updates, incorporating \textbf{Pos-Adv-Reweight (Entropy-guided)} under the same experimental setup as Qwen2.5-Math-7B. The entropy dynamics of Llama-3.1-8B-Instruct trained with GRPO and its variants are shown in Figure~\ref{fig:entropy_plot_llama}, and the corresponding performance on multiple benchmarks is reported in Table~\ref{tab:all_results_llama}.

As illustrated in Figure~\ref{fig:entropy_plot_llama}, Llama-3.1-8B-Instruct trained with GRPO, as well as the variant trained exclusively on tokens with non-negative advantages, both suffer from entropy collapse. In contrast, training exclusively on tokens with non-positive advantages results in pronounced entropy fluctuations. Conversely, training with \textbf{Pos-Adv-Reweight (Entropy-guided)} maintains the model entropy at a stable level of approximately 0.2 during training. Furthermore, the results in Table~\ref{tab:all_results_llama} indicate that restricting training to either non-negative-advantage tokens or non-positive-advantage tokens leads to inferior average Avg@64 scores on in-domain benchmarks compared with GRPO. By contrast, Llama-3.1-8B-Instruct trained with \textbf{Pos-Adv-Reweight (Entropy-guided)} outperforms GRPO in terms of average Avg@64 scores. These results further demonstrate that \textbf{Pos-Adv-Reweight}, which adaptively adjusts the loss weights of tokens with positive advantages, effectively controls the entropy of LLMs while improving their overall performance.

\end{document}